\documentclass[letterpaper]{article} 
\usepackage{aaai25}  
\usepackage{times}  
\usepackage{helvet}  
\usepackage{courier}  
\usepackage[hyphens]{url}  
\usepackage{graphicx} 
\usepackage{natbib}  
\usepackage{caption} 
\usepackage{algorithm}
\usepackage{algorithmic}
\usepackage{amssymb,amsmath}
\usepackage{booktabs}
\usepackage{multirow}
\usepackage{dsfont,pifont} 

\newcommand{\ie}{\textit{i}.\textit{e}., }
\newcommand{\eg}{\textit{e}.\textit{g}., }

\providecommand{\fref}[1]{Figure~\ref{#1}}

\usepackage{newfloat}
\usepackage{listings}
\DeclareCaptionStyle{ruled}{labelfont=normalfont,labelsep=colon,strut=off} 
\floatstyle{ruled}
\newfloat{listing}{tb}{lst}{}
\floatname{listing}{Listing}
\title{Generalized Zero-Shot Learning for Point Cloud Segmentation with Evidence-Based Dynamic Calibration}
\author{
    Hyeonseok Kim\textsuperscript{\rm 1}, Byeongkeun Kang\textsuperscript{\rm 1}, Yeejin Lee\textsuperscript{\rm 1}\thanks{Corresponding Author.}\\
}
\affiliations{
    \textsuperscript{\rm 1}Seoul National University of Science and Technology, Republic of Korea\\
    \{khslab, byeongkeun.kang, yeejinlee\}@seoultech.ac.kr
}

\usepackage{bibentry}

\begin{document}

\maketitle

\begin{abstract}
Generalized zero-shot semantic segmentation of 3D point clouds aims to classify each point into both seen and unseen classes. A significant challenge with these models is their tendency to make biased predictions, often favoring the classes encountered during training. This problem is more pronounced in 3D applications, where the scale of the training data is typically smaller than in image-based tasks. To address this problem, we propose a novel method called E3DPC-GZSL, which reduces overconfident predictions towards seen classes without relying on separate classifiers for seen and unseen data. E3DPC-GZSL tackles the overconfidence problem by integrating an evidence-based uncertainty estimator into a classifier. This estimator is then used to adjust prediction probabilities using a dynamic calibrated stacking factor that accounts for pointwise prediction uncertainty. In addition, E3DPC-GZSL introduces a novel training strategy that improves uncertainty estimation by refining the semantic space. This is achieved by merging learnable parameters with text-derived features, thereby improving model optimization for unseen data. Extensive experiments demonstrate that the proposed approach achieves state-of-the-art performance on generalized zero-shot semantic segmentation datasets, including ScanNet v2 and S3DIS.
\end{abstract}
\section{Introduction}
Semantic segmentation of 3D point clouds refers to a task that classifies each point into a specific semantic category. Most existing methods for this task mainly use supervised learning techniques that rely on a labeled dataset where each point is already categorized~\cite{c:215, c:217}.  Although these supervised models perform well at segmenting categories they have been trained on, they face challenges when dealing with novel or previously unseen categories, reducing their effectiveness in real-world scenarios. This limitation arises because these methods heavily rely on labeled training data, which may not cover all possible real-world scenarios or objects. 

To overcome this limitation, zero-shot learning~(ZSL) provides a valuable alternative by enabling models to generalize learned knowledge to novel, previously unseen categories. However, even with zero-shot learning, achieving accurate segmentation for these novel categories remains a significant challenge. This is particularly critical in high-stakes 3D applications such as autonomous driving and medical imaging, where precise segmentation is crucial for safety and reliability. 

\begin{figure}[t]
    \centering
    \begin{minipage}{0.99\linewidth}
        \centering
        \includegraphics[width=0.99\linewidth]{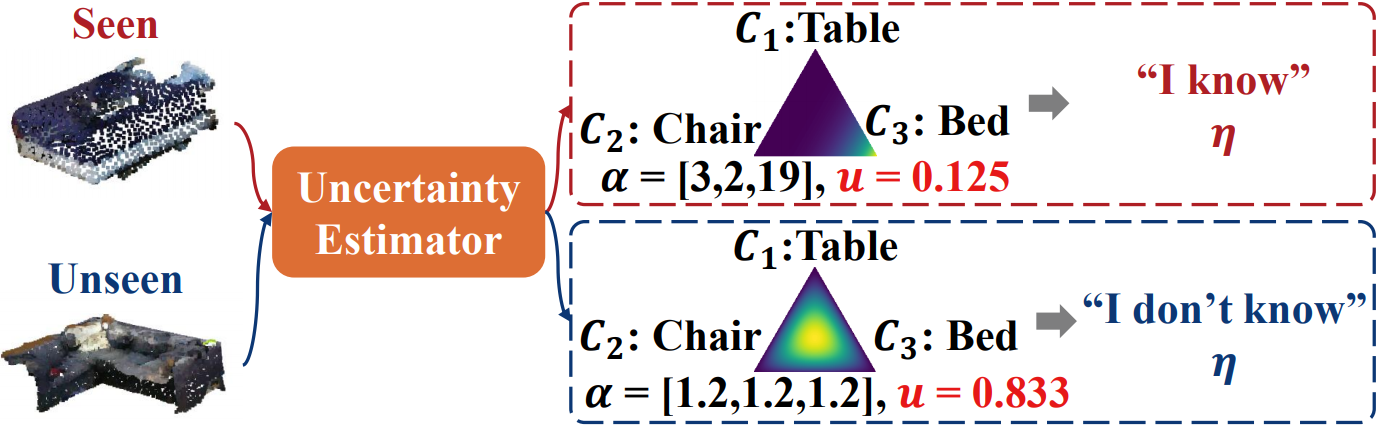}\\ 
    \end{minipage}
    \\
    \begin{minipage}{0.99\linewidth}
        \centering
        \includegraphics[width=0.99\linewidth]{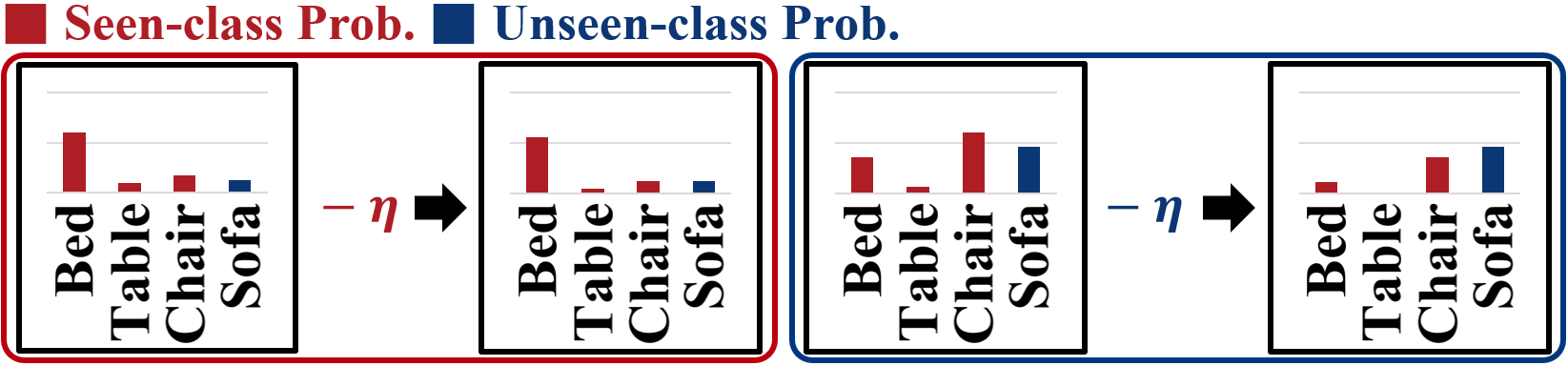}\\
    \end{minipage}
    \\
    \caption{An illustration of E3DPC-GZSL. E3DPC-GZSL mitigates the challenge of the overconfidence problem, as shown in the bottom figures, by using a dynamic calibration~($\eta$). The upper figure shows how $\eta$ is obtained from the uncertainty estimate ($u$), which is parameterized by the evidence ($\alpha$), resulting in an adaptive calibration for both the seen (bottom left) and unseen (bottom right) datasets.} 
   \label{fig:concept}
\end{figure}

Despite its significant potential impact in various applications, research on ZSL in the 3D domain has received limited attention in the literature. Early efforts to apply ZSL to point clouds~\cite{c:223} focused on a classification task by learning to map point cloud features into a word embedding space. Subsequent studies~\cite{c:224} introduced an unsupervised skewness loss to address the hubness problem, which arises from the phenomenon that the nearest neighbors of many data points converge to a single hub in a high-dimensional space. 

To address more practical scenarios, these techniques have been extended to generalized zero-shot learning~(GZSL). GZSL seeks to recognize both seen and unseen categories during inference. There are two different configurations to achieve GZSL: transductive and inductive settings. Transductive settings allow the use of unlabeled data, including unseen points without labels, whereas inductive settings strictly avoid using unseen points during the training phase. A critical issue with GZSL models trained in this inductive setting is that they often produce biased predictions, with a tendency to favor the seen categories used during training~\cite{c:237}. 

\begin{figure}[t]
    \centering
    \begin{tabular}{@{}c@{}c@{}}
         \hspace{-0.2cm} \includegraphics[width=0.44\linewidth]{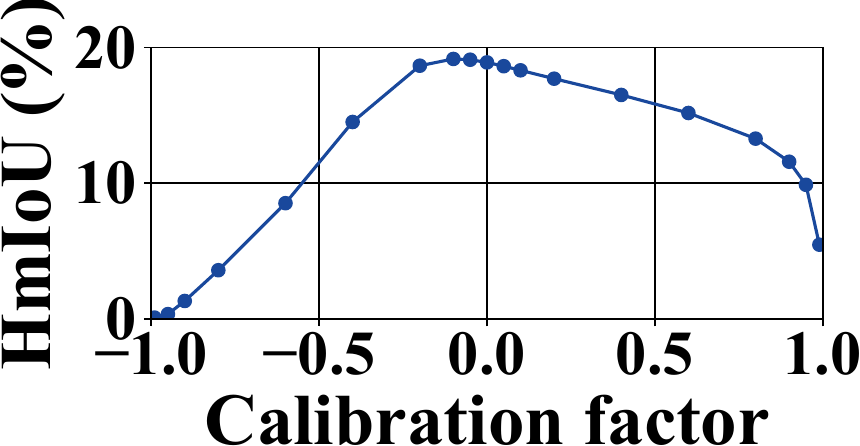} & \includegraphics[width=0.54\linewidth]{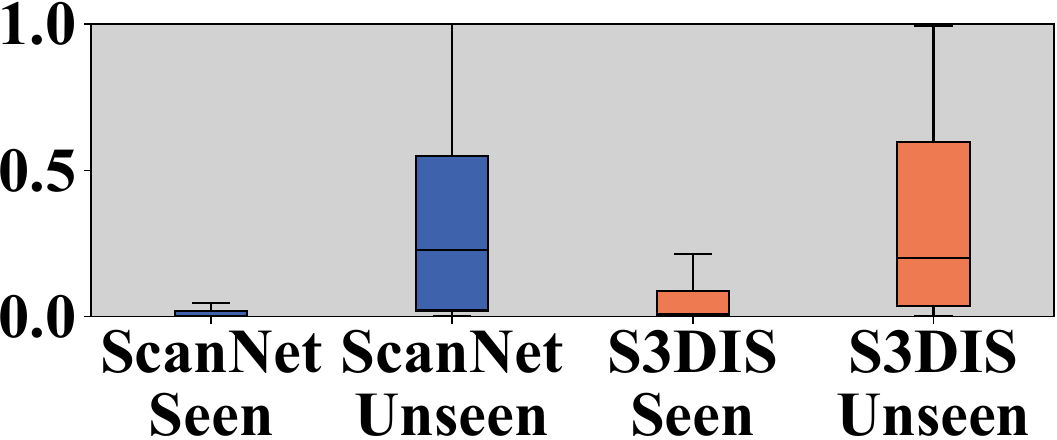}\\
         {\footnotesize (a)} & {\footnotesize (b)}\\
    \end{tabular}
    \caption{(a) Segmentation performance variation with different calibration factors on S3DIS. (b) Uncertainty difference between seen and unseen points.} \label{fig:motivation}
\end{figure}

To overcome this problem, two main approaches are commonly used: 1) the binary classification approach and 2) the calibrated stacking approach. Methods using the binary classification approach~\cite{c:235, c:2215} first use binary classification to determine whether the input data belong to seen or unseen categories. For seen categories, a supervised task is applied, while for unseen categories, a ZSL task is applied. In the calibrated stacking approach, the models~~\cite{c:237, c:225, c:226} adjust the final prediction probability of the GZSL model by reducing the probability of seen categories using predefined hyperparameters. By lowering the probability of seen categories, it increases the relative probability of unseen categories. However, both approaches rely heavily on hyperparameters, as demonstrated in Figure \ref{fig:motivation}(a), and face the challenge of applying the same hyperparameters consistently across all input data. Figure \ref{fig:motivation}(a) shows how performance varies with different predefined calibration factors, indicating that performance is dependent on them. (See also the section on uncertainty estimation).

To tackle this issue, we propose a novel method named E3DPC-GZSL, which integrates two main approaches in GZSL. First, E3DPC-GZSL enables point-wise calibrated stacking without relying on predefined hyperparameters. Unlike previous methods that apply the same stacking parameter to all samples, E3DPC-GZSL introduces a learnable calibration parameter. This parameter adjusts prediction probabilities based on the characteristics of individual samples. It is derived from estimated uncertainty levels, which act as an indicator for identifying unseen samples, but without the need for explicitly distinguishing between seen and unseen samples~(See Figure \ref{fig:motivation}(b)). By evaluating the prediction evidence of reliable seen data with labels, E3DPC-GZSL can implicitly distinguish between seen and unseen samples and perform dynamic calibration to the predicted class probabilities. This calibration adjusts proportionally to the uncertainty of the unseen samples -- larger adjustments are made for higher uncertainties, while smaller adjustments are made for lower uncertainties. Moreover, E3DPC-GZSL introduces a new training strategy for data augmentation for unseen classes to overcome data scarcity issues. Unlike the typical visual-text space alignment in ZSL, this strategy incorporates a new approach to semantic space refinement by fusing learnable tuning parameters into text-derived features. The proposed strategy tunes text embeddings and aligns feature vectors to the refined space, resulting in a better understanding of scenes.

In summary, our contributions are as follows:
\begin{itemize}
    \item We show that the performance of existing methods can vary when using the standard calibrated stacking approach.
    \item We propose a novel method, E3DPC-GZSL, which mitigates the overconfidence of zero-shot models on seen categories by redistributing prediction probabilities using estimated uncertainty.
    \item We propose a new strategy for tuning semantic embeddings to overcome data scarcity in 3D zero-shot learning.
    \item We show that the proposed method outperforms state-of-the-art~(SOTA) methods for both seen and unseen classes. Extensive analysis confirms the effectiveness of our approach for generalized zero-shot semantic segmentation in 3D datasets.
\end{itemize}

\section{Related Works} \label{sec:related_works}

\textbf{Point Cloud Semantic Segmentation.}
Several approaches have been proposed to accomplish semantic segmentation of 3D point clouds.
One such method uses a multilayer perceptron~(MLP) that processes individual points as input, including models~\cite{c:211, c:212} specifically designed to handle unordered points directly.
Another approach employs point-wise convolution techniques, which extract features by applying kernel operations to the point cloud~\cite{c:213, c:214}.
This includes sparse convolution methods~\cite{c:215, c:216}, which map the point cloud to grid cells and perform 3D convolution operations only where data exists to extract features.
More recently, transformer-based methods~\cite{c:217, r:211} have been proposed to encode the point cloud using attention mechanisms. However, all these methods have been designed for fully supervised learning with labels. In contrast, few methods have been developed for unsupervised settings.

In the GZSL setting, a generator usually synthesizes unseen samples using semantic information, which is then used to train semantic segmentation classifiers~\cite{c:225, c:226}. Certain methods adopt a pseudo-labeling approach for the unseen data~\cite{c:228, r:222}. Additionally, some approaches incorporate extra information, such as projected 2D images~\cite{c:229} or geometric primitives~\cite{c:2210}, to enhance segmentation performance.

\noindent \textbf{Mitigating Bias of Prediction towards Seen Categories.} 
ZSL models typically train on samples from seen categories, resulting in a bias toward those categories. This bias is more pronounced in inductive settings, where unseen samples are not available during training. To address this issue, one approach is to distinguish samples as either seen or unseen and then categorize them accordingly.
This approach effectively breaks down the GZSL task into a supervised learning task for seen categories and a ZSL task for unseen categories. Some methods use gating networks that perform binary classification before classifying the specific category of the input data~\cite{c:235, c:2215}. A more recent method~\cite{r:232} uses multiple gating networks to classify data not only as seen or unseen, but also as ambiguous.

Another approach is to adjust the predicted probabilities or scores of a model to more accurately represent the true probability or confidence of each prediction. 
A simple but effective calibration technique, known as calibrated stacking, is used in point cloud semantic segmentation models~\cite{c:225, c:226} by increasing the probabilities of unseen predictions to a specified level. 

However, both approaches heavily rely on hyperparameters and face the challenge of applying the same hyperparameters consistently across all input data.
\section{Problem Definition}
\textbf{ZSL Setup.} Given a point cloud, a training set $\mathcal{D}_{tr}$ contains $N_{tr}$ labeled point samples: $\mathcal{D}_{tr} = \left\{ \left( \mathbf{x}_i, y_i\right)\right\}_{i=1}^{N_{tr}}$. Each sample in the training set consists of a point $\mathbf{x}_i \in \mathbb{R}^{N_p}$ from the point cloud and its corresponding label $y_i$.
In $\mathcal{D}_{tr}$, the labels come from a set of $\mathcal{Y}^{s}$ of $N_s$ seen classes. The point $\mathbf{x}$ can comprise either 3D spatial coordinates or spatial coordinates with color components $(r, g, b)$. Additionally, during training, class description vectors $\mathbf{t} \in \mathbb{R}^{N_t}$, either associated with $N_s$ seen classes or $N_u$ unseen classes, can be accompanied by $y$ in the form of semantic attributes or natural language embeddings. Note that the experiments presented in this paper consider the inductive setting, where only class description vectors are provided and no unseen points are included. 

At inference, a set of points $\mathcal{D}_{te}^{u}=\left\{ \mathbf{x}_i\right\}_{i=1}^{N_{te}^{u}}$ is provided from a set $\mathcal{Y}^{u}$ of $N_{u}$ unseen classes that is completely disjoint from $\mathcal{Y}^{s}$. In other words, there are no overlapping classes between $\mathcal{Y}^{u}$ and $\mathcal{Y}^{s}$, meaning $\mathcal{Y}^{u} \cap \mathcal{Y}^{s} = \emptyset$.

\noindent \textbf{GZSL Setup.}
In the ZSL setting, samples are drawn exclusively from unseen classes during inference. However, the GZSL setting allows samples to be drawn from both seen and unseen classes. In this setting, the inference set $\mathcal{D}_{te}=\left\{ \mathbf{x}_i\right\}_{i=1}^{N_{te}}$ consists of sample points from $\mathcal{Y} = \mathcal{Y}^{u} \cup \mathcal{Y}^{s}$.

\noindent \textbf{Generalized Zero-Shot Semantic Segmentation.} 
Given the label set $\mathcal{Y} = \left\{c_k\right\}_{k=1}^{N_c}$, where $N_c$ is the total number of seen and unseen classes~($N_s+N_u$), generalized zero-shot semantic segmentation can be achieved by introducing a weight matrix $\mathbf{w}_c \in \mathbb{R}^{N_f \times N_c}$. This matrix maps an input point $\mathbf{x}$ to the output space and is obtained by using the feature vector $\mathbf{f}$ extracted by a trained encoder $E$, \ie $\mathbf{f} = E(\mathbf{x}) \in \mathbb{R}^{N_f}$. 

Concretely, a classifier for 3D semantic segmentation can be implemented by computing class posterior probabilities $p$ from a logit vector $\boldsymbol{\ell} = \mathbf{w}_{c}^{T}\mathbf{f}$, as follows: 
\begin{align} \label{eq:class_posterior}
    p_k = p\left(c_{k}|\mathbf{x}\right) = \frac{e^{\ell_{k}}}{\sum_{j=1}^{N_c} e^{\ell_{j}} },
\end{align}
where $\ell_k$ and $\ell_j$ are the $k$-th and $j$-th elements of the vector $\boldsymbol{\ell}$, respectively, with $k$ ranging from 1 to $N_c$. The predicted class for a given $\mathbf{x}$ is then computed as $\hat{y} = \underset{c_k}{\mathrm{argmax}}\:\: p_k$. 
The classifier is typically optimized using the cross-entropy loss:
\begin{align} \label{eq:cross_entropy}
    \mathcal{L}_{CE} = -\frac{1}{N_b}\sum\limits_{j=1}^{N_b}\sum\limits_{c_k \in \mathcal{Y}} \mathds{1}\left(c_k = c_{y_j} \right) \log p_{k},
\end{align}
where $\mathds{1}(\cdot)$ is the indicator function and $N_b$ is the number of points in a minibatch. The cross-entropy loss function reaches its minimum when the predicted probabilities perfectly match the ground truth labels for all samples, \ie when the predicted probability for the correct class is $1$, and for all other classes, it is $0$. Thus, minimizing the cross-entropy loss tends to amplify the differences between the logit values, in particular by increasing the highest logit while suppressing the others. This happens because the loss function heavily penalizes incorrect classifications, forcing the model to make highly confident predictions for the most likely class~\cite{c:310}. However, this tendency to make overly confident predictions can be problematic in GZSL. When the model encounters novel unseen classes, it may struggle to generalize effectively because it has been trained to make very sharp distinctions between the classes it has seen before. 
\begin{figure}[t]
    \centering
    \includegraphics[width=0.99\linewidth]{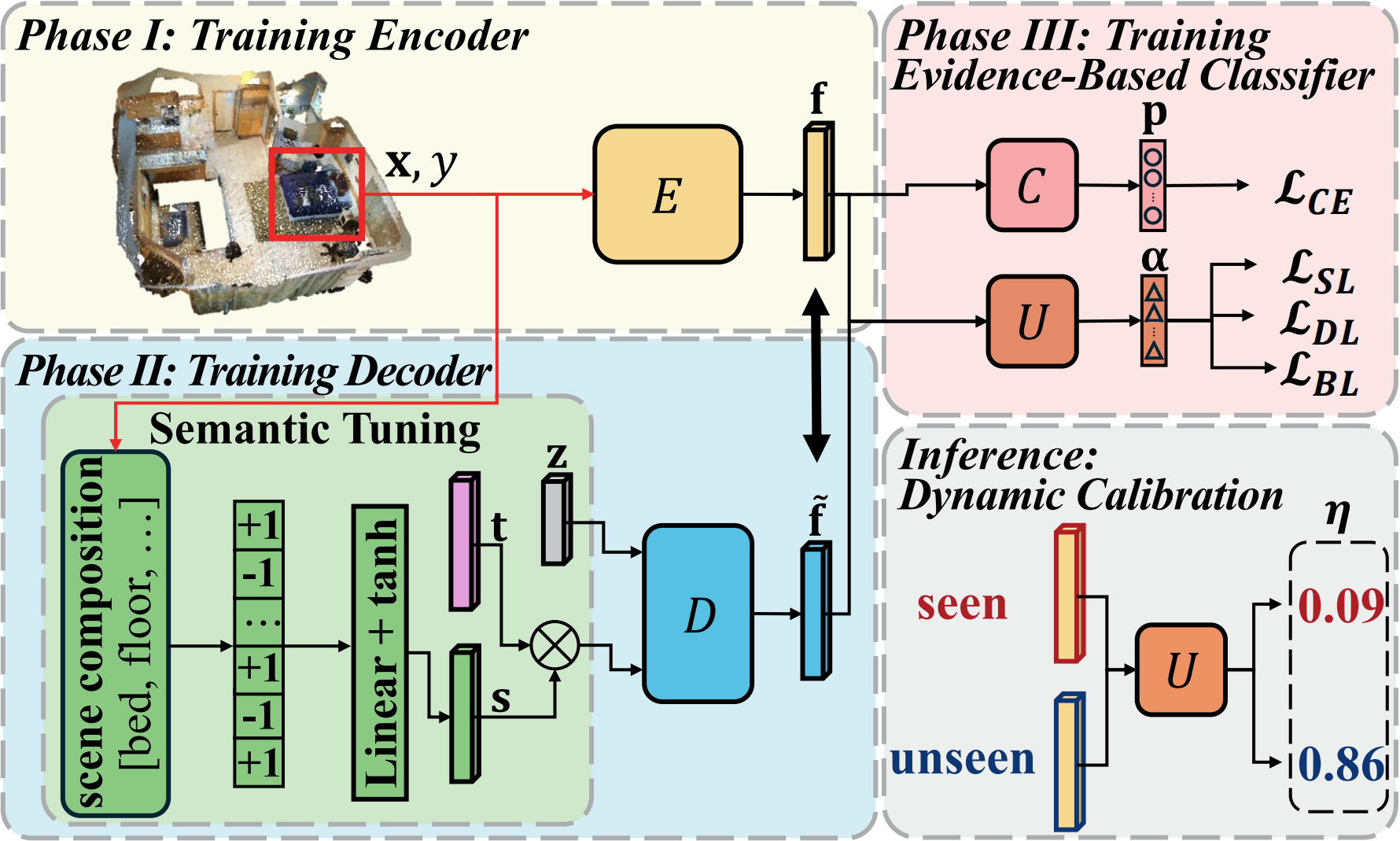}
    \caption{The E3DPC-GZSL architecture. The encoder $E$ extracts features from the point cloud; the decoder $D$ generates synthesized features by aligning visual features with a semantically tuned text space for training the classifier $C$. The classifier $C$ then predicts the class for each point by incorporating an uncertainty estimator to adjust the predicted probabilities.}
    \label{fig:overall_architecture}
\end{figure}
\section{Proposed Approach}
In this section, to overcome the aforementioned issues, we propose a novel method, E3DPC-GZSL, for generalized zero-shot semantic segmentation of 3D point clouds. The overall architecture of E3DPC-GZSL is shown in Figure \ref{fig:overall_architecture}. E3DPC-GZSL consists of three main components: 1) an encoder $E$ that extracts feature vectors $\mathbf{f}$ from input points $\mathbf{x}$, 2) a decoder $D$ that generates synthesized features, helping to train a classifier for segmentation and transfer knowledge from seen to unseen categories, and 3) a classifier $C$ with an uncertainty estimator $U$ that performs segmentation by using the feature vectors produced by $E$ and $D$. Note that during inference, only the feature encoder $E$, the classifier $C$, and the uncertainty estimator $U$ are used, while the decoder, which assisted the classifier in learning potential unseen features, is detached.

The training process of E3DPC-GZSL is divided into three phases. Below, we describe each phase and outline our strategies for achieving reliable predictions.

\subsection{Phase I: Training Encoder}
This training process aims to optimize the encoder $E$ to extract a distinctive feature vector $\mathbf{f}$ from an input $\mathbf{x}$ for category identification. The knowledge gained from $E$ is then used to train the decoder $D$ for feature vector synthesis. The feature vectors produced by $D$ are then used to learn the parameters for $C$ and $U$.

In this phase, the encoder $E$ is trained from scratch by optimizing the cross-entropy loss over seen samples of $\mathcal{D}_{tr}$. This approach is taken for two main reasons: first, there are no effective pre-trained foundation models for 3D point clouds; second, the configuration of 3D datasets varies significantly from task to task. For instance, in classification datasets~\cite{c:302, c:303}, each point cloud sample typically contains a single object without background, whereas segmentation and detection datasets~\cite{c:304, c:305} contain multiple objects with background in point cloud samples.
\subsection{Phase II: Training Decoder} 
This training process has two goals: 1) to overcome data scarcity by training the decoder $D$ to enable feature synthesis, and 2) to improve feature representation for synthesis by conditioning on both scene- and point-wise semantics.

Given a point $\mathbf{x}$ of the class $c_y$ from $\mathcal{D}_{tr}$, the pre-trained encoder $E$ extracts the feature vector $\mathbf{f}$. As described in Figure \ref{fig:overall_architecture},
the decoder $D$ is then trained to generate the feature vectors $\tilde{\mathbf{f}}$ from a vector $\mathbf{z} \in \mathbb{R}^{N_z}$. The vector $\mathbf{z}$ is randomly sampled from a uniform distribution in the range $[0, 1)$. To mitigate data scarcity issues in zero-shot learning, $D$ is conditioned on auxiliary prior knowledge about the class in the form of a text-driven feature embedding $\mathbf{t}$, which represents the class for feature synthesis. 

\noindent \textbf{Semantic Tuning. } In E3DPC-GZSL, the text embedding $\mathbf{t}$ is tuned by introducing a learnable feature vector derived from the scene semantics, denoted as $\mathbf{s}  \in \mathbb{R}^{N_t}$. This vector is constructed based on a scene composition descriptor in $\mathbb{R}^{N_c}$, whose elements are either $1$ or $-1$. A value of $1$ indicates the presence of a class in the scene, while a value of $-1$ indicates the absence of objects belonging to that class. 

To learn $\mathbf{s}$, the scene composition descriptor is passed through a fully connected layer with a hyperbolic tangent activation function. This introduction of $\mathbf{s}$ has two key advantages: it improves the quality of synthesized features by exploiting both global scene-based and local point-based conditions, and it acts similarly to adjusting prompts that include a word for a class name~(\eg [chair]) to prompts that include the class name with the scene description~(\eg [chair] near [table] in the room). 
Once $\mathbf{s}$ is obtained, $\mathbf{t}$ is tuned by adding point by point. The synthesized feature vector $\tilde{\mathbf{f}}$ is then produced from $\mathbf{z}$ and $\mathbf{t} \otimes \mathbf{s}$, \ie $\tilde{\mathbf{f}} = D(\mathbf{z}, \mathbf{t} \otimes \mathbf{s})$, where $\otimes$ denotes an point-wise multiplication operation. 

\noindent \textbf{Opimization. } The decoder $D$ is trained by following the approach in \cite{c:226}, minimizing the decoder loss. This decoder loss consists of the discrepancy loss between $\mathbf{f}$ and $\tilde{\mathbf{f}}$, the contrastive loss between positive and negative pairs, as well as the prototype distance loss across classes. Once the parameters of $D$ are learned, the synthesized features align with the semantic space derived from tunable text representations.
\subsection{Phase III: Training Evidence-Based Classifier for 3D 
Semantic Segmentation} 
The goal of this training process is twofold: first, to obtain the parameters of a classifier $C$ for 3D semantic segmentation, and second, to learn parameters of $U$ for evaluating prediction confidence. 

\noindent \textbf{Learning Segmentation. } As shown in Figure \ref{fig:overall_architecture}, 
to learn the parameters of $C$, the pre-trained encoder $E$ is used to encode the samples of seen classes, while the pre-trained decoder $D$ generates synthetic features for unseen classes. $C$ is then optimized using the cross-entropy loss on feature vectors produced by $E$ and $D$, following the standard framework outlined in (\ref{eq:class_posterior}) and (\ref{eq:cross_entropy}).
Here, the resulting class probabilities are represented as a vector $\boldsymbol{p} \in \mathbb{R}^{N_c}$, where $p_k$ denotes the probability of the class $c_k$.

\noindent \textbf{Uncertainty Estimation. } 
To avoid overconfident prediction due to prior knowledge of seen classes, at inference, the calibrated stacking method~\cite{c:237} is used to compute class probabilities, resulting in the vector $\boldsymbol{p}^{\prime} \in [0,1]^{N_c}$:
\begin{align} \label{eq:calibration}
     p^{\prime}_k = p_k - \eta \cdot
     \mathds{1}_{\scriptscriptstyle \mathcal{Y}^{s}}\left( c_k\right),
\end{align}
where $p_{k}^{\prime}$ represents the probability of the class $c_k$; $\mathds{1}_{\scriptscriptstyle A}(x)$ is denoted as $\mathds{1}\left(x \in A\right)$, \ie $\mathds{1}\left(c_k \in \mathcal{Y}^{s} \right)$; and the prediction probability is redistributed using a calibration factor $\eta$ in the range $[0,1]$. In (\ref{eq:calibration}), $\eta$ is used to decrease the probabilities for seen classes, which consequently increases the relative probabilities of unseen classes.
Once the class probabilities are computed, the predicted class is assigned as $\hat{y} = \underset{c_k}{\mathrm{argmax}}\:\:p_{k}^{\prime}$. 
However, segmentation performance can degrade when applying a pre-defined $\eta$ to (\ref{eq:calibration}), as demonstrated in Figure \ref{fig:motivation}(a). To mitigate this performance degradation, $\eta$ is treated as an adjustable parameter in E3DPC-GZSL. It is learned from uncertainty estimation derived based on evidence theory~\cite{b:111} and subjective logic~\cite{c:313}. 

In standard uncertainty estimation~\cite{c:231, r:402, c:233, c:314}, the Dirichlet distribution, characterized by a concentration parameter vector $\boldsymbol{\alpha}$, is used as a conjugate prior for Bayesian inference. When the Dirichlet distribution is uniform, it acts as a non-informative prior, representing complete uncertainty about the outcomes. For example, when minimal prior knowledge is assigned to all concentration parameters relevant to class probability~(\eg $\alpha_k = 1$ for all $k$ classes, where $\alpha \geq 1$), it implies no preference for any particular class (See Figure \ref{fig:concept}). This reflects maximum uncertainty and a lack of evidence, meaning there is no bias or skew toward the probability of any particular class.

This uncertainty~(or, conversely, evidence) can be modeled using evidence theory~\cite{b:111,c:231}, where the degree of uncertainty is inversely proportional to the total amount of evidence represented by $\alpha$ with the relationship $\alpha = evidence + 1$:
\begin{align} \label{eq:u}
u = \frac{K}{\alpha_0},
\end{align}
where $\alpha_0$ is defined as the sum of all concentration parameters, \ie $\alpha_0 = \sum_{k=1}^{K} \alpha_k$ and $K$ denotes the number of the parameter $\alpha$.

To derive $\eta$ based on prediction uncertainty, the uncertainty is measured exclusively using reliable sets of seen labels with $K=N_s$, considering two scenarios: 1) For unseen classes, $u$ ideally reaches its maximum by accumulating the evidence~($\alpha$) from seen classes. This occurs when all $\alpha$ are associated with seen classes, their class probabilities have minimal evidence. As a result, given unseen samples, high uncertainties associated with seen classes result in low class probabilities and a large $\eta$, thereby redistributing class probabilities more evenly for unseen classes. 2) For seen classes with labels, a skewed distribution towards a given class, driven by low uncertainty, increases prediction evidence with a small $\eta$.
Based on this reasoning, the process for estimating $\eta$ is as follows: model the evidence distribution as Dirichlet, estimate its concentration parameters $\alpha$~(substituting the evidence), compute the uncertainty $u$ from the estimated $\alpha$, and then estimate $\eta$ from $u$ (See Figure \ref{fig:concept}). 

\noindent \textbf{Learning $\boldsymbol{\eta}$. } 
The module $U$ for estimating $\alpha$ is optimized on the expected probability of evidence $\pi_k$ for each class, defined as $\pi_k = \alpha_k/\alpha_0$. 
Three objectives guide it: first, to estimate parameters that minimize Bayesian risk to improve evidence for class prediction~($\mathcal{L}_{SL}$); second, to regularize the optimization process ensuring balanced prediction for seen and unseen classes~($\mathcal{L}_{DL}$); and third, to improve evidence estimates by identifying unseen classes from seen ones~($\mathcal{L}_{BL}$):
\begin{align} \label{eq:overall_loss}
\mathcal{L}_{EV} = \mathcal{L}_{SL} + \lambda_{DL}\mathcal{L}_{DL} + \lambda_{BL}\mathcal{L}_{BL}, 
\end{align}
where $\lambda_{DL}$ and $\lambda_{BL}$ are the weighting coefficients.

The \textbf{segmentation loss~(SL)} is the posterior expected loss designed to minimize Bayesian risk using the cross-entropy loss function. Assuming a Dirichlet conjugate prior, the loss is formulated as follows~(for a detailed derivation, refer to Appendix A):
\begin{align} \label{loss_sl} \begin{array}{l}
    \mathcal{L}_{SL} \\
    =\!{\displaystyle \frac{1}{N_b^{\prime}}}\sum\limits_{j=1}^{N_b^{\prime}}{\displaystyle\int}\!\frac{1}{B(\boldsymbol{\alpha_j})}\prod\limits_{k^\prime=1}^{N_s}\pi_{j,k^\prime}^{\alpha_{j,k^\prime}-1}\!
    {\scriptstyle\left[ -\sum\limits_{k=1}^{N_s} \mathds{1}\left(c_k =c_{y_j} \right)\!\log \pi_{j,k}\right]\! }d\boldsymbol{\pi} \\ 
    = -{\displaystyle \frac{1}{N_b^{\prime}}}\sum\limits_{j=1}^{N_b^{\prime}}
    \sum\limits_{k=1}^{N_s} \mathds{1}\left(c_k = c_{y_j} \right) \big( \psi\left(\alpha_{j,k} \right) - \psi\left(\alpha_{j,0}\right) \big),
\end{array} \end{align}
where $B(\cdot)$ represents the multivariate beta function; $\Gamma(\cdot)$ denotes the gamma function, and $\psi(\cdot)$ is the digamma function, defined as $\psi(x) = \frac{d}{dx}\log \Gamma(x)$; and $N_b^{\prime}$ is the number of seen samples in a minibatch. 

The \textbf{divergence loss~(DL)} regularizes the model by reducing the Kullback-Leibler divergence from a uniform Dirichlet distribution to ensure reliable uncertainty prediction. A modified vector $\tilde{\alpha}$ is introduced to avoid misleading evidence when labeled samples are provided and to ensure high uncertainty for unseen ones, similar to \cite{c:231}:  
\begin{align} \begin{array}{l}
    \tilde{\boldsymbol{\alpha}} = \mathds{1}_{\scriptscriptstyle \mathcal{Y}^{s}}\left( c_{y}\right)\cdot \left\{ \boldsymbol{y} \otimes (\boldsymbol{1} + \frac{1}{\sqrt{\boldsymbol{\alpha}}}) + (\boldsymbol{1}-\boldsymbol{y}) \otimes \sqrt{\boldsymbol{\alpha}}\right\} \\      
    \hspace{1.2cm} + \mathds{1}_{\scriptscriptstyle \mathcal{Y}^{u}}\left( c_{y}\right)\cdot \boldsymbol{\alpha}, 
\end{array}\end{align}
where $\boldsymbol{1} \in \mathbb{R}^{N_s}$ is the vector of ones, $\otimes$ denotes element-wise multiplication, and $\boldsymbol{y}$ is provided in a one-hot encoded format.
Using $\tilde{\boldsymbol{\alpha}}$, the module estimates the evidence for each class by preventing incorrect predictions for unseen samples from being biased towards any seen classes~(see Appendix A for full derivation):
\begin{align} \label{loss_dl} \begin{array}{ll}
    \mathcal{L}_{DL} \hspace{-0.3cm} &= {\displaystyle \frac{1}{N_b}}\sum\limits_{j=1}^{N_b} \textrm{KL} \left[ Dir\left( \boldsymbol{\pi}_j|\tilde{\boldsymbol{\alpha}_j}\right) \bigg|\bigg| \: Dir\left( \boldsymbol{\pi}_j|\boldsymbol{1}\right)\right]\\ 
    &= {\displaystyle \frac{1}{N_b}}\sum\limits_{j=1}^{N_b} \big(\log {\displaystyle \frac{1}{\Gamma(N_s)B(\widetilde{\boldsymbol{\alpha}}_j)} } \\
    & \hspace{1.5cm} + \sum\limits_{k=1}^{N_s} (\tilde{\alpha}_{j,k}-1)\big(\psi(\tilde{\alpha}_{j,k}) - \psi(\alpha_{j,0})\big) \big). 
\end{array}\end{align}

In addition to the loss functions, the \textbf{binary loss~(BL)} is introduced to ensure that the evidence is not biased toward any class when handling unseen samples. The loss function reduces uncertainty for seen samples and simultaneously increases uncertainty for unseen samples, encouraging balanced predictions:
\begin{align} \begin{array}{ll} \label{eq:binary_loss} 
    \mathcal{L}_{BL} \hspace{-0.3cm} &=
    - {\displaystyle \frac{1}{N_b}}\sum\limits_{j=1}^{N_b} \big[ \mathds{1}_{\scriptscriptstyle \mathcal{Y}^{s}}\left( c_{y_j}\right)\log u_{j} \big. \\
    & \hspace{2.5cm} \big. + \mathds{1}_{\scriptscriptstyle \mathcal{Y}^{u}}\left( c_{y_j}\right) \log (1-u_{j}) \big]. 
\end{array}\end{align}
In (\ref{eq:binary_loss}), by reducing uncertainty for seen samples, the module gains confidence in its predictions for seen classes, leading to more decisive and accurate results. In contrast, by increasing uncertainty for unseen samples, the module avoids making overly confident predictions about unseen samples. This increase in uncertainty prevents overfitting and ensures that the module recognizes its limitations when it encounters new inputs.

\noindent \textbf{Inference. } Using the learned parameters of $U$ to estimate $\alpha$, the dynamic calibration factor is defined as $\eta = u - \bar{u}$, where $\bar{u}$ is defined as the average estimated uncertainty of unseen samples predicted from $C$ before applying calibrated stacking.

\begin{table*}[t]
\centering
  \setlength{\tabcolsep}{1.0mm}
  \begin{tabular}{c|c|c|ccc|c|ccc|c}
    \toprule
     & \multicolumn{2}{c|}{Training set} & \multicolumn{4}{c|}{ScanNet v2} & \multicolumn{4}{c}{S3DIS} \\ \cline{2-11}
     & \multirow{2}{*}{Encoder} & \multirow{2}{*}{Classifier} & \multicolumn{3}{c|}{mIoU} & \multirow{2}{*}{HmIoU} & \multicolumn{3}{c|}{mIoU} & \multirow{2}{*}{HmIoU} \\ \cline{4-6} \cline{8-10}
     & & & Seen & Unseen & All & & Seen & Unseen & All & \\
     \midrule
     Full supervision & $\mathcal{Y}^{s} \cup \mathcal{Y}^{u}$ & $\mathcal{Y}^{s} \cup \mathcal{Y}^{u}$ & 43.3 & 51.9 & 45.1 & 47.2 & 74.0 & 50.0  & 66.6 & 59.6 \\
     Full supervision only for classifier & $\mathcal{Y}^{s}$ & $\mathcal{Y}^{s} \cup \mathcal{Y}^{u}$ & 41.5 & 39.2 & 40.3 & 40.3 & 60.9 & 21.5 & 48.7 & 31.8 \\
     Supervision with seen & $\mathcal{Y}^{s}$ & $\mathcal{Y}^{s}$ & 39.0 & 0.0 & 31.3 & 0.0 & 70.2 &  0.0 & 48.6 & 0.0 \\
     \midrule
     3DGenZ \cite{c:225} & $\mathcal{Y}^{s}$ & $\mathcal{Y}^{s} \cup \mathcal{Y}^{\tilde{u}}$ & 32.8 & 7.7 & 27.8 & 12.5 & 53.1 & 7.3 & 39.0 & 12.9 \\
     3DPC-GZSL \cite{c:226} & $\mathcal{Y}^{s}$ & $\mathcal{Y}^{s} \cup \mathcal{Y}^{\tilde{u}}$ & 34.5 & 14.3 & 30.4 & 20.2 & 58.9 & 9.7 & 43.8 & 16.7\\
     \midrule
     \textbf{E3DPC-GZSL (ours)} & $\mathcal{Y}^{s}$ & $\mathcal{Y}^{s} \cup \mathcal{Y}^{\tilde{u}}$ & \textbf{36.1} & \textbf{15.4} & \textbf{32.0} & \textbf{21.6} & \textbf{67.9} & \textbf{12.0} & \textbf{50.7} & \textbf{20.4}\\
    \bottomrule
\end{tabular} 
\caption{Performance Comparisons of 3D GZSL semantic segmentation benchmarks in terms of mIoU(\%) and HmIoU(\%).}
\label{tab:GZSL_semantic_seg_benchmarks}
\end{table*}
\section{Experiments}
In this section, we provide details on various experiments designed to evaluate E3DPC-GZSL. Note that additional results and analyses are provided in the supplementary material.

\subsection{Experimental Setup}
\noindent \textbf{Datasets.}
We evaluate the proposed E3DPC-GZSL method using the S3DIS dataset~\cite{c:305} and the ScanNet v2 dataset~\cite{c:304}, following the data splitting protocol outlined in previous studies~\cite{c:225, c:226}.

The ScanNet v2 dataset, collected indoors, consists of $1,201$ point cloud scenes for training and $312$ point cloud scenes for evaluation. The training scenes encompass sixteen seen classes~($N_s=16$), while the evaluation scenes include four unseen classes (desk, bookshelf, sofa, and toilet, $N_u = 4$) and sixteen seen classes.

The S3DIS dataset consists of $272$ scenes from $6$ indoor areas, covering $13$ classes. Areas 2, 3, 4, 5, and 6~($228$ scenes) are used for training with nine seen classes~($N_s=9$), while Area $1$ with $44$ scenes is designated as the evaluation dataset. This evaluation set includes four unseen classes~(beam, column, window, and sofa, $N_u=4$) and nine seen classes.

\noindent \textbf{Evaluation Metrics.} The performance of E3DPC-GZSL is assessed using two metrics. The first metric is the mean Intersection-over-Union~(mIoU), which measures the average overlap between the predicted and ground truth segmentations across all classes. We assess three types of mIoU: for seen classes, unseen classes, and a combined measure for both. The second metric used is the Harmonic mean of mIoU~(HmIoU), defined as $2 \times \left( \textrm{mIoU}(\mathcal{Y}^{s}) \times \textrm{mIoU}(\mathcal{Y}^{u})\right) / \left( \textrm{mIoU}(\mathcal{Y}^{s}) + \textrm{mIoU}(\mathcal{Y}^{u}) \right)$.

\noindent \textbf{Implementation Details. } We implement the proposed method using the PyTorch framework. 
Following the 3D GZSL semantic segmentation setting~\cite{c:225}, the sample points contain spatial coordinates for the ScanNet v2~($N_p=3$) and spatial coordinates plus color information for the S3DIS~($N_p=6$). The experiments use the FKAConv~\cite{c:306} network on the ScanNet v2 and the ConvPoint~\cite{r:301} network on the S3DIS as the encoder and classifier. The encoder $E$ encodes the inputs into $64$-dimensional feature vectors~($N_f=64$) for ScanNet v2 and $128$ for S3DIS. The decoder $D$ used is the generative moment matching network~(GMMN)~\cite{c:312}. We employ $600$-dimensional~($N_t=600$) text embeddings, which are a concatenation of  GloVe~\cite{C:222} and Word2Vec~\cite{c:221}, as class description vectors.
The proposed model is trained with a learning rate of $7e-2$, a batch size of 4, and a poly learning rate scheduler with a base of $0.9$ and $30$ epochs. The Adam optimizer is used for ScanNet v2, while the SGD optimizer is used for S3DIS. Each minibatch contains $8192$ sample points~($N_b=8192$). $\lambda_{DL}$ and $\lambda_{BL}$ are set to $0.005$ and $0.01$ for ScanNet v2 and $0.005$ and $0.1$ for S3DIS, respectively. The uncertainty estimator utilizes the same network as the classifier, employing the exponential function as its activation function.
\subsection{Experimental Results}

The effectiveness of our proposed methods is compared with SOTA methods in the inductive generalized zero-shot learning setting, where the encoder is trained with seen data~($\mathcal{Y}^s$) and the classifier is trained with both seen and augmented unseen data~$\mathcal{Y}^{\tilde{u}}$. The methods compared are 3DGenZ~\cite{c:225} and 3DPC-GZSL~\cite{c:226}. The results of this comparison are summarized in Table \ref{tab:GZSL_semantic_seg_benchmarks}. Additionally, for reference, we provide performance for three different supervised settings using the same encoder and classifier as the zero-shot models.
Our proposed method achieves improvements over the state-of-the-art method in both seen and unseen mIoU metrics. Specifically, it demonstrates an increase of 1.4\% HmIoU on the ScanNet v2 and 3.7\% HmIoU on the S3DIS.
Note that qualitative results and additional evaluations on outdoor benchmarks are provided in the supplementary material.
\begin{table}[t]
  \centering
  \setlength{\tabcolsep}{1.5mm}
  \begin{tabular}{@{}c@{}|ccc|ccc|c}
    \toprule
    \multirow{2}{*}{Dataset} & \multirow{2}{*}{B} & \multirow{2}{*}{S} & \multirow{2}{*}{U} & \multicolumn{3}{c|}{mIoU} & \multirow{2}{*}{HmIoU} \\
    \cline{5-7}
    & & & & Seen & Unseen &All & \\
    \midrule
    \multirow{4}{*}{ScanNet v2} & \checkmark & -- &  -- & 34.78 & 14.80 & 30.79 & 20.77\\
     & \checkmark & -- & \checkmark & 34.86 & 14.87 & 30.87 & 20.85 \\
     & \checkmark & \checkmark & -- & 36.09 & 15.23 & 31.91 & 21.42 \\
     & \checkmark & \checkmark & \checkmark & \textbf{36.11} & \textbf{15.40} & \textbf{31.97} & \textbf{21.59} \\
    \midrule
    \multirow{4}{*}{S3DIS} & \checkmark & -- &  -- & 65.03 & 10.17 & 48.15 & 17.60\\
     & \checkmark & -- & \checkmark & 66.58 & 11.27 & 49.56 & 19.28 \\
     & \checkmark & \checkmark & -- & 66.46 & 11.02 & 49.40 & 18.90 \\
     & \checkmark & \checkmark & \checkmark & \textbf{67.90} & \textbf{12.01} & \textbf{50.70} & \textbf{20.42} \\
    \bottomrule
\end{tabular} 
\caption{Analysis of the effects of each module on segmentation performance.}
\label{tab:Module_component_analysis}
\end{table} 
\subsection{Further Discussion}
\noindent \textbf{Analysis of Component Effectiveness. }
The effectiveness of the proposed semantic tuning and the uncertainty estimator is validated by removing and adding each component. The analysis results are summarized in Table~\ref{tab:Module_component_analysis}. In the table, ``B'' represents the baseline consisting of the encoder without calibrated stacking~($\eta=0$), decoder without semantic tuning, and standard classifier; ``S'' represents the addition of the semantically tuned decoder; and ``U'' represents the addition of the dynamically calibrated classifier. It is evident that dynamic calibration using module $U$ effectively adjusts the module's predictions independent of semantic tuning. Moreover, when semantic tuning is applied, it enhances classifier performance by increasing the expressive power of the decoder.

\begin{table}[t]
  \centering
  \setlength{\tabcolsep}{0.7mm}
  \begin{tabular}{@{}c@{}|ccc|ccc|c}
    \toprule
    \multirow{2}{*}{Dataset} & \multirow{2}{*}{$\mathcal{L}_{SL}$} & \multirow{2}{*}{ $\mathcal{L}_{DL}$} & \multirow{2}{*}{$\mathcal{L}_{BL}$} & \multicolumn{3}{c|}{mIoU} & \multirow{2}{*}{HmIoU} \\
    \cline{5-7}
    & & & & Seen & unseen &All & \\
    \midrule
    \multirow{5}{*}{ScanNet v2} & -- & -- & -- & 36.09 & 15.23 & 31.91 & 21.42 \\    
    & \checkmark & -- & -- & \textbf{36.14} & 15.31 & \textbf{31.98} & 21.51\\
    & \checkmark & \checkmark & -- & 36.13 & 15.33 & 31.97 & 21.53 \\
    & \checkmark & -- & \checkmark & \textbf{36.14} & 15.32 & \textbf{31.98} & 21.51 \\
    & \checkmark & \checkmark & \checkmark & 36.11 & \textbf{15.40} & 31.97 & \textbf{21.59} \\
    \midrule
    \multirow{5}{*}{S3DIS} & -- & -- & -- & 66.46 & 11.02 & 49.40 & 18.90 \\ 
    & \checkmark & -- & -- & 66.64 &  11.19 & 49.58 & 19.17 \\
    & \checkmark & \checkmark & -- & 67.19 & 11.66 & 50.10 & 19.89 \\
    & \checkmark & -- & \checkmark & 66.84 & 11.39 & 49.78 & 19.46 \\
    & \checkmark & \checkmark & \checkmark & \textbf{67.90} & \textbf{12.01} & \textbf{50.70} & \textbf{20.42} \\
    \bottomrule
\end{tabular}
\caption{Ablation study on loss functions for uncertainty estimation.}
\label{tab:Dirichlet_loss}
\end{table}

\noindent \textbf{Analysis of Uncertainty Estimator.} In addition, to further analyze the proposed uncertainty estimation module, we evaluate the effectiveness of each loss function. The analysis results are presented in Table \ref{tab:Dirichlet_loss}. For reference, the performance of the B+S model in Table~\ref{tab:Module_component_analysis} is reported in the first row for each dataset as the baseline. As expected, segmentation accuracy on unseen data improves by redistributing the model's prediction probabilities, which coincides with HmIoU improvements, reflecting enhanced performance on unseen data.

\begin{figure}[t]
    \centering
    \begin{tabular}{@{}c@{}c@{}}
         {ScanNet v2} & { S3DIS}\\
        \includegraphics[width=0.485\linewidth]{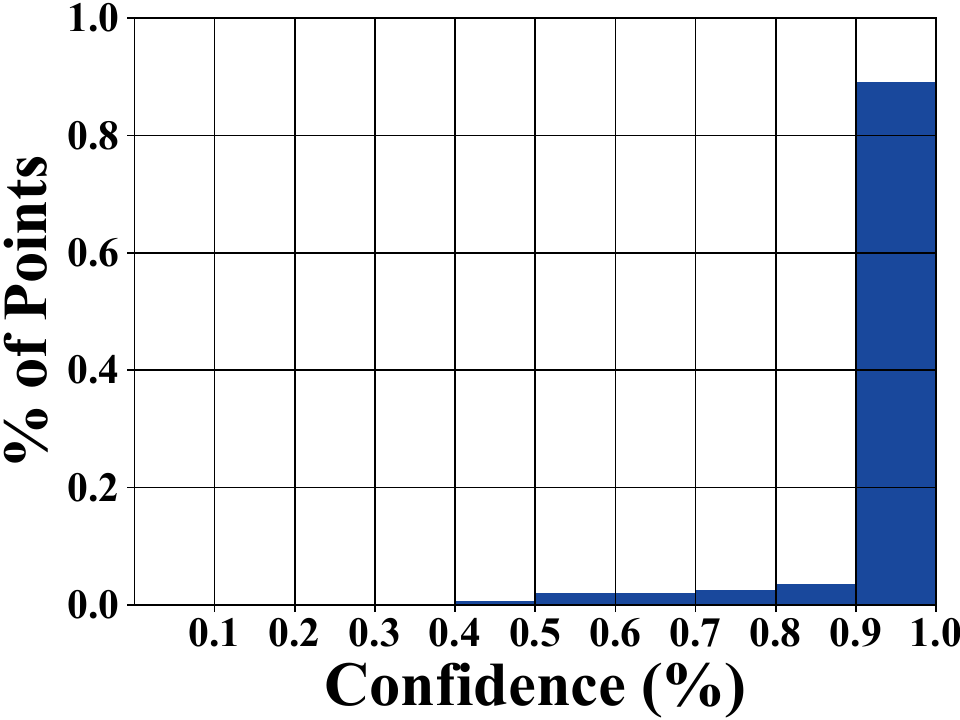}  \hspace{0.2cm}& \includegraphics[width=0.485\linewidth]{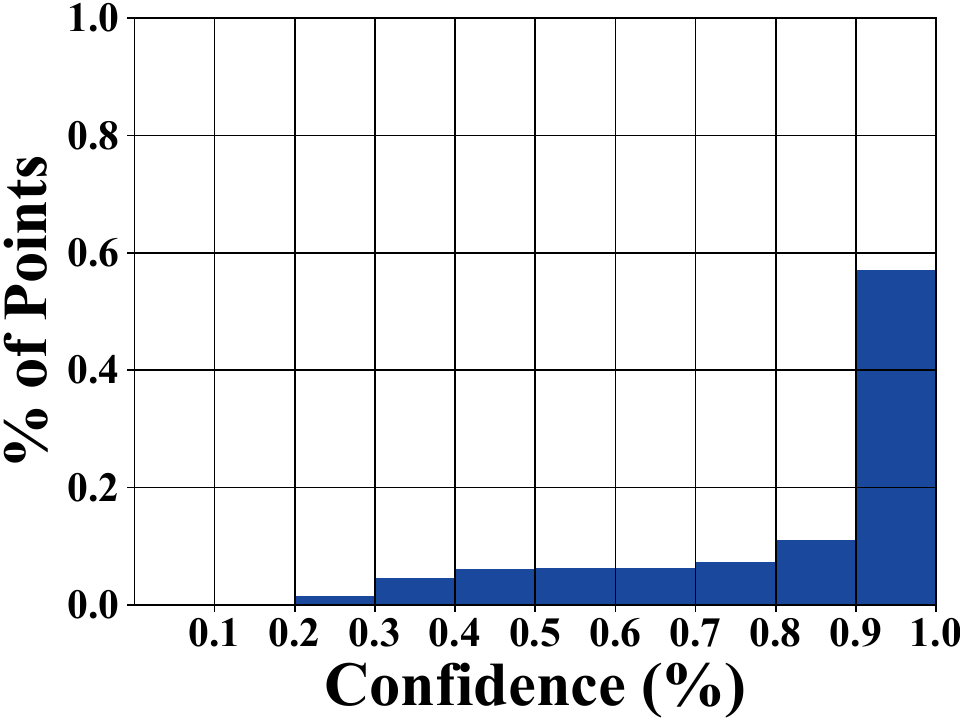} \\
        \includegraphics[width=0.485\linewidth]{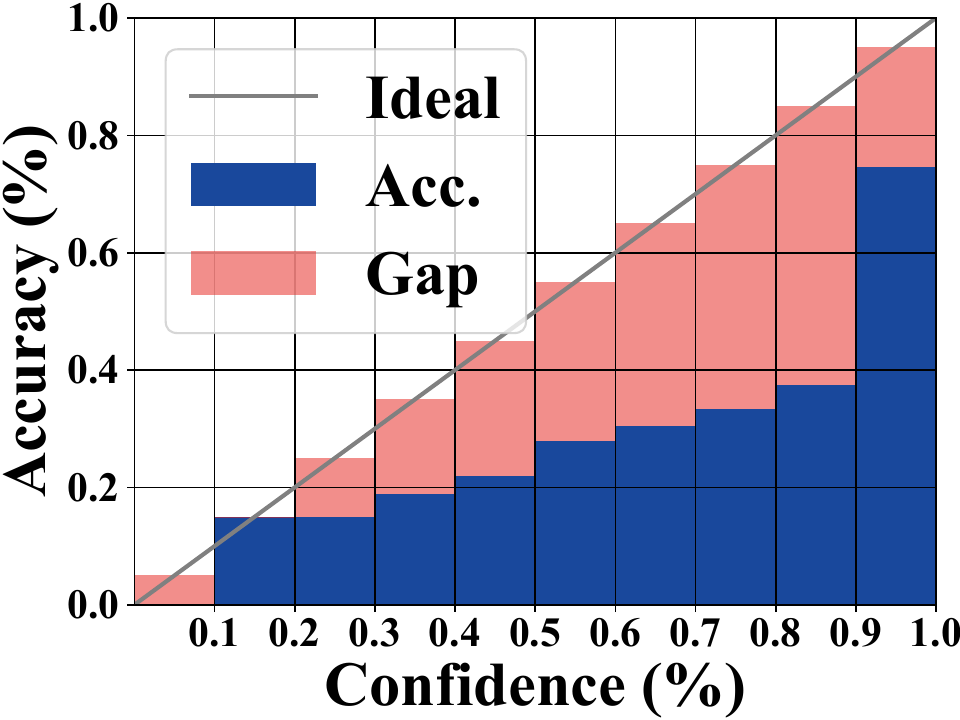}  \hspace{0.01cm} 
        & \includegraphics[width=0.485\linewidth]{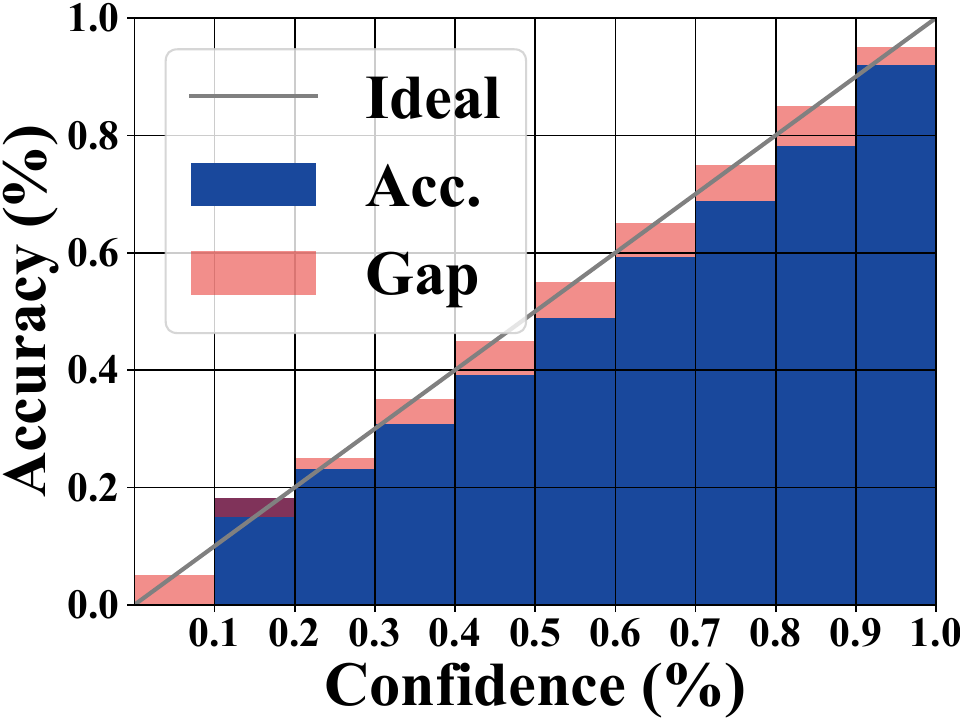} \\
    \end{tabular}
    \caption{Analysis of model confidence on ScanNet v2 and S3DIS: confidence histograms~(top) and reliability diagrams~(bottom). The Gap represents the difference between the ideal accuracy and the measured accuracy.} \label{fig:model_confidence}
\end{figure}

For the S3DIS dataset, the results clearly demonstrate the effectiveness of the uncertainty estimator for unseen data. However, its impact on ScanNet v2 is less pronounced, showing only marginal improvement. To explore this further, we analyze and measure the confidence histograms and reliability diagrams of the models without applying calibrated stacking. The confidence histograms in the top row of Figure \ref{fig:model_confidence} show the distribution of prediction confidence, represented as softmax probabilities of the classifier $C$ associated with the predicted class label. The reliability diagrams in the bottom row of Figure \ref{fig:model_confidence} show accuracy as a function of confidence, calculating the accuracy of each bin in the confidence histograms~(\ie the ratio of correct predictions of each bin). If the model produces balanced outputs, the diagram would show the identity function labeled ``ideal'' in the graph. Any deviation from this identity function indicates that the model produces overconfident outputs, with one class of probability significantly higher than the others~\cite{c:310}. 
As evidenced in the confidence histogram for ScanNet v2, the model tends to produce outputs with a significantly high probability of one class. This overconfidence limits the effectiveness of calibrated stacking, which is intended to reduce the bias toward the classes seen. Since the probabilities are already very high, subtracting $\eta$ does not effectively shift the peak of the probability distribution. In contrast, the model confidence in the outputs for S3DIS is relatively lower compared to ScanNet v2. Consequently, the effect of calibrated stacking is much more pronounced for S3DIS, resulting in a more effective redistribution of the probability distribution. 

\section{Conclusion}
We proposed E3DPC-GZSL, a novel approach for generalized zero-shot point cloud semantic segmentation. Our method exploits the uncertainty of input points to dynamically calibrate classifier predictions. This uncertainty-based strategy helps to mitigate the bias of zero-shot models towards seen classes, thereby improving the generalization performance of the model. Additionally, to address the issue of data scarcity, we introduced a novel training strategy that refines the semantic space by applying semantic tuning to text embeddings. Our experiments show that E3DPC-GZSL outperforms SOTA methods in 3D semantic segmentation. Despite the significant performance improvements over SOTA, the impact is less pronounced on models that tend to produce overconfident results with high probabilities. Regularizing the model's overconfidence with biased predictions in a zero-shot setting could improve performance in such cases. We leave this as a direction for future research.
\section{Acknowledgments}
This work was supported in part by the Institute of Information and Communications Technology Planning and Evaluation~(IITP) grant funded by the Korean government~(MSIT), 3D Digital Media Streaming Service Technology, under Grant No. RS-2023-00229330; and in part by the National Research Foundation of Korea~(NRF) grant funded by the Korea government~(MSIT)~(No. 2022R1F1A1062950).
\section*{Note}
This arXiv version includes a correction to Equation (2). The logarithm function was inadvertently omitted in the original cross-entropy loss formula. This has been fixed in this version for clarity and correctness. The originally published version in AAAI contains this minor typographical error. Additionally, this version includes an supplementary material. The code is available at \url{https://github.com/Hsgalaxy-Kim/E3DPC-GZSL}.


%

\bibliography{aaai25}

{
\appendix
\onecolumn

\begin{center}
\bf \LARGE Supplementary Material: Generalized Zero-Shot Learning for Point Cloud Segmentation with Evidence-based Dynamic Calibration \\ 
\end{center}
\vspace{2cm}

\noindent \textbf{\Large A. Derivation of Loss Functions} 
\vspace{0.5cm}

\noindent This appendix section derives the segmentation loss~($\mathcal{L}_{SL}$) of (\ref{loss_sl}) and divergence loss~($\mathcal{L}_{DL}$) of (\ref{loss_dl}) functions. \\

\noindent The following is derived under the assumption that $\pi_k \geq 0 \:\: \forall k \in N_s $: \vspace{0.1cm}
\begin{align} \begin{array}{ll}
    \mathcal{L}_{SL} \hspace{-0.3cm} &= {\displaystyle \frac{1}{N_b^{\prime}}\sum\limits_{j=1}^{N_b^{\prime}} \int \frac{1}{B(\boldsymbol{\alpha_j})} \prod\limits_{k^\prime=1}^{N_s}\pi_{j,k^\prime}^{\alpha_{j,k^\prime}-1} \left[ -\sum\limits_{k=1}^{N_s} \mathds{1}\left(c_k = c_{y_j} \right)\log \pi_{j,k}\right] d\boldsymbol{\pi} }\vspace{0.2cm}\\ 
    & \displaystyle = -\frac{1}{N_b^{\prime}}\sum\limits_{j=1}^{N_b^{\prime}}\sum\limits_{k=1}^{N_s} \mathds{1}\left(c_k = c_{y_j} \right) \left( \frac{\Gamma(\alpha_{j,0})}{\prod_{k^\prime=1}^{N_s} \Gamma(\alpha_{j, k^\prime})} 
    {\displaystyle\int} \log \pi_{j,k} \prod\limits_{k^\prime = 1}^{N_s} \pi_{j, k^\prime}^{\alpha_{j, k^\prime}-1} d\boldsymbol{\pi}\right) \vspace{0.2cm}\\
    & \displaystyle =  -\frac{1}{N_b^{\prime}}\sum\limits_{j=1}^{N_b^{\prime}}\sum\limits_{k=1}^{N_s} \mathds{1}\left(c_k = c_{y_j} \right) \: \mathbb{E}\big[ \log \pi_{j, k}\big] \vspace{0.2cm} \\
    & \displaystyle = -\frac{1}{N_b^{\prime}}\sum\limits_{j=1}^{N_b^{\prime}}\sum\limits_{k=1}^{N_s} \mathds{1}\left(c_k = c_{y_j} \right) \big( \psi\left(\alpha_{j,k} \right) - \psi\left(\alpha_{j,0}\right) \big), \\
\end{array}\end{align}
where $\mathbb{E}\big[ \log \pi_{k}\big]$ quantifies as $\psi\left(\alpha_k \right) - \psi\left(\alpha_0\right)$. \\

\noindent The Kullback-Leibler (KL) Divergence between the estimated prediction probability and the uniform Dirichlet distribution is derived as follows:
\begin{align}\begin{array}{ll}
    \mathcal{L}_{DL} \hspace{-0.3cm} &= 
    \displaystyle \frac{1}{N_b}\sum\limits_{j=1}^{N_b} \textrm{KL} \left[ Dir\left( \boldsymbol{\pi}_j|\tilde{\boldsymbol{\alpha}_j}\right) \bigg|\bigg| \: Dir\left( \boldsymbol{\pi}_j|\boldsymbol{1}\right)\right] \vspace{0.2cm}\\
    & \displaystyle = \: \frac{1}{N_b}\sum\limits_{j=1}^{N_b} \mathbb{E}\big[ \log \frac{p\left(\boldsymbol{\pi_{j}}|\widetilde{\boldsymbol{\alpha}}_{j}\right)}{p\left(\boldsymbol{\pi_{j}}|\boldsymbol{1}\right)}\big] \vspace{0.2cm} \\
    & \displaystyle = \: \frac{1}{N_b}\sum\limits_{j=1}^{N_b} \big( \mathbb{E}\big[ \log p\left(\boldsymbol{\pi_{j}}|\widetilde{\boldsymbol{\alpha}}_{j}\right)\big] - \mathbb{E}\big[ \log p\left(\boldsymbol{\pi_{j}}|\boldsymbol{1}\right)\big] \vspace{0.2cm} \big)\\
    & \displaystyle = \frac{1}{N_b}\sum\limits_{j=1}^{N_b} \big( \mathbb{E}\big[ -\log B (\widetilde{\boldsymbol{\alpha}}_j) + \sum_{k=1}^{N_{s}} (\tilde{\alpha}_{j,k} - 1)\log \pi_{j, k} \big] { \displaystyle -  \mathbb{E}\big[ -\log B \left( \boldsymbol{1} \right) \big]} \vspace{0.2cm} \big)\\
    & \displaystyle = \frac{1}{N_b}\sum\limits_{j=1}^{N_b} \big( \log \frac{B(\boldsymbol{1})}{B(\widetilde{\boldsymbol{\alpha}}_j)} \:+\:   \mathbb{E}\big[ \sum_{k=1}^{N_{s}} (\tilde{\alpha}_{j, k} - 1)\log \pi_{j, k} \big] \big) \vspace{0.2cm}\\
    & \displaystyle = \frac{1}{N_b}\sum\limits_{j=1}^{N_b} \big( \log \frac{1}{\Gamma(N_{c})B(\widetilde{\boldsymbol{\alpha}}_j)} + \sum_{k=1}^{N_{s}} (\tilde{\alpha}_{j, k} - 1)\mathbb{E}\big[\log \pi_{j, k} \big] \big) \vspace{0.2cm}\\
     & \displaystyle = \frac{1}{N_b}\sum\limits_{j=1}^{N_b} \big( \log \frac{1}{\Gamma(N_c)B(\widetilde{\boldsymbol{\alpha}}_j)} {\displaystyle + \sum\limits_{k=1}^{N_s} (\tilde{\alpha}_{j,k}-1)\big(\psi(\tilde{\alpha}_{j,k}) - \psi(\alpha_{j,0})\big)} \big). \vspace{1cm} \\ 
\end{array}\end{align}
\noindent \textbf{\Large B. Additional Results}
\vspace{0.5cm}

\noindent \textbf{Per-Class Semantic Segmentation Results.}
In addition to the mIoU and HmIoU results presented in Table~\ref{tab:GZSL_semantic_seg_benchmarks} of the main manuscript, we provide class-wise segmentation accuracies in Table~\ref{tab:ScanNet_classwise} for ScanNet v2~\cite{c:304} and Table~\ref{tab:S3DIS_classwise} for S3DIS~\cite{c:305}. For comparison, we also include results from fully supervised models at varying levels of supervision. Furthermore, we report detailed comparisons against existing zero-shot segmentation methods, including 3DGenZ~\cite{c:225} and 3DPC-GZSL~\cite{c:226}, to highlight the strengths of our approach in both seen and unseen categories.

On ScanNet (Table~\ref{tab:ScanNet_classwise}), our method achieves a notable improvement of +9.1\% IoU in the ``door'' class over 3DPC-GZSL and +3.3\% IoU over 3DGenZ. Compared to 3DPC-GZSL, additional gains are observed in the ``picture'' (+3.1\% IoU) and ``window'' (+7.1\% IoU) classes, while relative to 3DGenZ, our method achieves -0.3\% IoU in ``picture'' but a substantial +8.0\% IoU improvement in ``window''. Among unseen categories, the ``toilet'' class shows an IoU increase of +3.1\% over 3DPC-GZSL and +7.6\% over 3DGenZ, underscoring our method's ability to generalize to novel categories.

On S3DIS (Table~\ref{tab:S3DIS_classwise}), our method significantly outperforms prior works. Compared to 3DPC-GZSL, it yields improvements of +28.6\%, +14.0\%, and +10.1\% IoU in the ``door'', ``chair'', and ``table'' classes, respectively. When compared to 3DGenZ, these gains are even more pronounced, with increases of +43.6\%, +14.9\%, and +7.0\% IoU, respectively. For unseen categories, our method improves the ``sofa'' class by +11.4\% IoU over 3DPC-GZSL and +12.7\% IoU over 3DGenZ. These results highlight the robustness of our approach, particularly in challenging indoor scenes. However, performance on the column class remains limited, likely due to its strong visual similarity to the wall class, making accurate differentiation more difficult.

\begin{table*}[!h]
\centering
  \setlength{\tabcolsep}{1.5pt}
  \begin{tabular}{l|c|cccccccccccccccc|cccc}
    \toprule
    & & \multicolumn{16}{c|}{Seen classes} & \multicolumn{4}{c}{Unseen classes} \\
    \cline{3-22}
    ScanNet v2& HmIoU& \rotatebox[origin=c]{90}{bath} & \rotatebox[origin=c]{90}{bed} & \rotatebox[origin=c]{90}{cabinet} & \rotatebox[origin=c]{90}{chair} & \rotatebox[origin=c]{90}{counter} & \rotatebox[origin=c]{90}{curtain} & \rotatebox[origin=c]{90}{door} & \rotatebox[origin=c]{90}{floor} & \rotatebox[origin=c]{90}{other} & \rotatebox[origin=c]{90}{picture} & \rotatebox[origin=c]{90}{fridge} & \rotatebox[origin=c]{90}{s. cur.} & \rotatebox[origin=c]{90}{sink} & \rotatebox[origin=c]{90}{table} & \rotatebox[origin=c]{90}{wall} & \rotatebox[origin=c]{90}{window} & \rotatebox[origin=c]{90}{bkshf} & \rotatebox[origin=c]{90}{desk} & \rotatebox[origin=c]{90}{sofa} & \rotatebox[origin=c]{90}{toilet}\\
    \midrule
    Full-Sup & 47.2 & 58.0 & 67.5 & 21.2 & 75.5 & 12.0 & 35.2 & 13.6 & 96.5 & 20.6 & 10.7 & 39.9 & 63.3 & 34.2 & 59.5 & 81.1 & 4.8 & 56.9 & 30.0 & 57.4 & 63.4\\
    \midrule
    3DGenZ & 12.5 & 64.9 & 44.0 & 16.9 & 63.2 & 15.3 & 33.8 & 10.4 & 91.0 & 10.1 & \textbf{4.3} & 26.1 & 0.2 & 27.5 & 43.1 & 71.3 & 2.8 & 6.3 & 3.3 & 13.1 & 8.1\\
    3DPC-GZSL & 20.2 & 61.0 & \textbf{46.2} & \textbf{18.6} & \textbf{63.3} & 14.2 & 31.1 & 4.6 & \textbf{90.7} & 11.2 & 0.9 & 27.2 & \textbf{30.9} & \textbf{29.0} & \textbf{46.5} & 72.4 & 3.7 & 11.1 & 9.9 & \textbf{23.6} & 12.6\\
    \midrule
    \textbf{Ours} & \textbf{21.6} & \textbf{62.3} & 44.9 & 16.1 & 63.1 & \textbf{16.2} & \textbf{37.9} & \textbf{13.7} & \textbf{90.7} & \textbf{11.9} & 4.0 & \textbf{29.5} & \textbf{30.9} & 26.7 & 46.2 & \textbf{72.9} & \textbf{10.8} & \textbf{12.5} & \textbf{10.0} & 23.3 & \textbf{15.7}\\
  \bottomrule
\end{tabular} 
\caption{Class-wise performance comparisons of 3D GZSL Semantic Segmentation benchmarks in terms of mIoU(\%) and HmIoU(\%) on ScanNet v2 dataset.}
\label{tab:ScanNet_classwise} 
\end{table*}
\begin{table*}[!h]
\centering
  \setlength{\tabcolsep}{2.2mm}
  \begin{tabular}{l|c|ccccccccc|cccc}
    \toprule
    & & \multicolumn{9}{c|}{Seen classes} & \multicolumn{4}{c}{Unseen classes} \\
    \cline{3-15}
    S3DIS& HmIoU& \rotatebox[origin=c]{90}{board} & \rotatebox[origin=c]{90}{bookcase} & \rotatebox[origin=c]{90}{ceiling} & \rotatebox[origin=c]{90}{chair} & \rotatebox[origin=c]{90}{clutter} & \rotatebox[origin=c]{90}{door} & \rotatebox[origin=c]{90}{floor} & \rotatebox[origin=c]{90}{table} & \rotatebox[origin=c]{90}{wall} & \rotatebox[origin=c]{90}{beam} & \rotatebox[origin=c]{90}{column} & \rotatebox[origin=c]{90}{sofa} & \rotatebox[origin=c]{90}{window}\\
    \midrule
    Full-Sup & 59.6 & 53.9 & 54.4 & 96.5 & 75.9 & 66.0 & 78.7 & 96.0 & 70.3 & 74.1 & 63.1 & 10.2 & 54.1 & 72.4\\
    \midrule
    3DGenZ & 12.9 & 19.1 & 34.1 & 92.8 & 56.3 & 39.2 & 25.4 & 91.5 & 57.3 & 62.3 & 13.9 & \textbf{2.4} & 4.9 & 8.1 \\
    3DPC-GZSL & 16.7 & 33.0 & 48.3 & \textbf{96.0} & 57.2 & 44.3 & 40.4 & 91.9 & 54.2 & 64.8 & \textbf{22.3} & 1.2 & 6.2 & 9.3 \\
    \midrule
    \textbf{Ours} & \textbf{20.4} & \textbf{43.4} & \textbf{52.4} & 95.7 & \textbf{71.2} & \textbf{52.8} & \textbf{69.0} & \textbf{94.6} & \textbf{64.3} & \textbf{67.6} & 17.7 & 0.4 & \textbf{17.6} & \textbf{12.3} \\
  \bottomrule
\end{tabular} 
\caption{Class-wise performance comparisons of 3D GZSL Semantic Segmentation benchmarks in terms of mIoU(\%) and HmIoU(\%) on S3DIS dataset.}
\label{tab:S3DIS_classwise}
\end{table*}

\clearpage

\noindent \textbf{Analysis of Dynamic Calibration.} This section investigates the impact of dynamic calibration on both seen and unseen classes, based on experiments conducted on the ScanNet v2~\cite{c:304} and S3DIS~\cite{c:305} datasets. \fref{fig:class_wise_results} presents per-class IoU scores along with the corresponding estimated uncertainty, ranging from 0 to 1--where higher values indicate greater prediction uncertainty. In the plots, blue bars denote IoU without dynamic calibration, green bars represent IoU with dynamic calibration applied, and the red line indicates the estimated uncertainty for each class.

As expected, the estimated uncertainty tend to be higher for unseen classes (shown in purple) and lower for seen classes (shown in black). For classes with high uncertainty, dynamic calibration consistently enhances segmentation performance, as demonstrated by a noticeable increase between the blue and green bars. This effect is particularly prominent in the S3DIS dataset, where dynamic calibration plays a more substantial role--consistent with the ablation results reported in Table~\ref{tab:Module_component_analysis} of the main manuscript. Furthermore, we observe that the gap in uncertainty between seen and unseen classes is smaller in ScanNet v2 compared to S3DIS, suggesting a less pronounced effect of dynamic calibration on ScanNet v2.

For well-predicted seen classes with high IoU, such as ``floor'' and ``wall'' in ScanNet v2 and ``ceiling'' and ``floor'' in S3DIS, the estimated uncertainty remains low, indicating high model confidence and accurate predictions. In contrast, certain seen classes like ``picture'' and ``window'' in ScanNet v2 exhibit both low IoU and high uncertainty. This is attributed to the nature of the data: ScanNet v2 lacks RGB input, making it challenging to distinguish planar objects such as ``pictures'' and ``windows'' from ``walls''. Specifically, over 87\% of points labeled picture'' are incorrectly segmented into the wall'' class, and approximately 56\% of window'' points are also misclassified as wall'', as further illustrated in the qualitative results in \fref{fig:qualitative_comparison_ScanNet}.

\begin{figure}[!h]
    \centering
    \begin{tabular}{@{}cc@{}}
         {ScanNet v2} & {S3DIS}\\
        \includegraphics[width=0.37\textwidth]{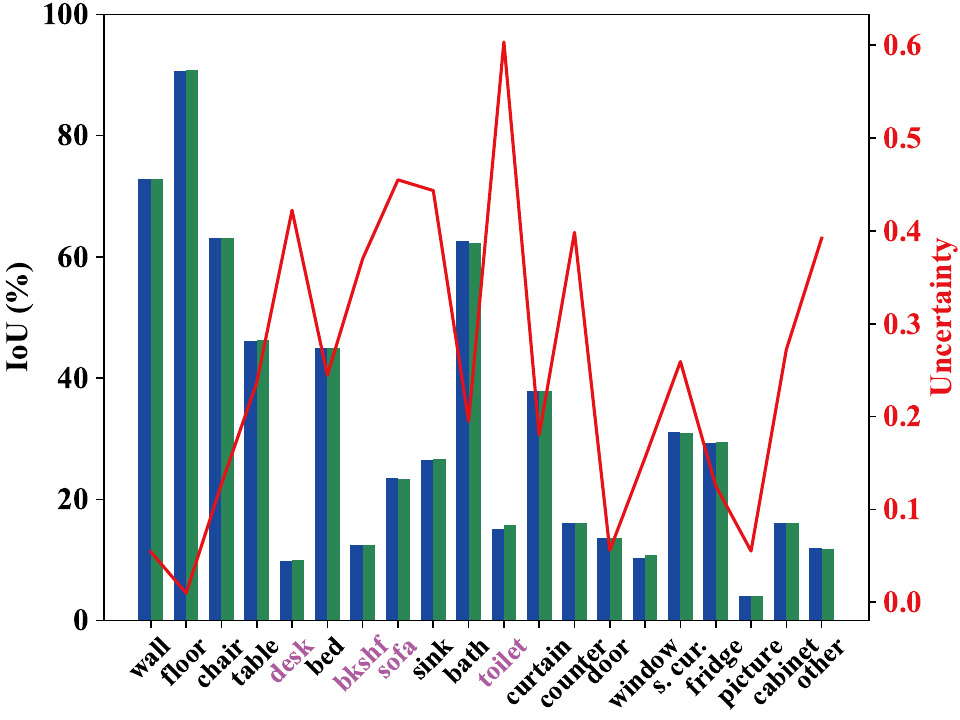}  \hspace{0.2cm}& \includegraphics[width=0.37\textwidth]{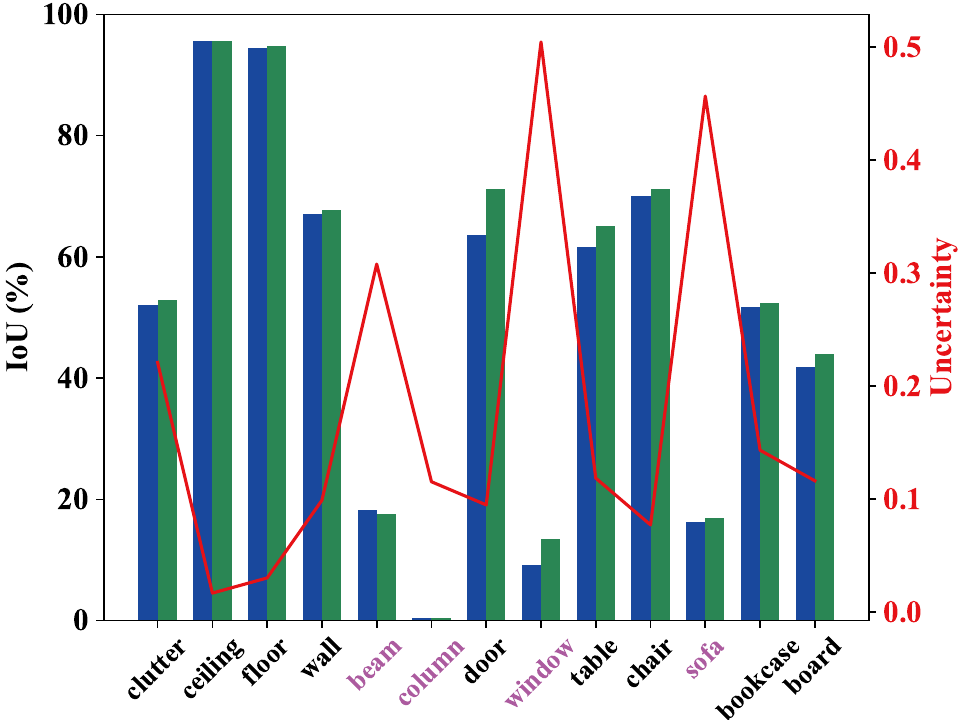} \\
        \multicolumn{2}{c}{\includegraphics[width=0.35\textwidth]{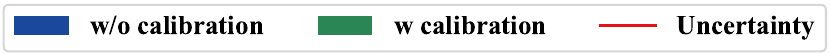}}\\
    \end{tabular}
    \caption{Trend in estimated uncertainty and IoU(\%) variations with and without the proposed dynamic calibration for each class. The black classes represent seen classes, and the magenta classes represent unseen classes.} \label{fig:class_wise_results}
\end{figure}

\noindent \textbf{Analysis of Semantic Tuning.} To enrich the semantic information encoded in the text embeddings, we introduce a learnable scene-conditioned tuning vector. To assess its effectiveness, we compare it against a baseline that uses randomly initialized vectors drawn from a standard normal distribution. We conduct experiments on both the ScanNet v2~\cite{c:304} and S3DIS~\cite{c:305} datasets using 10 different random seeds, and report the results as the mean ± standard deviation. To validate the effectiveness of the semantic tuning vector, we use a baseline model without calibrated stacking~\cite{c:237}. Table~\ref{tab:Semantic_tuning} presents the experimental results, where ``Proposed'' denotes the use of the scene-conditioned tuning vector, and ``Random'' corresponds to the use of randomly initialized vectors. Experimental results show that using a randomly initialized vector for semantic tuning leads to performance degradation in both seen and unseen mIoU, indicating that the scene-conditioned tuning vector meaningfully contributes to the synthesis of decoder features. Notably, as shown in Table~\ref{tab:Random_seeds}, our method surpasses the performance of 3DPC-GZSL~\cite{c:226} even without calibrated stacking.

\begin{table}[!h]
  \centering
  \renewcommand{\tabcolsep}{4.0mm}{
  \begin{tabular}{c|c|ccc|c}
    \toprule
    \multirow{2}{*}{Dataset} & \multirow{2}{*}{Semantic Vector} & \multicolumn{3}{c|}{mIoU} & \multirow{2}{*}{HmIoU} \\
    \cline{3-5}
    & & Seen & Unseen &All & \\
    \midrule
    \multirow{2}{*}{ScanNet v2} & \textbf{Proposed} & \textbf{34.83 ± 1.28} & \textbf{15.23 ± 0.89} & \textbf{30.91 ± 0.99} & \textbf{21.17 ± 0.82}\\
     & Random & 31.30 ± 1.10 & 4.51 ± 1.20 & 25.94 ± 0.88 & 7.83 ± 1.75\\
    \midrule
    \multirow{2}{*}{S3DIS} & \textbf{Proposed} & \textbf{65.47 ± 2.60} & \textbf{10.06 ± 2.17} & \textbf{48.42 ± 2.28} & \textbf{17.38 ± 3.29} \\
     & Random & 56.18 ± 4.45 & 5.59 ± 3.24 & 40.61 ± 4.01 & 9.98 ± 5.36 \\
    \bottomrule
\end{tabular}
}
\caption{Performance comparison of semantic tuning methods in terms of mIoU (\%) and HmIoU (\%) on the ScanNet v2 and S3DIS datasets. Results are averaged over 10 random seeds and reported as mean ± standard deviation.}
\label{tab:Semantic_tuning}
\end{table} 

\clearpage

\noindent \textbf{Performance Analysis Across Random Seeds. }
Following the baseline implementation~\cite{c:225, c:226}, the default random seed is set to 1, and results reported with this seed are used as a reference. To assess the stability of the methods, we further evaluate both 3DPC-GZSL~\cite{c:226} and the proposed approach across 10 random seeds. Table~\ref{tab:Random_seeds} summarizes the averaged performance and standard deviation. On the ScanNet v2~\cite{c:304} dataset, the proposed method achieves an average improvement of +1.19\% in HmIoU, while on the S3DIS~\cite{c:305} dataset, it yields an average gain of +2.57\% HmIoU. These results demonstrate that our method consistently enhances performance across different random seeds, indicating its robustness and stability.

\begin{table}[!h]
  \centering
  \renewcommand{\tabcolsep}{4.0mm}{
  \begin{tabular}{@{}c@{}|c|ccc|c}
    \toprule
    \multirow{2}{*}{Dataset} & \multirow{2}{*}{Methods} & \multicolumn{3}{c|}{mIoU} & \multirow{2}{*}{HmIoU} \\
    \cline{3-5}
    & & Seen & Unseen &All & \\
    \midrule
    \multirow{2}{*}{ScanNet v2 } & 3DPC-GZSL &  33.78 ± 1.14 & 14.27 ± 0.68 & 29.86 ± 0.91 & 20.05 ± 0.69 \\
     & \textbf{E3DPC-GZSL(ours)} & \textbf{34.92 ± 1.26} & \textbf{15.29 ± 0.89} & \textbf{30.99 ± 0.98} & \textbf{21.24 ± 0.82} \\
    \midrule
    \multirow{2}{*}{S3DIS} & 3DPC-GZSL & 62.34 ± 2.42 & 9.26 ± 1.59 & 49.01 ± 1.76 & 16.08 ± 2.43 \\
     & \textbf{E3DPC-GZSL(ours)} & \textbf{66.66 ± 2.01} & \textbf{10.88 ± 2.03} & \textbf{49.49 ± 1.84} & \textbf{18.65 ± 3.03} \\
    \bottomrule
\end{tabular}
}
\caption{Performance comparison of 3DPC-GZSL and the proposed method in terms of mIoU (\%) and HmIoU (\%) on the ScanNet v2 and S3DIS datasets. Results are averaged over 10 random seeds and reported as mean ± standard deviation.}
\label{tab:Random_seeds}
\end{table} 

\noindent \textbf{Qualitative Comparisons. } 
Figures \ref{fig:qualitative_comparison_ScanNet} and \ref{fig:qualitative_comparison_S3DIS} show the qualitative results of E3DPC-GZSL compared to previous studies on the ScanNet v2~\cite{c:304} and S3DIS~\cite{c:305} datasets, respectively. 

As shown in \fref{fig:qualitative_comparison_ScanNet}, the proposed E3DPC-GZSL method improves segmentation performance not only on unseen classes but also on seen classes in ScanNet v2 dataset. In particular, improved recognition of seen classes is observed in examples such as the ``refrigerator'' in Scene 5 and the ``door'' in Scenes 28, 35, 60, and 206. Notably, in Scene 35, while 3DGenZ~\cite{c:225} fails to correctly predict the unseen class and 3DPC-GZSL~\cite{c:226} overcompensates, leading to misclassification of the seen class ``curtain'', E3DPC-GZSL successfully recovers curtain segmentation by preventing excessive unseen class activation. Additionally, in Scene 128, all baseline methods misclassify the seen class ``table'' as ``desk'', whereas our method correctly identifies it as ``table''. Furthermore, the segmentation performance on unseen classes such as ``desk'' (Scene 60) and ``toilet'' (Scenes 35 and 206) is preserved, indicating that our method achieves balanced improvements without degrading unseen class recognition.

\fref{fig:qualitative_comparison_S3DIS} presents a qualitative comparison between prior works~\cite{c:225, c:226} and the proposed E3DPC-GZSL method on the S3DIS dataset. In hallway scenes, our method significantly improves the segmentation of seen classes such as ``door'', ``floor'', ``wall'', and `ceiling''. While previous methods often misclassify parts of ``door'' and ``wall'' as the unseen class ``window,'' our method correctly distinguishes these as seen classes, demonstrating improved robustness. In the conference room example, previous methods incorrectly classify the board as the unseen class ``column,'' whereas E3DPC-GZSL accurately identifies it as ``board.'' Similarly, in the office scene, prior methods show mixed predictions of ``window'' and ``column'' for ``window'' regions, while our method reduces this confusion and achieves more precise ``window'' segmentation. These results indicate that E3DPC-GZSL not only enhances seen class segmentation but also leads to more accurate recognition of unseen classes.

\begin{figure*}[!t]
    \centering
    \begin{tabular}{@{}c@{}c@{}c@{}c@{}c@{}}
    & Ground-truth & 3DGenZ & 3DPC-GZSL & E3DPC-GZSL~(Ours) \vspace{0.2cm}\\
    \rotatebox[origin=c]{90}{Scene 5} \vspace{0.5cm} & \raisebox{-.5\height}{\includegraphics[width=0.23\textwidth]{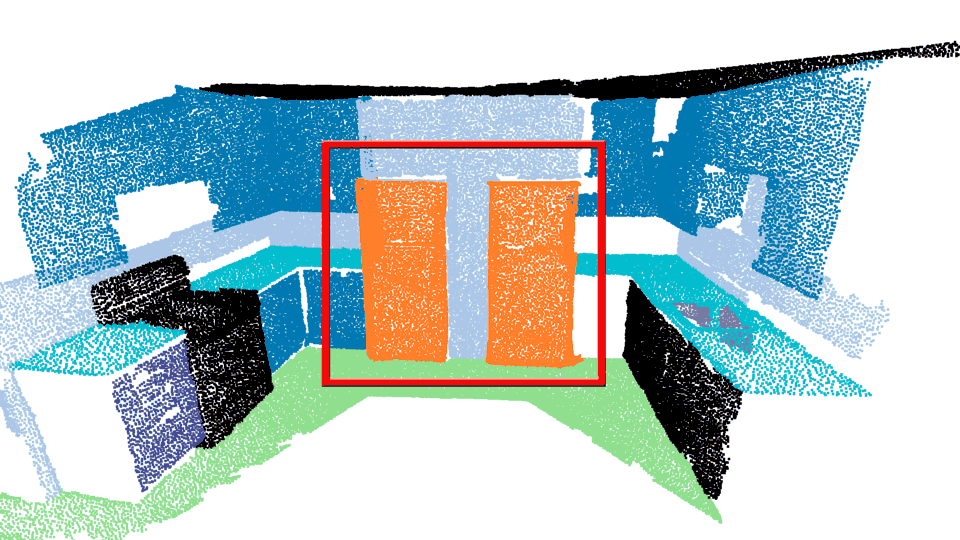}} & \raisebox{-.5\height}{\includegraphics[width=0.23\textwidth]{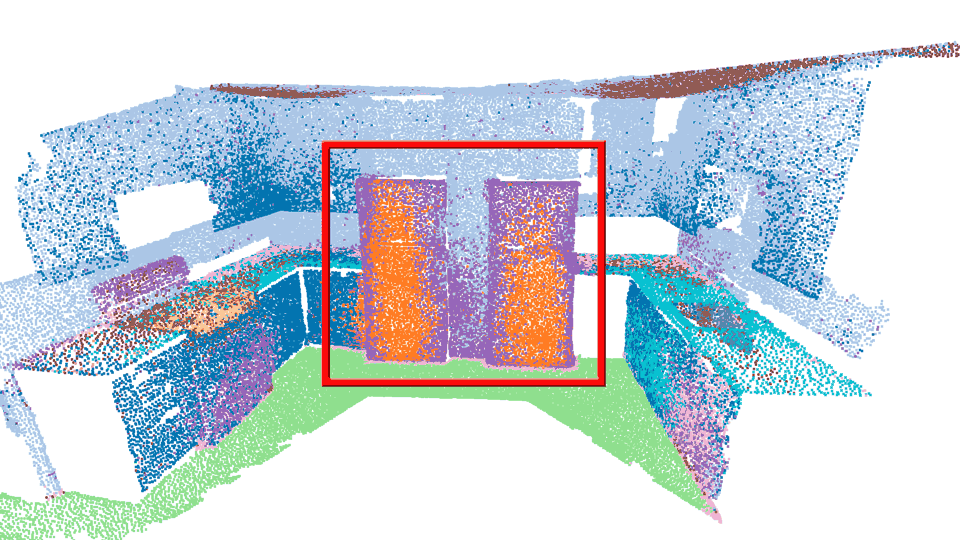}} & \raisebox{-.5\height}{\includegraphics[width=0.23\textwidth]{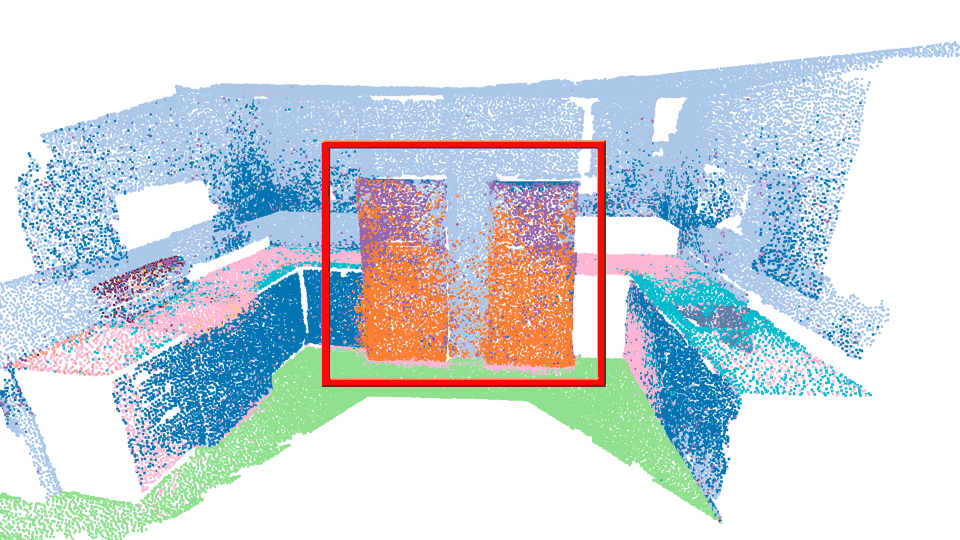}} & \raisebox{-.5\height}{\includegraphics[width=0.23\textwidth]{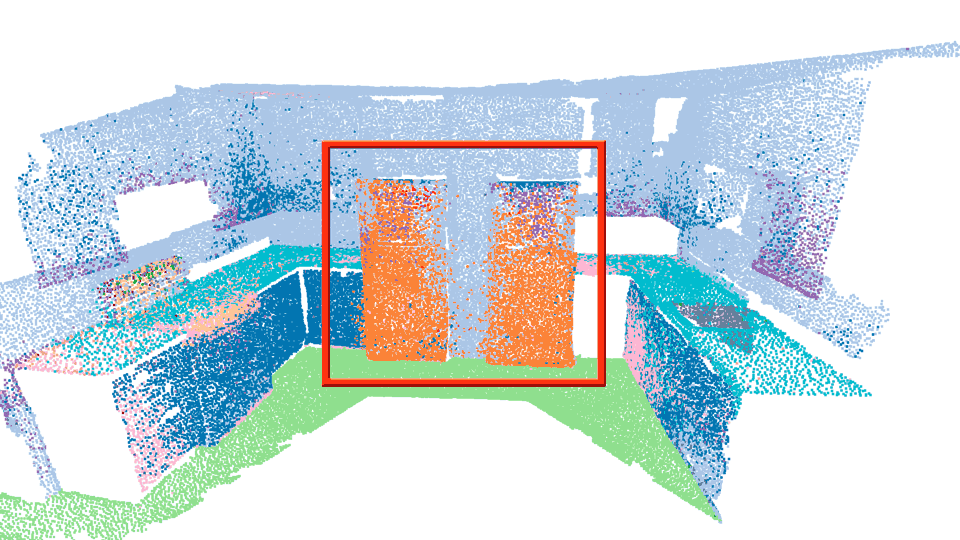}} \\ 
    \rotatebox[origin=c]{90}{Scene 28} \vspace{0.5cm} & \raisebox{-.5\height}{\includegraphics[width=0.23\textwidth]{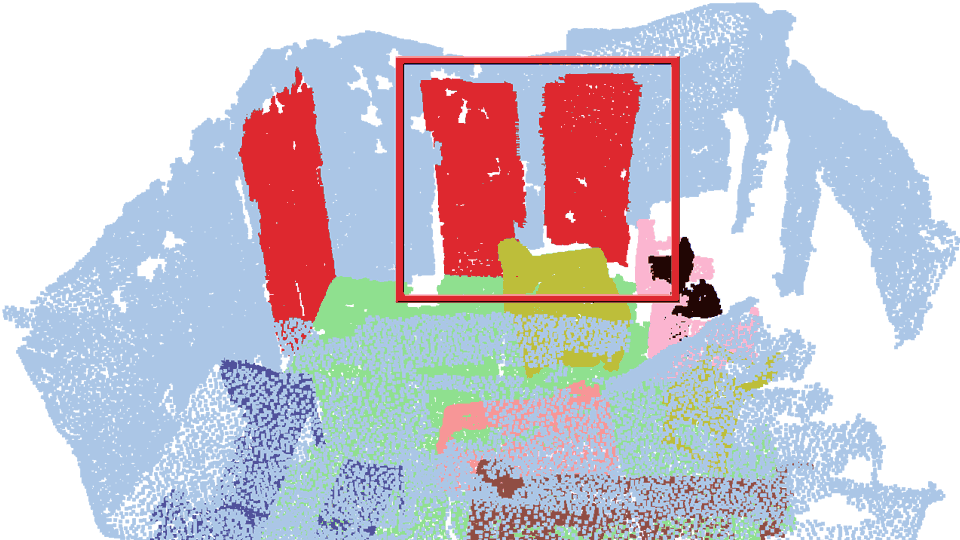}} & \raisebox{-.5\height}{\includegraphics[width=0.23\textwidth]{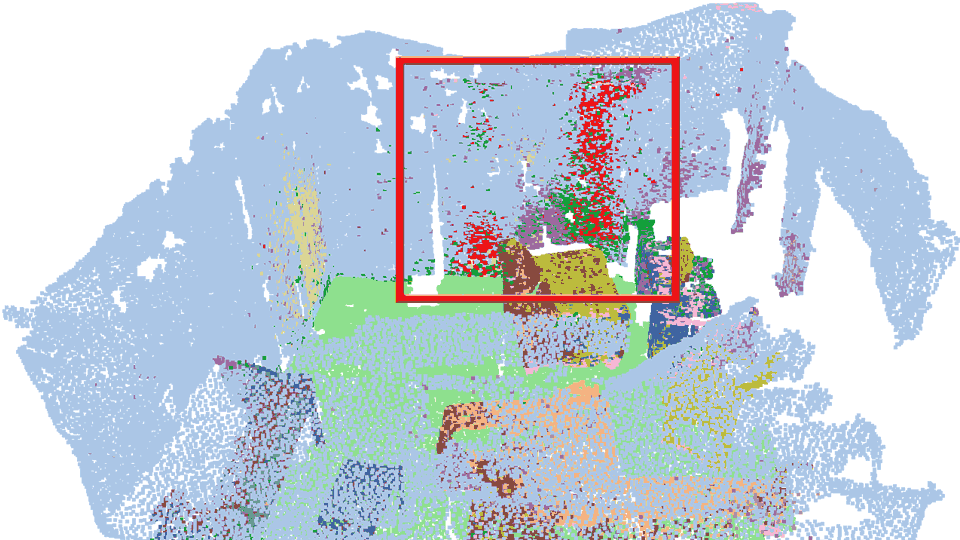}} & \raisebox{-.5\height}{\includegraphics[width=0.23\textwidth]{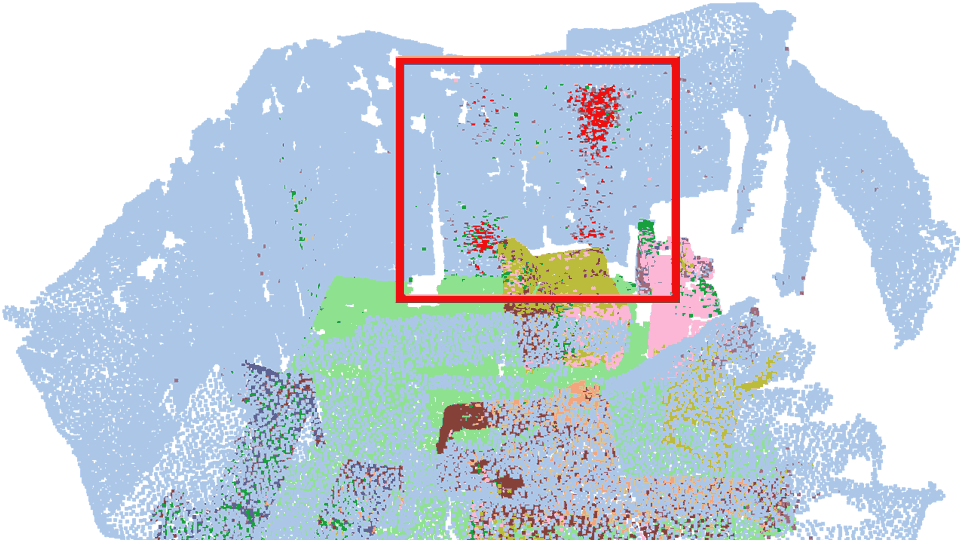}} & \raisebox{-.5\height}{\includegraphics[width=0.23\textwidth]{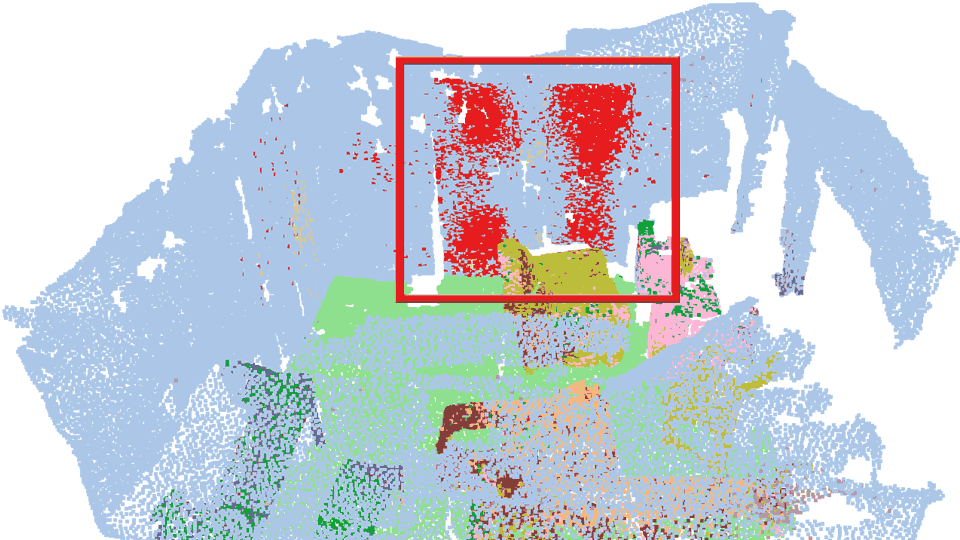}} \\ 
    \rotatebox[origin=c]{90}{Scene 35} \vspace{0.5cm} & \raisebox{-.5\height}{\includegraphics[width=0.23\textwidth]{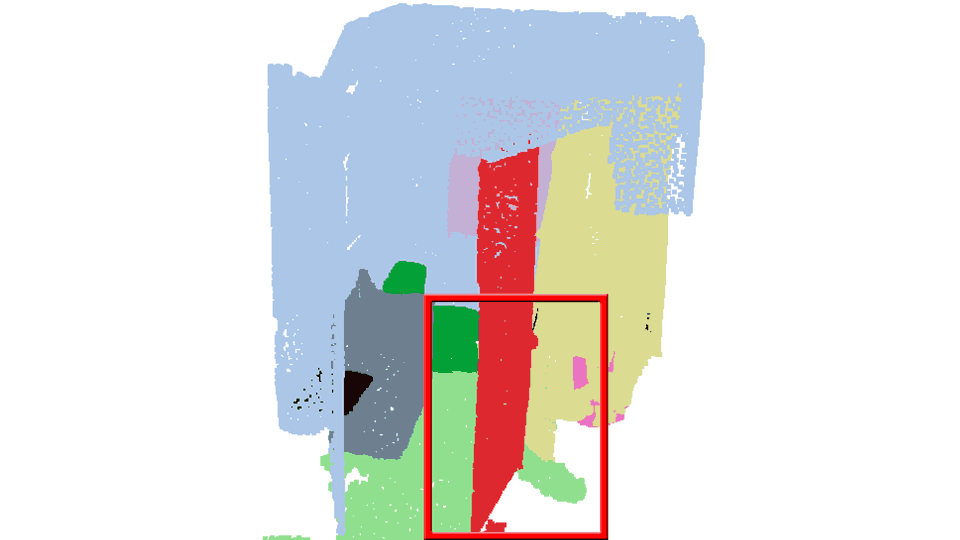}} & \raisebox{-.5\height}{\includegraphics[width=0.23\textwidth]{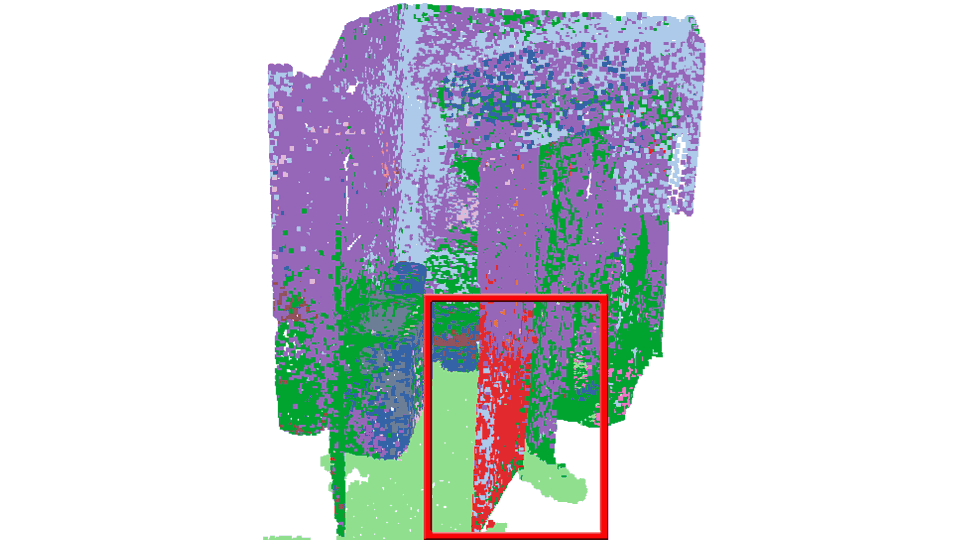}} & \raisebox{-.5\height}{\includegraphics[width=0.23\textwidth]{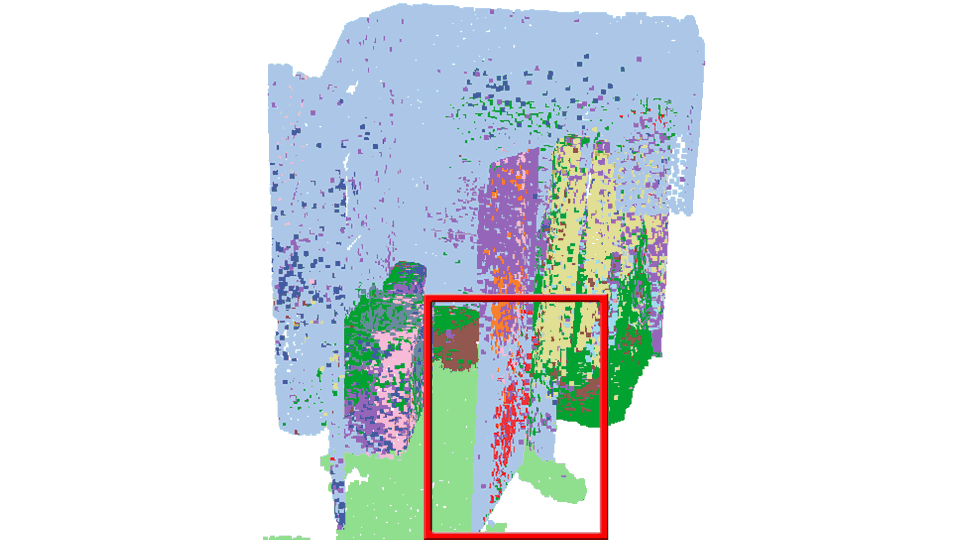}} & \raisebox{-.5\height}{\includegraphics[width=0.23\textwidth]{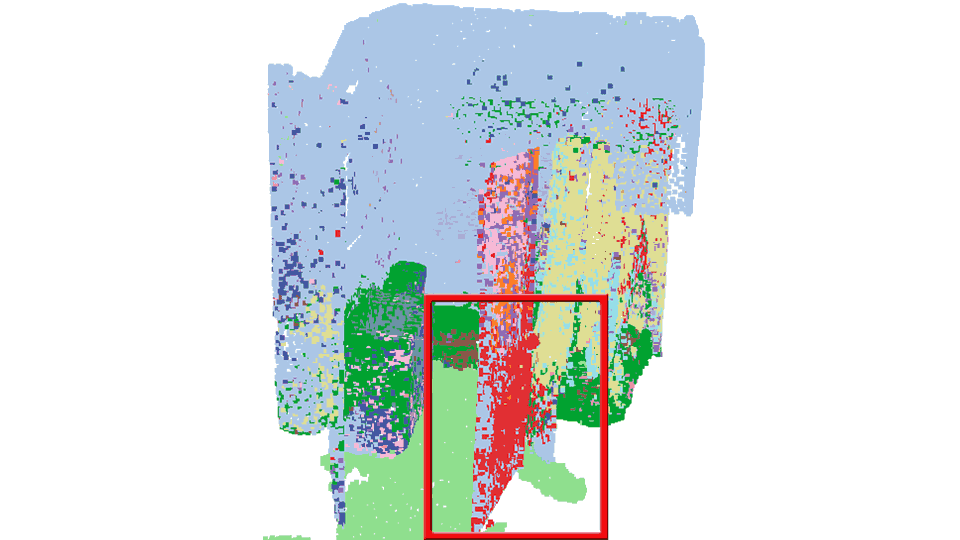}} \\
    \rotatebox[origin=c]{90}{Scene 40} \vspace{0.5cm} & \raisebox{-.5\height}{\includegraphics[width=0.23\textwidth]{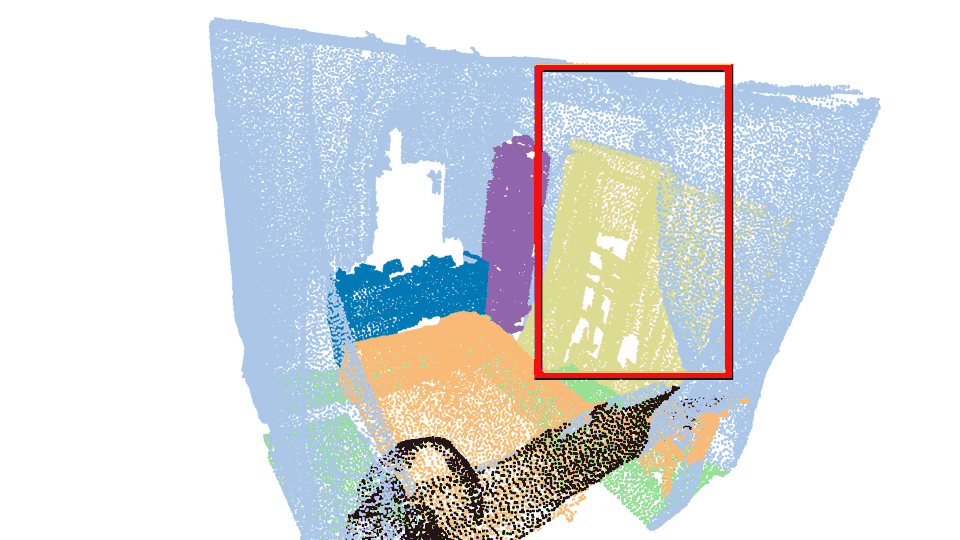}} & \raisebox{-.5\height}{\includegraphics[width=0.23\textwidth]{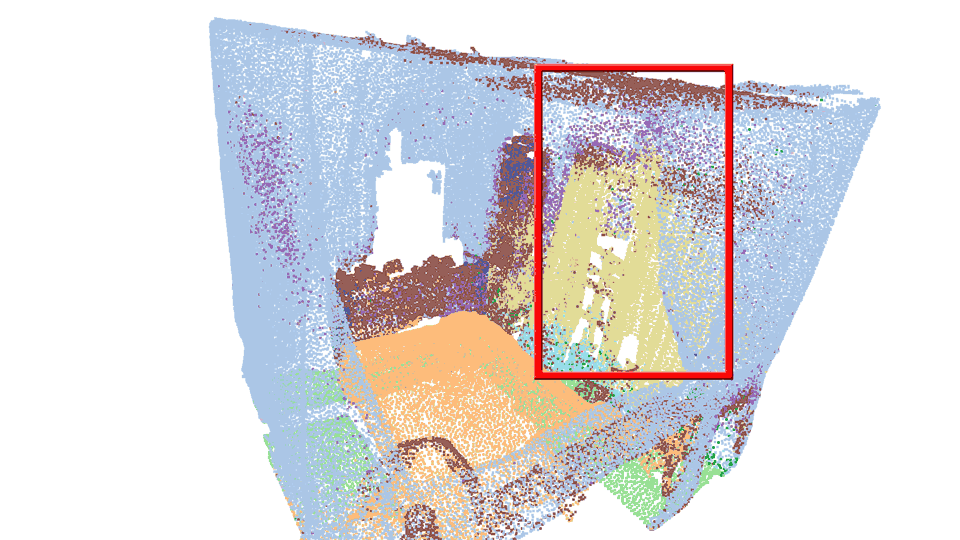}} & \raisebox{-.5\height}{\includegraphics[width=0.23\textwidth]{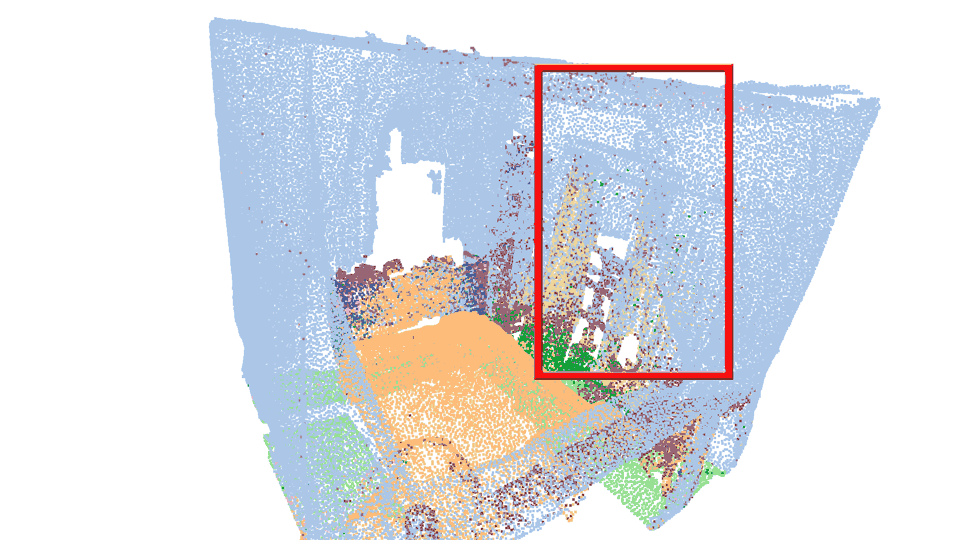}} & \raisebox{-.5\height}{\includegraphics[width=0.23\textwidth]{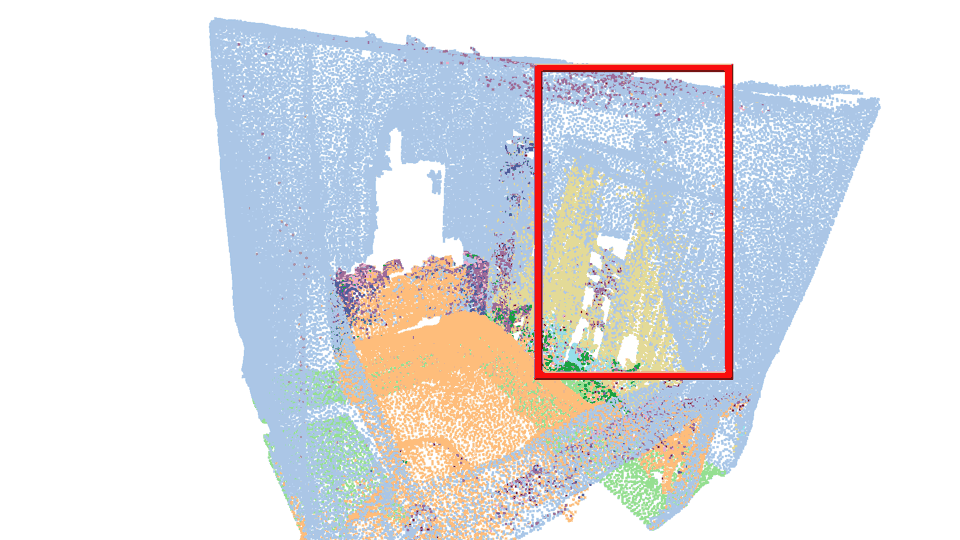}} \\
    \rotatebox[origin=c]{90}{Scene 60} \vspace{0.5cm} & \raisebox{-.5\height}{\includegraphics[width=0.23\textwidth]{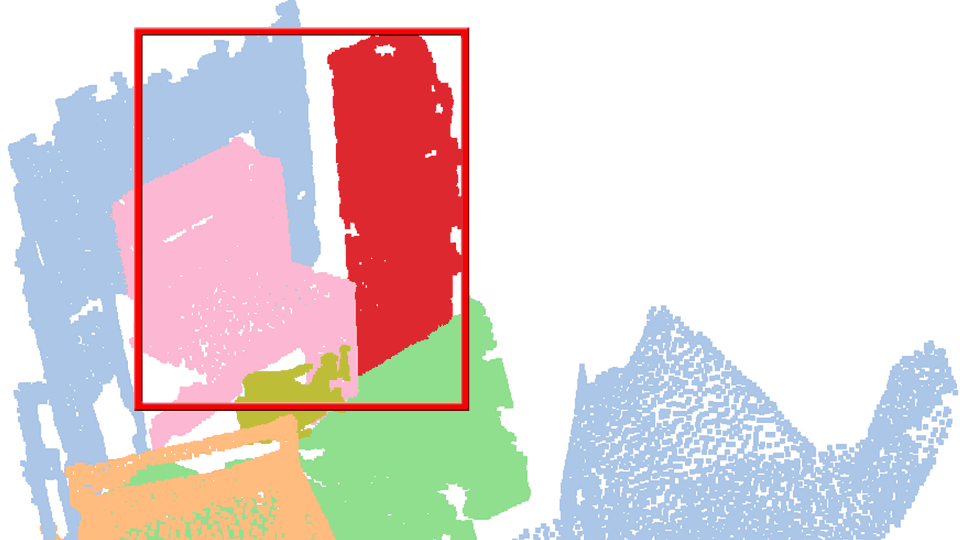}} & \raisebox{-.5\height}{\includegraphics[width=0.23\textwidth]{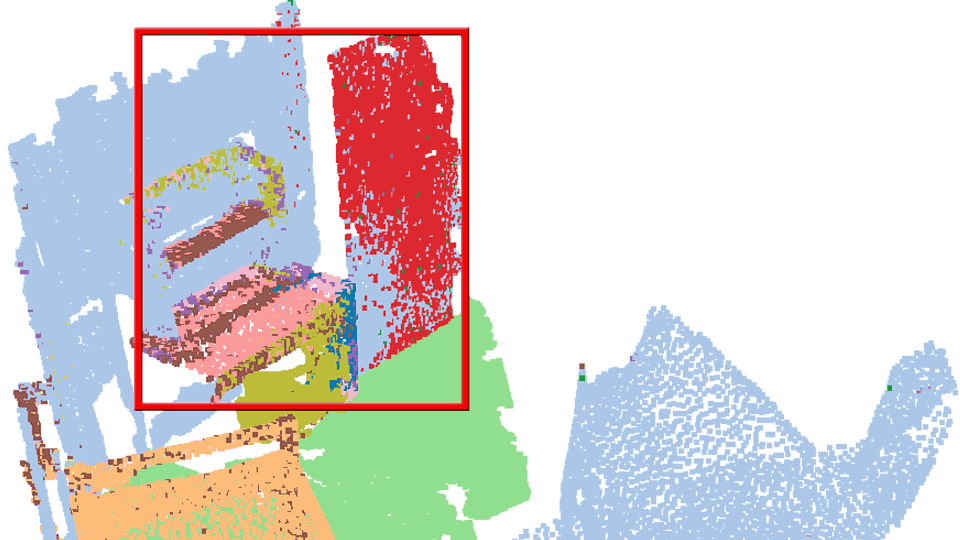}} & \raisebox{-.5\height}{\includegraphics[width=0.23\textwidth]{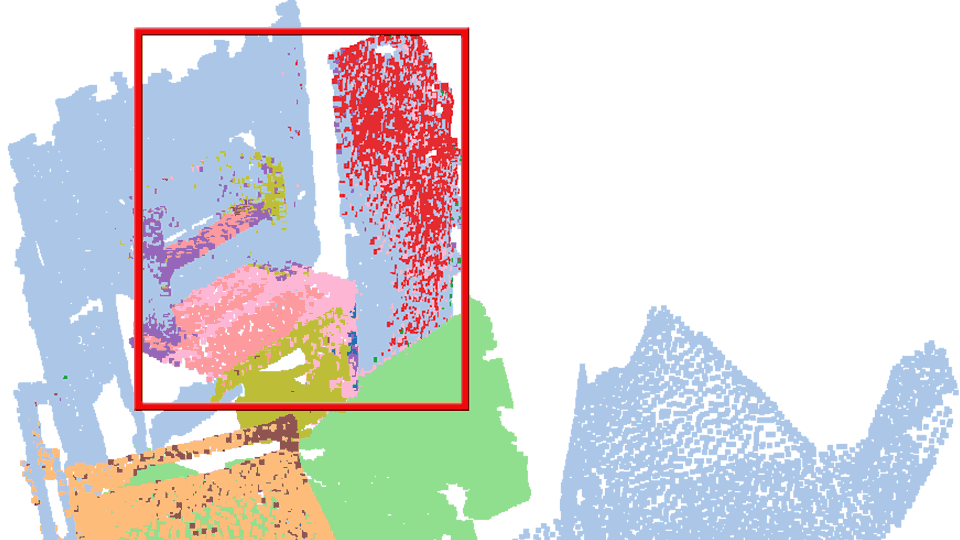}} & \raisebox{-.5\height}{\includegraphics[width=0.23\textwidth]{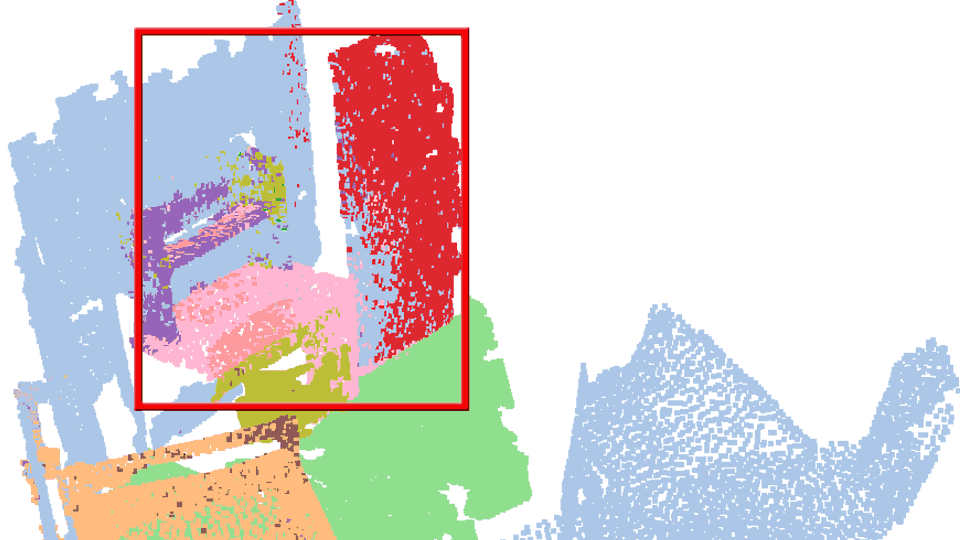}} \\
    \rotatebox[origin=c]{90}{Scene 128} \vspace{0.5cm} & \raisebox{-.5\height}{\includegraphics[width=0.23\textwidth]{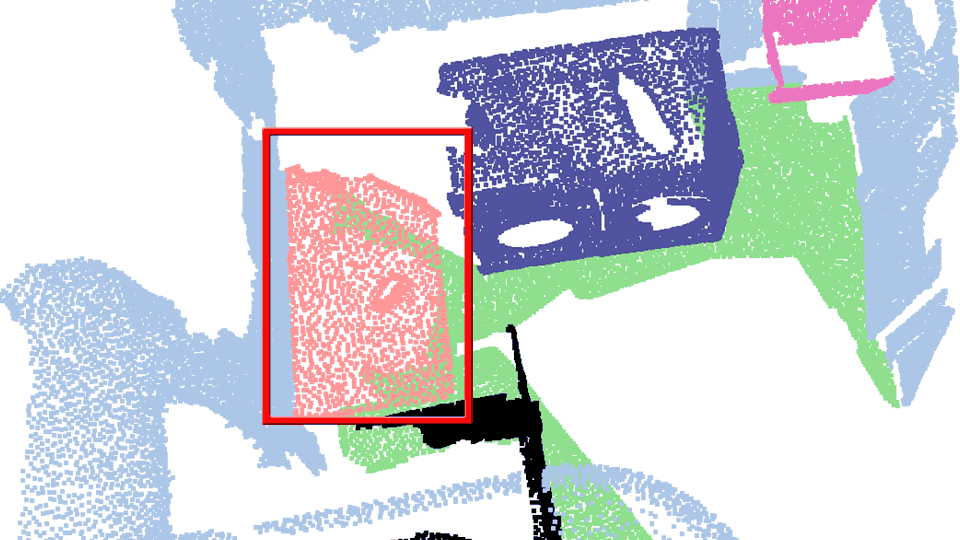}} & \raisebox{-.5\height}{\includegraphics[width=0.23\textwidth]{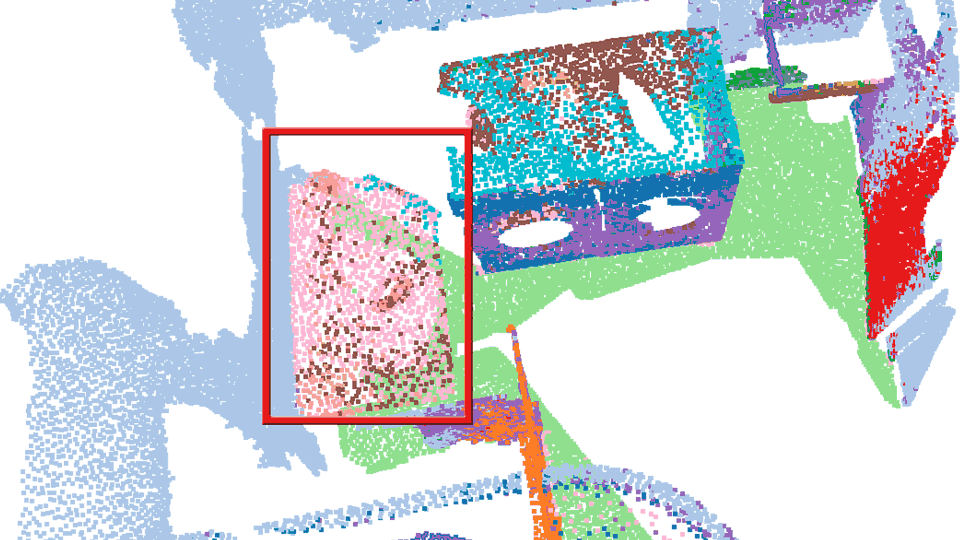}} & \raisebox{-.5\height}{\includegraphics[width=0.23\textwidth]{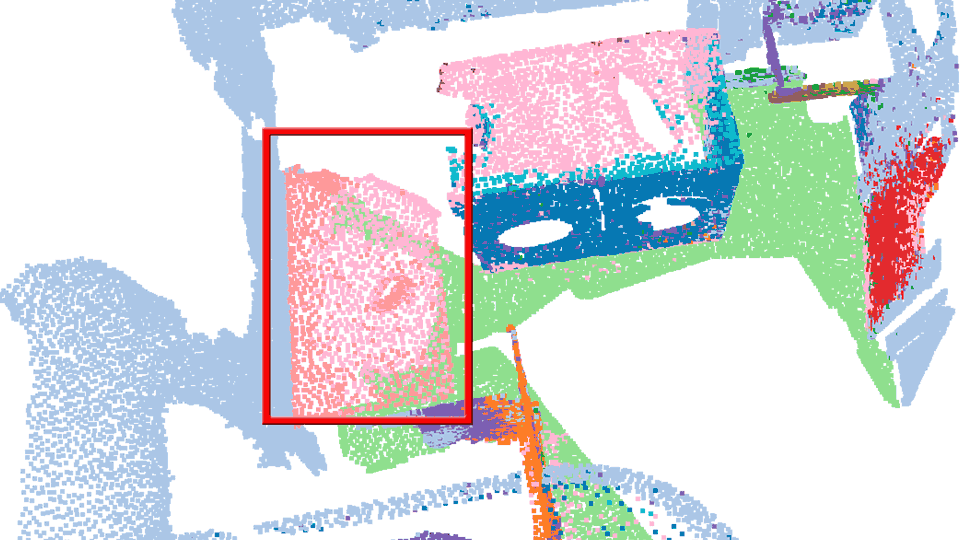}} & \raisebox{-.5\height}{\includegraphics[width=0.23\textwidth]{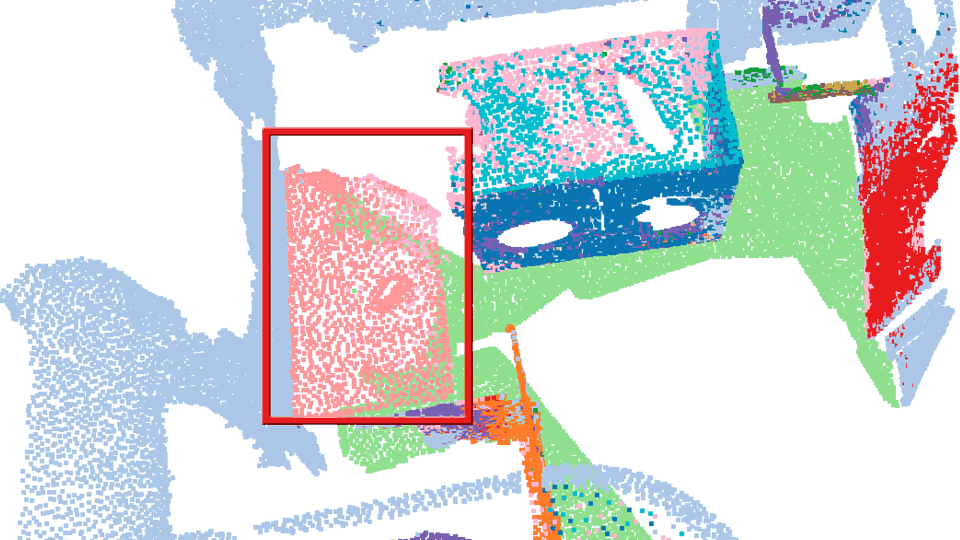}} \\
    \rotatebox[origin=c]{90}{Scene 206} \vspace{0.5cm} & \raisebox{-.5\height}{\includegraphics[width=0.23\textwidth]{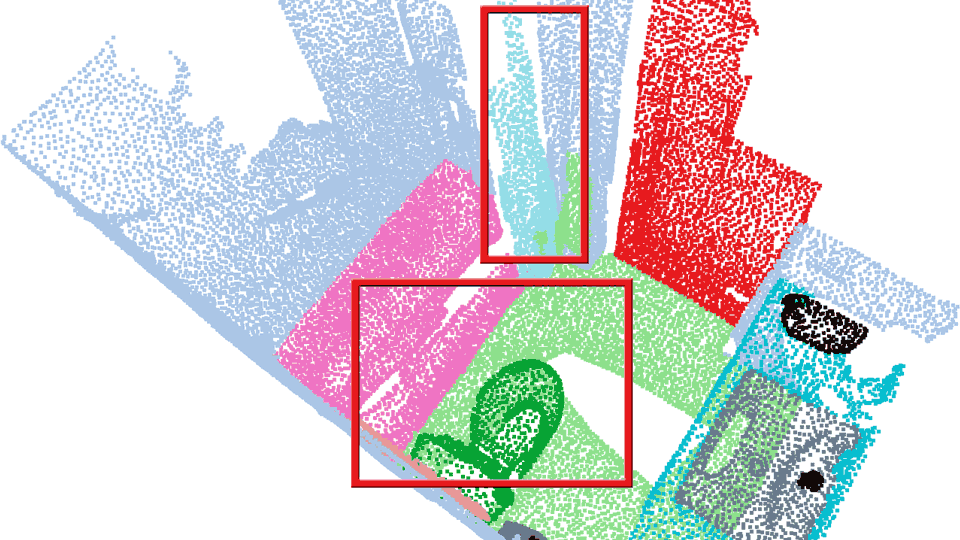}} & \raisebox{-.5\height}{\includegraphics[width=0.23\textwidth]{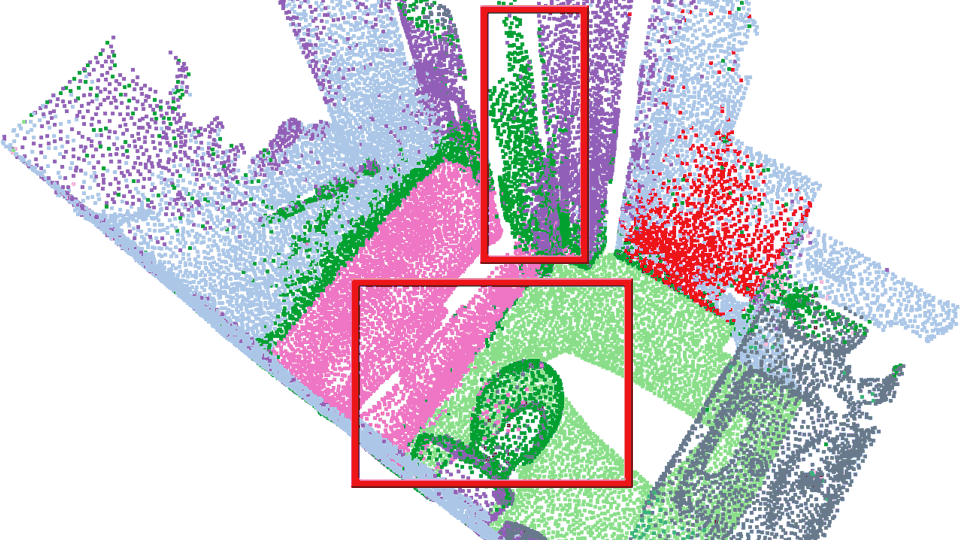}} & \raisebox{-.5\height}{\includegraphics[width=0.23\textwidth]{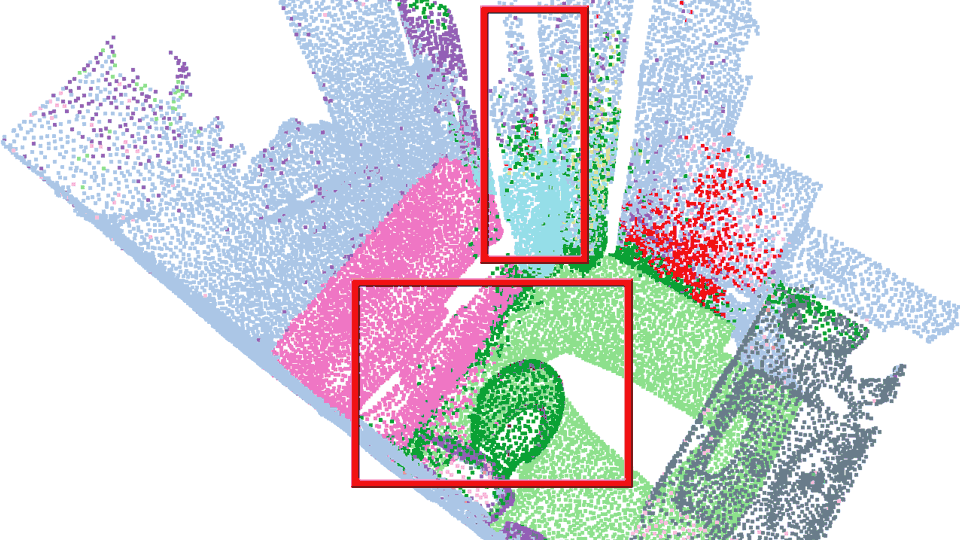}} & \raisebox{-.5\height}{\includegraphics[width=0.23\textwidth]{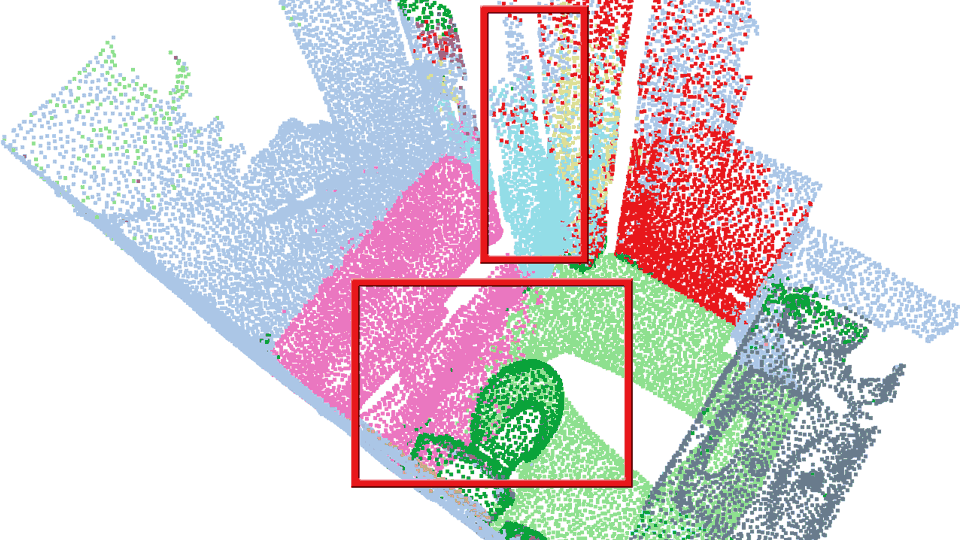}} \\
    \multicolumn{5}{c}{\includegraphics[width=0.95\textwidth]{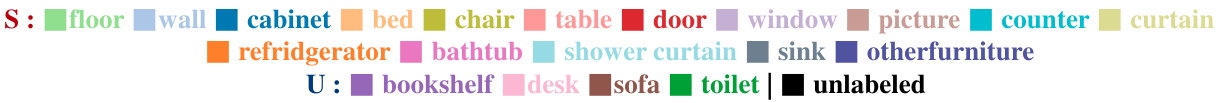}}\\
    \end{tabular}
    \caption{Qualitative comparison of the proposed model with other methods on ScanNet v2. The labels ``S'' in red denote seen classes, and ``U'' in blue denote unseen classes. } \label{fig:qualitative_comparison_ScanNet}
\end{figure*}

\begin{figure*}[t]
    \centering
    \begin{tabular}{@{}c@{}c@{}c@{}c@{}c@{}}
    & Ground-truth & 3DGenZ & 3DPC-GZSL & E3DPC-GZSL~(Ours) \vspace{0.2cm}\\
    \rotatebox[origin=c]{90}{hallway 2} \vspace{0.5cm} & \raisebox{-.5\height}{\includegraphics[width=0.23\textwidth]{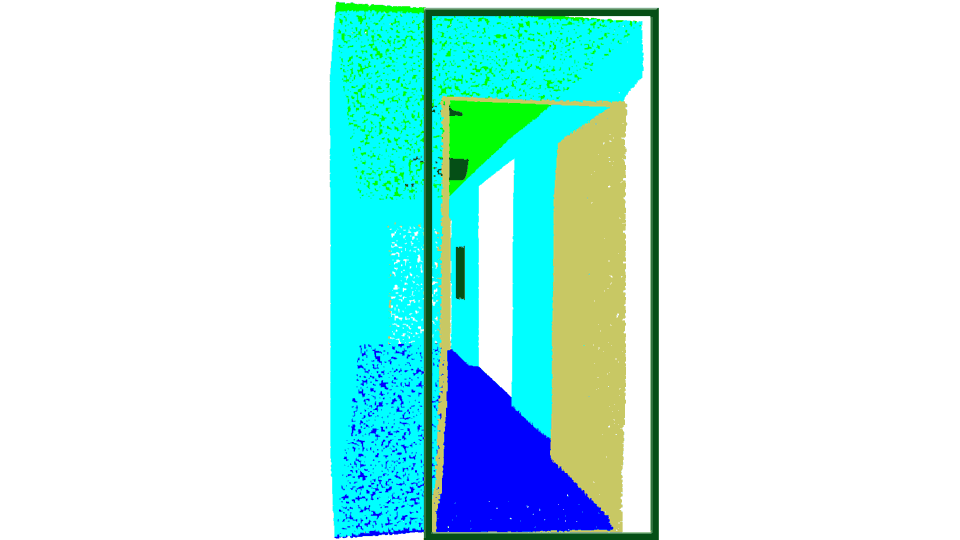}} & \raisebox{-.5\height}{\includegraphics[width=0.23\textwidth]{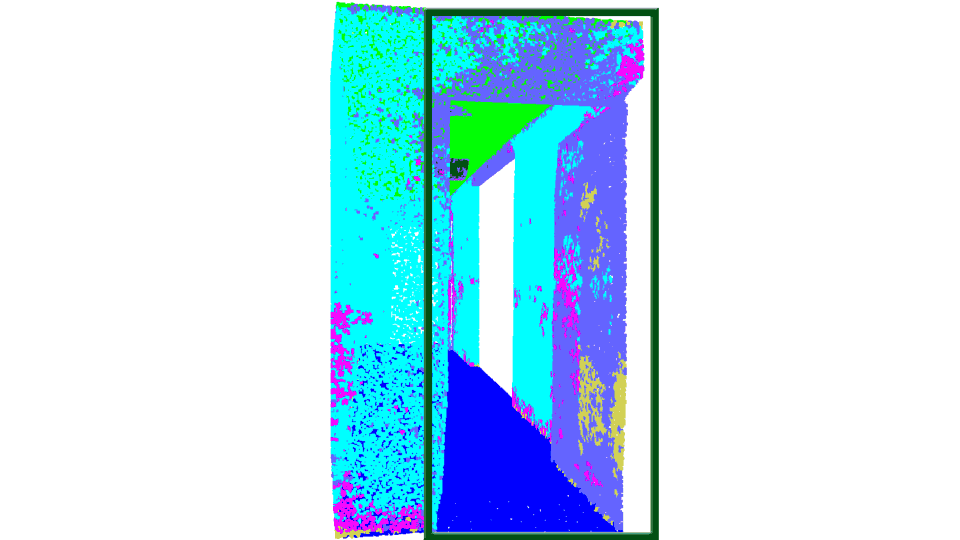}} & \raisebox{-.5\height}{\includegraphics[width=0.23\textwidth]{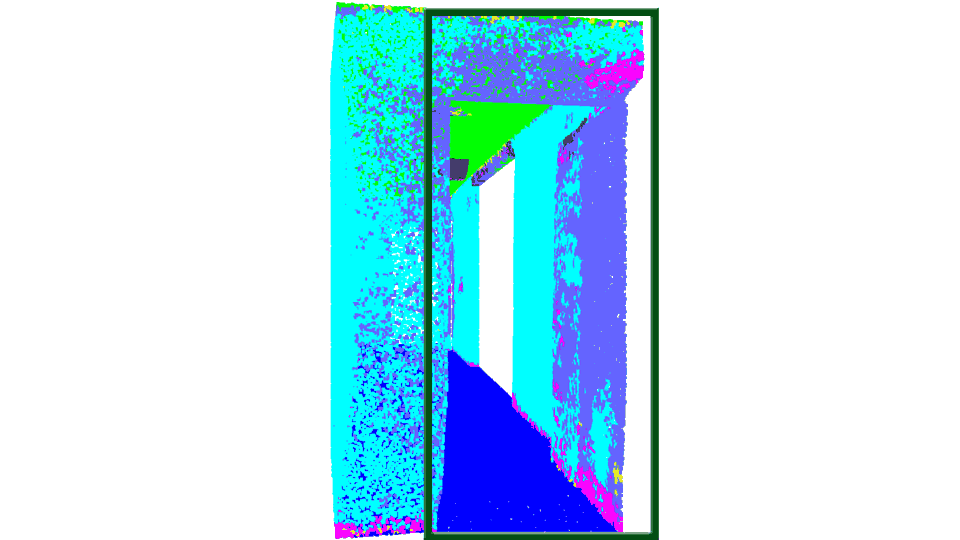}} & \raisebox{-.5\height}{\includegraphics[width=0.23\textwidth]{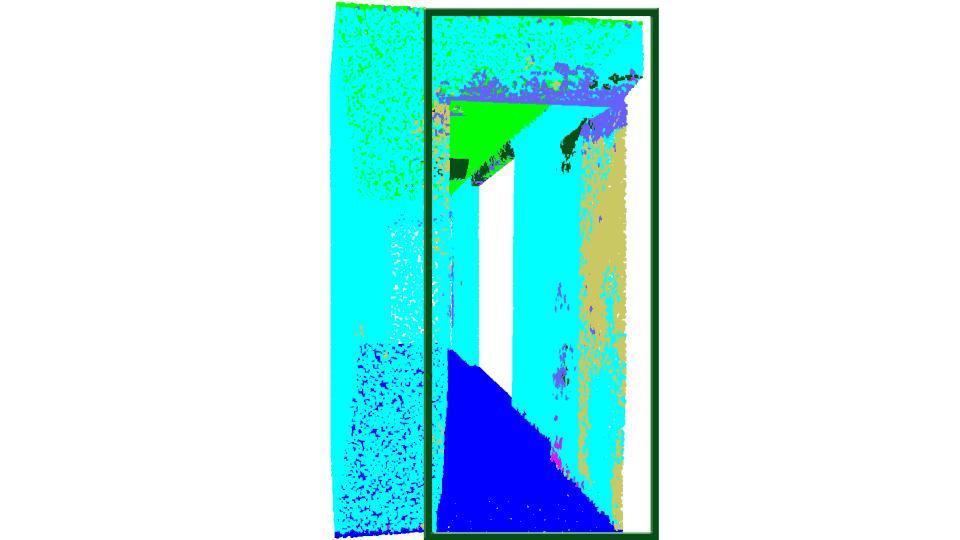}} \\
    \rotatebox[origin=c]{90}{hallway 3} \vspace{0.5cm} & \raisebox{-.5\height}{\includegraphics[width=0.23\textwidth]{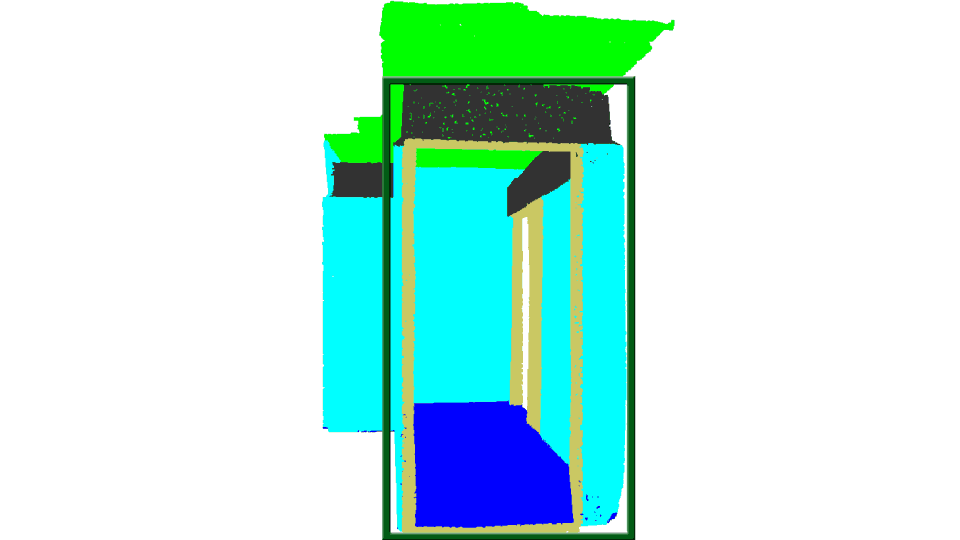}} & \raisebox{-.5\height}{\includegraphics[width=0.23\textwidth]{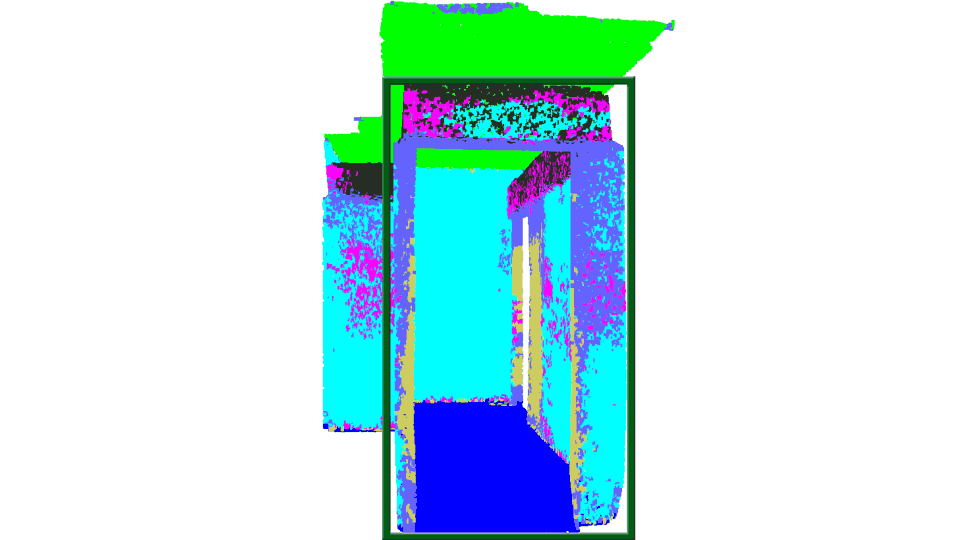}} & \raisebox{-.5\height}{\includegraphics[width=0.23\textwidth]{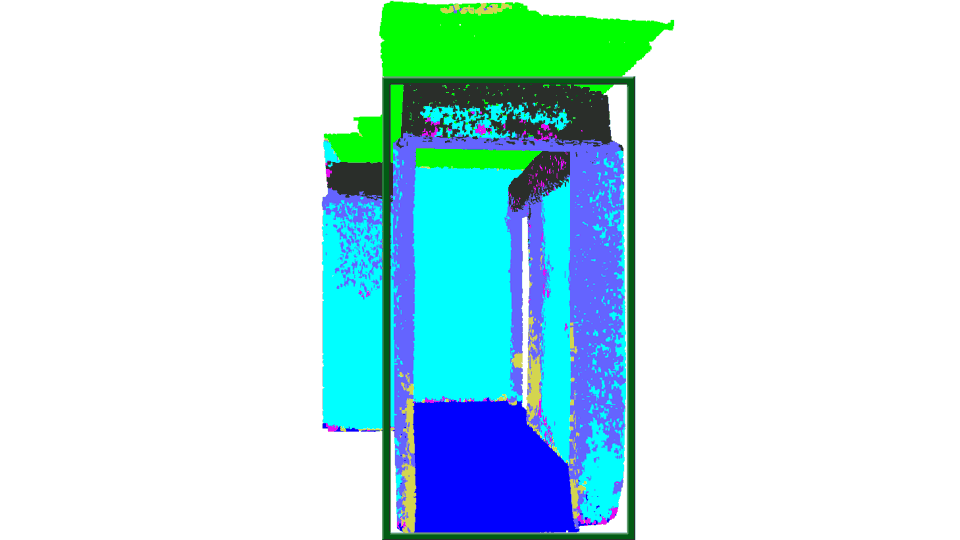}} & \raisebox{-.5\height}{\includegraphics[width=0.23\textwidth]{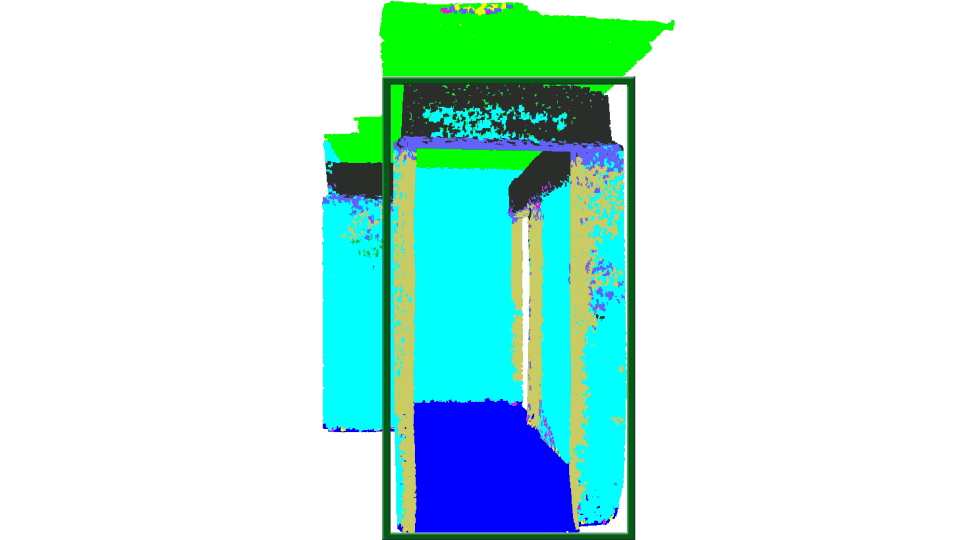}} \\
    \rotatebox[origin=c]{90}{hallway 4} \vspace{0.5cm} & \raisebox{-.5\height}{\includegraphics[width=0.23\textwidth]{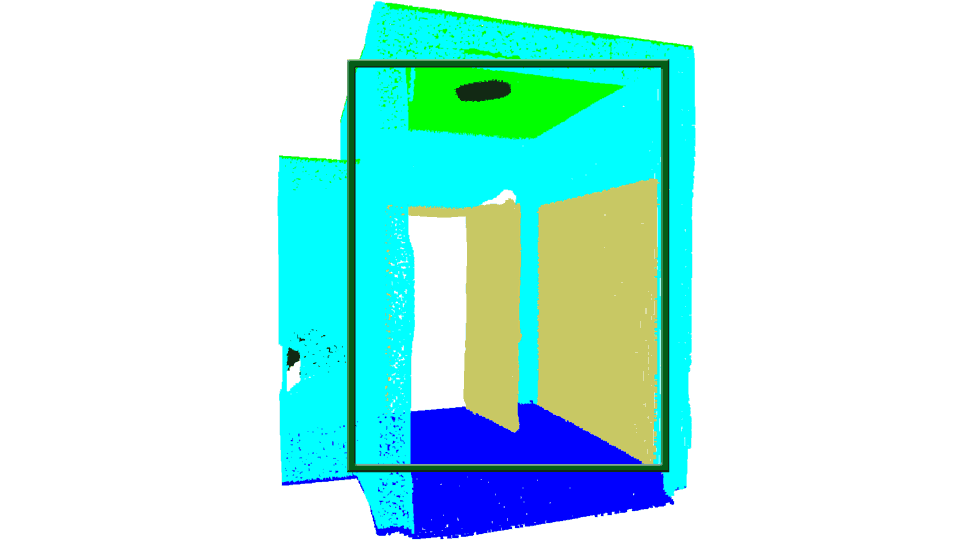}} & \raisebox{-.5\height}{\includegraphics[width=0.23\textwidth]{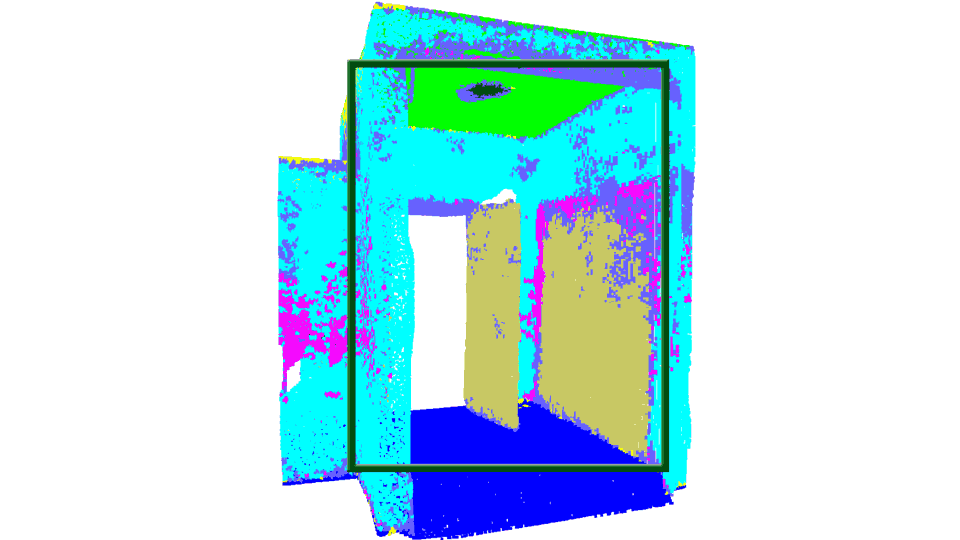}} & \raisebox{-.5\height}{\includegraphics[width=0.23\textwidth]{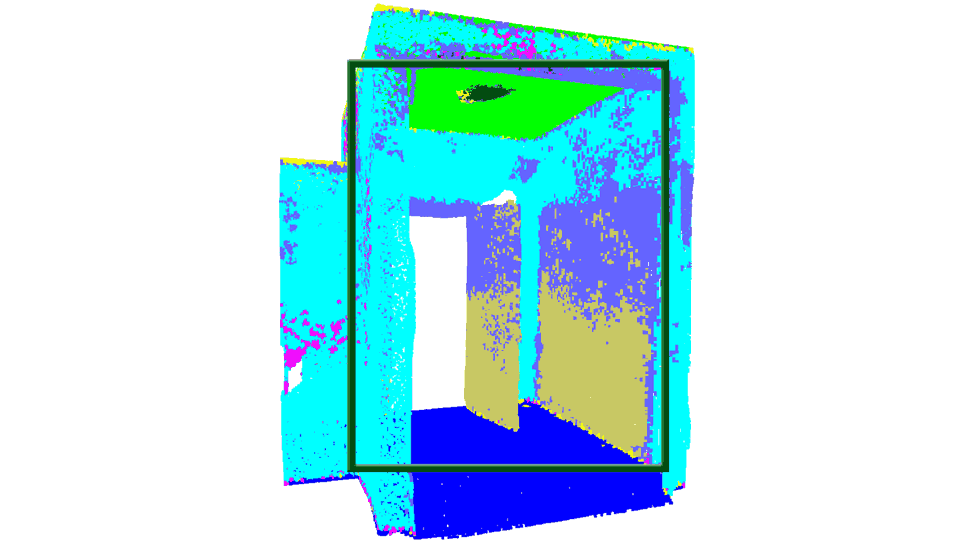}} & \raisebox{-.5\height}{\includegraphics[width=0.23\textwidth]{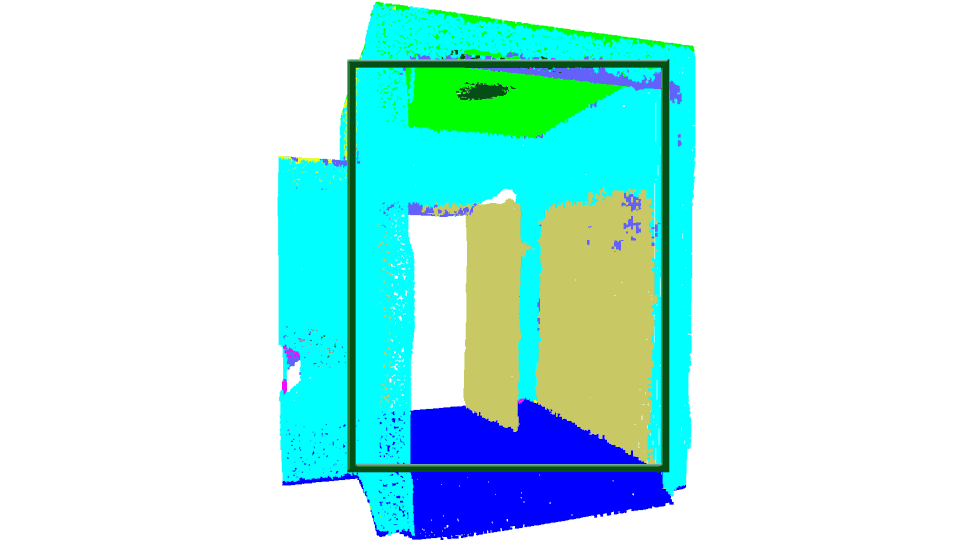}} \\
    \rotatebox[origin=c]{90}{conf.Room 1} \vspace{0.5cm} & \raisebox{-.5\height}{\includegraphics[width=0.23\textwidth]{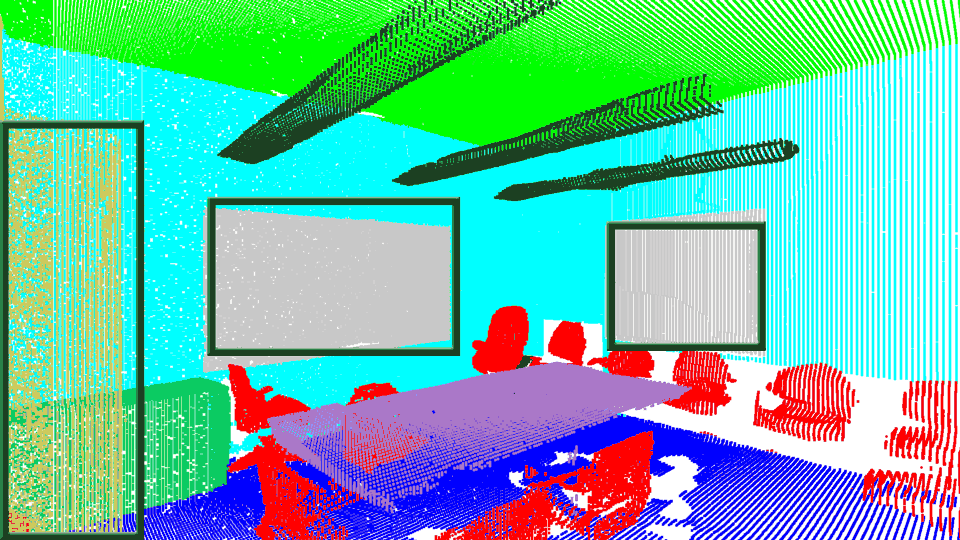}} & \raisebox{-.5\height}{\includegraphics[width=0.23\textwidth]{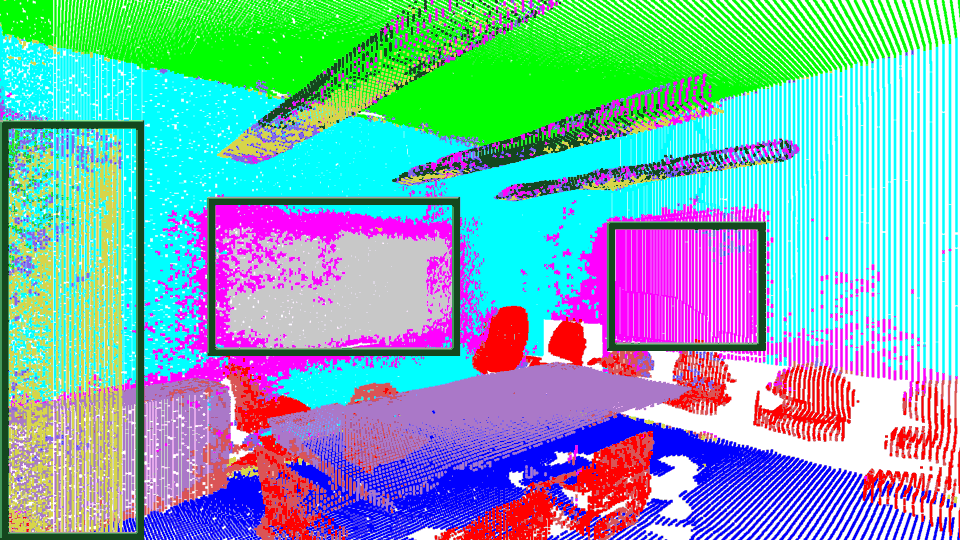}} & \raisebox{-.5\height}{\includegraphics[width=0.23\textwidth]{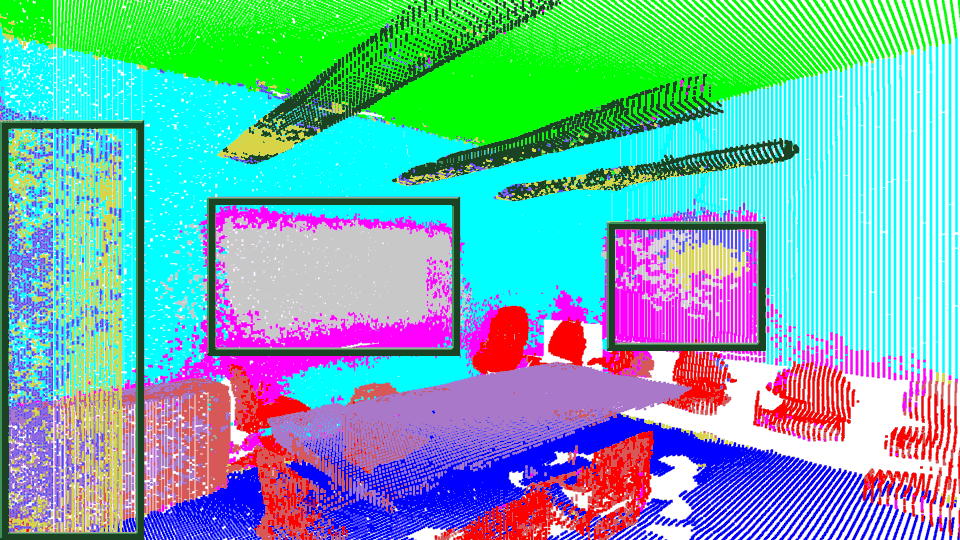}} & \raisebox{-.5\height}{\includegraphics[width=0.23\textwidth]{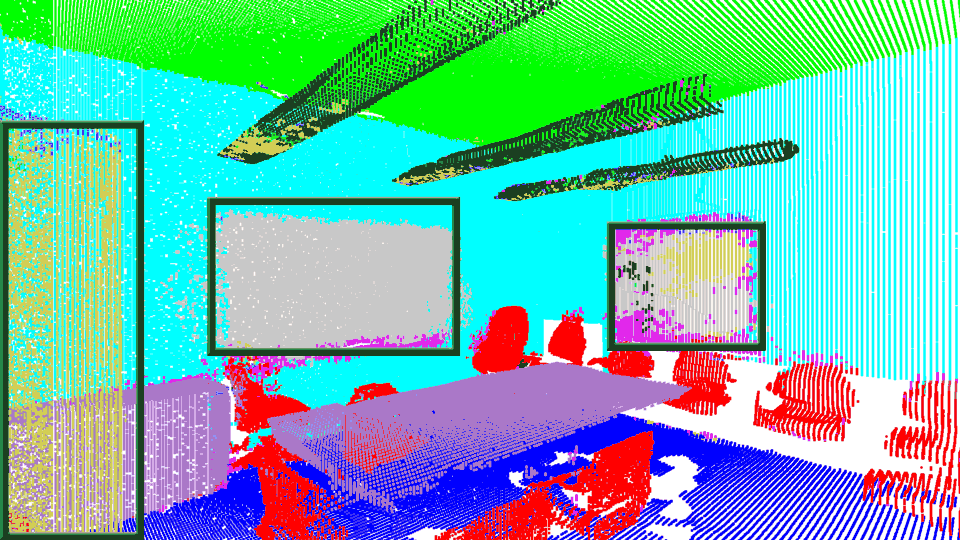}}\\
    \rotatebox[origin=c]{90}{office 5} \vspace{0.5cm} & \raisebox{-.5\height}{\includegraphics[width=0.23\textwidth]{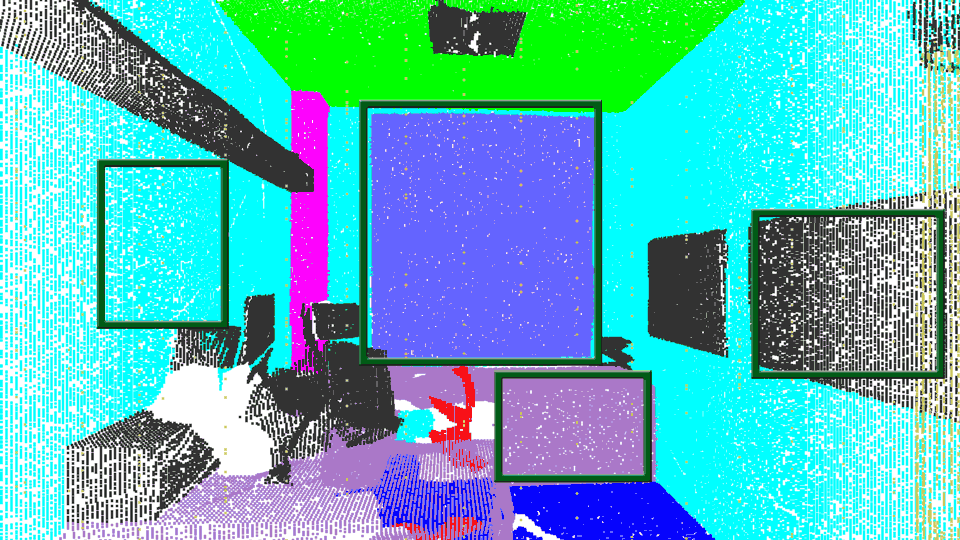}} & \raisebox{-.5\height}{\includegraphics[width=0.23\textwidth]{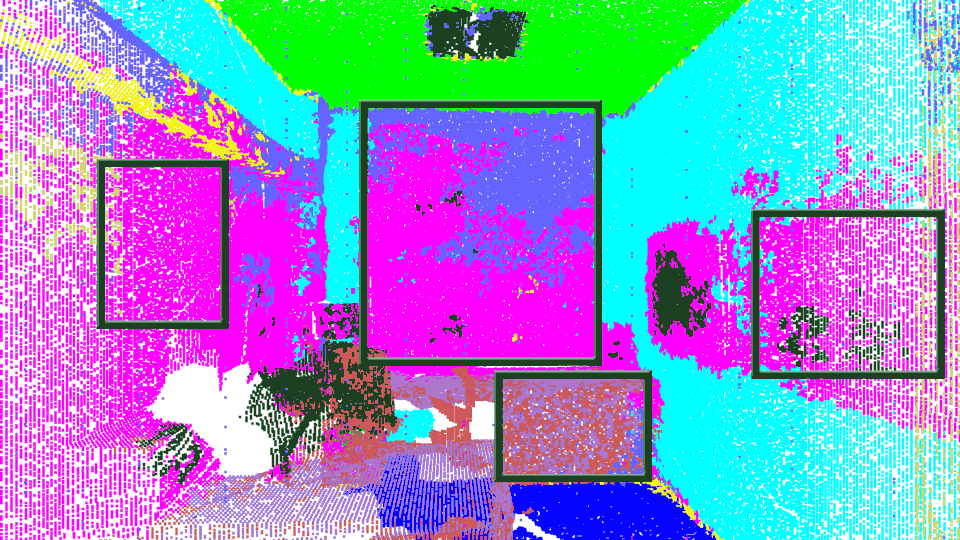}} & \raisebox{-.5\height}{\includegraphics[width=0.23\textwidth]{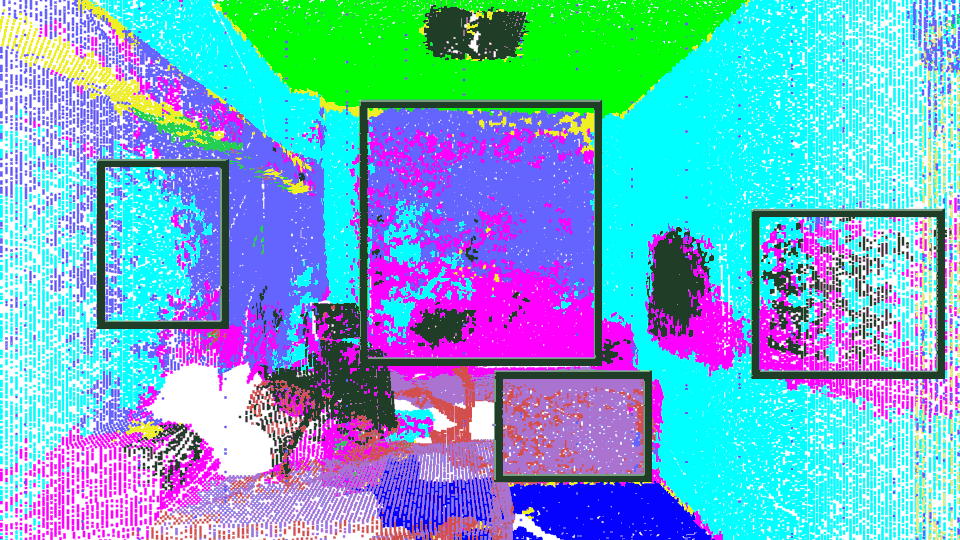}} & \raisebox{-.5\height}{\includegraphics[width=0.23\textwidth]{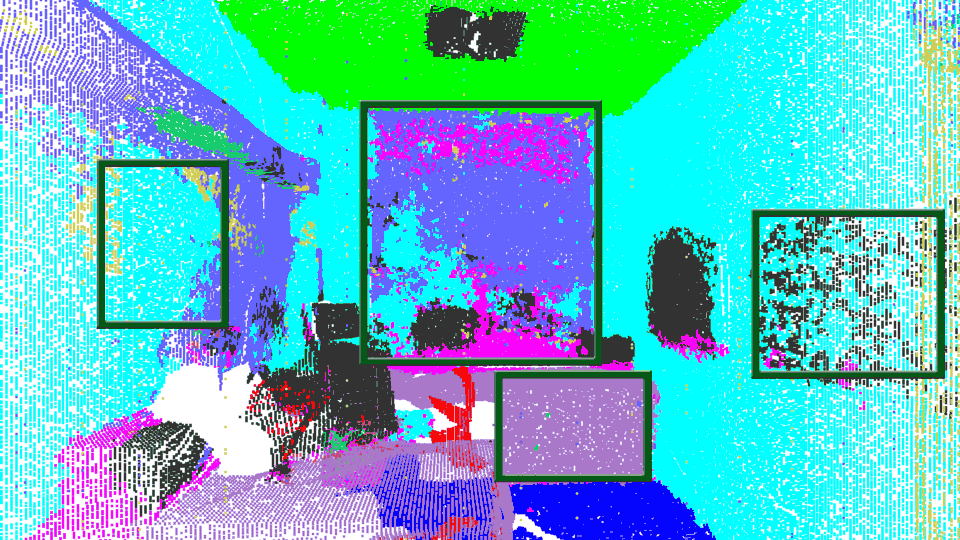}} \\
    \rotatebox[origin=c]{90}{office 9} \vspace{0.5cm} & \raisebox{-.5\height}{\includegraphics[width=0.23\textwidth]{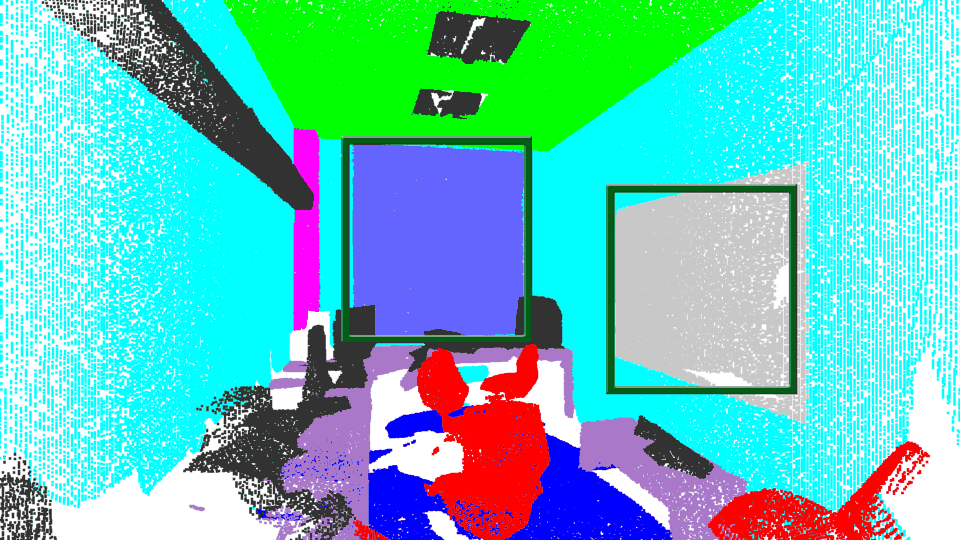}} & \raisebox{-.5\height}{\includegraphics[width=0.23\textwidth]{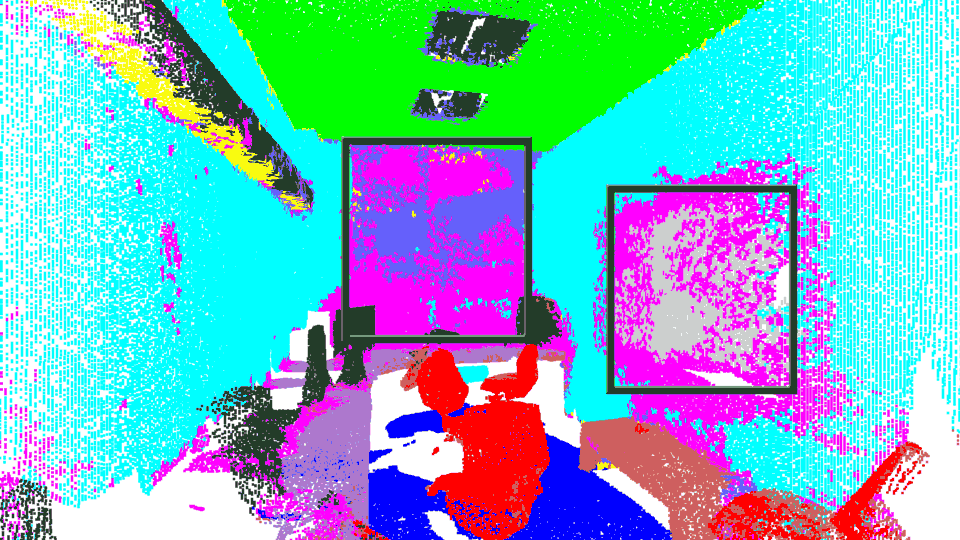}} & \raisebox{-.5\height}{\includegraphics[width=0.23\textwidth]{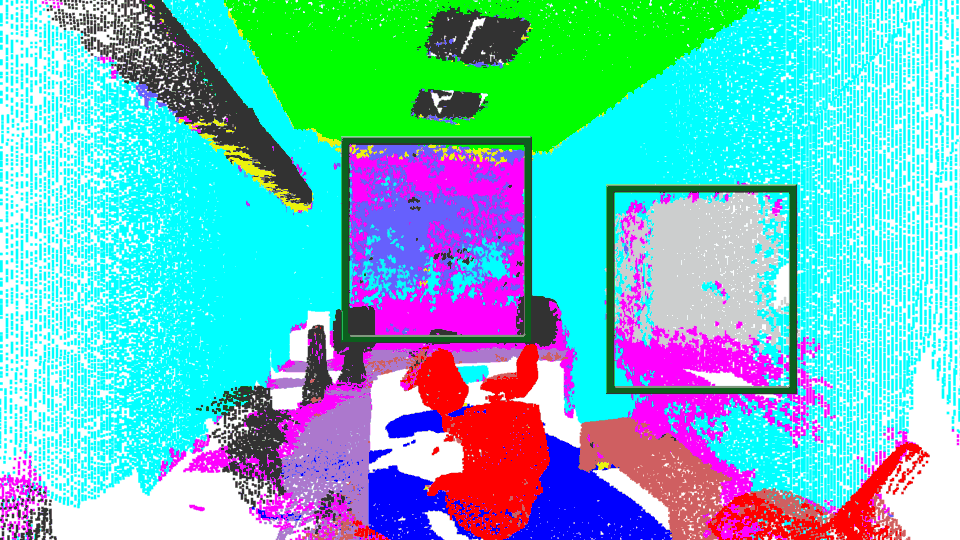}} & \raisebox{-.5\height}{\includegraphics[width=0.23\textwidth]{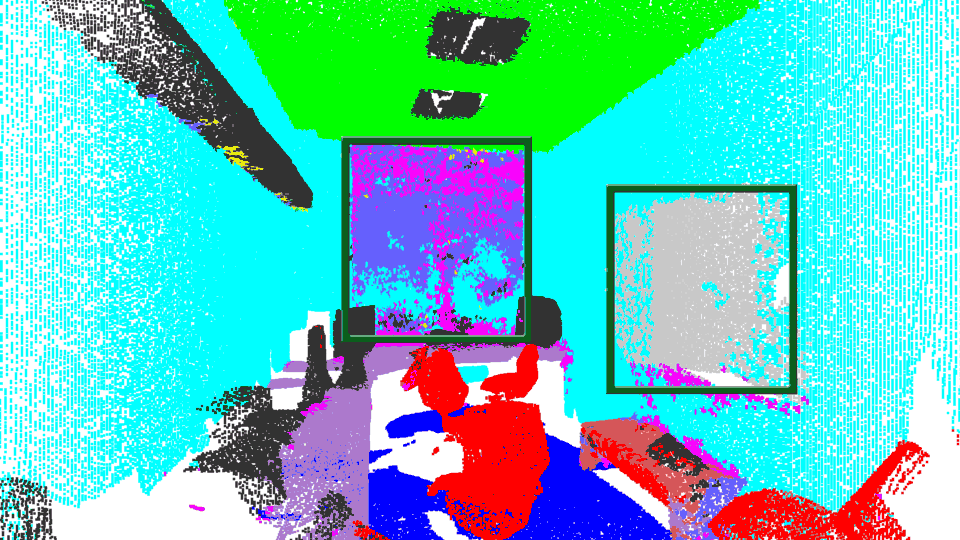}} \\
    \rotatebox[origin=c]{90}{office 11} \vspace{0.5cm} & \raisebox{-.5\height}{\includegraphics[width=0.23\textwidth]{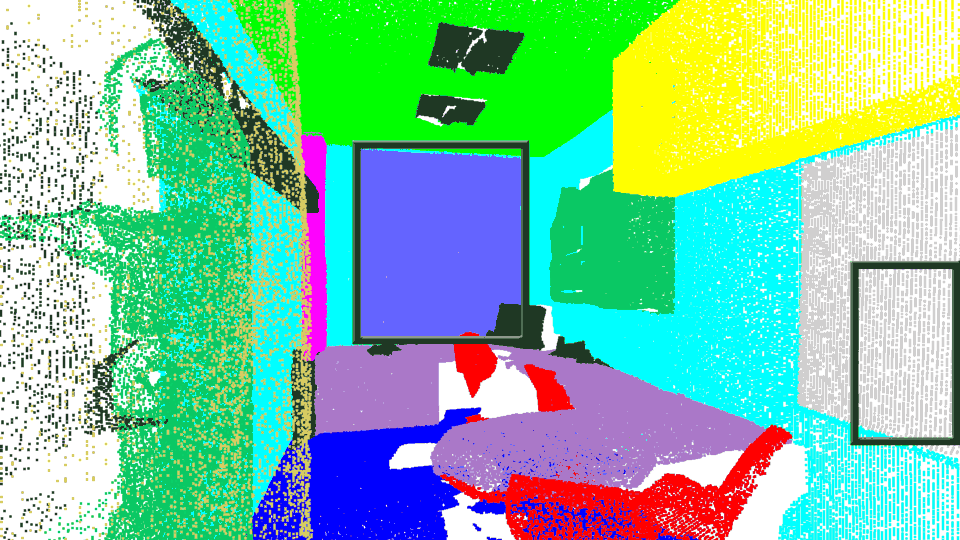}} & \raisebox{-.5\height}{\includegraphics[width=0.23\textwidth]{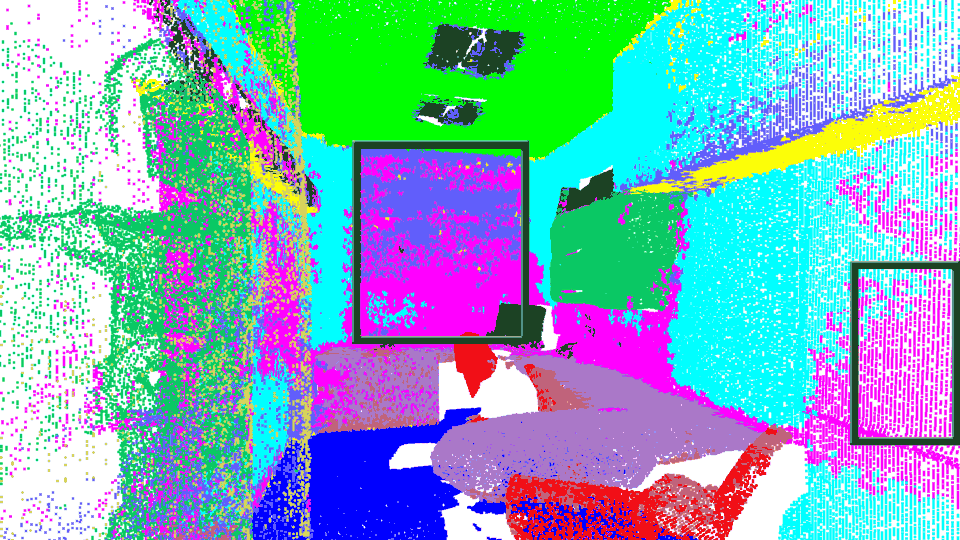}} & \raisebox{-.5\height}{\includegraphics[width=0.23\textwidth]{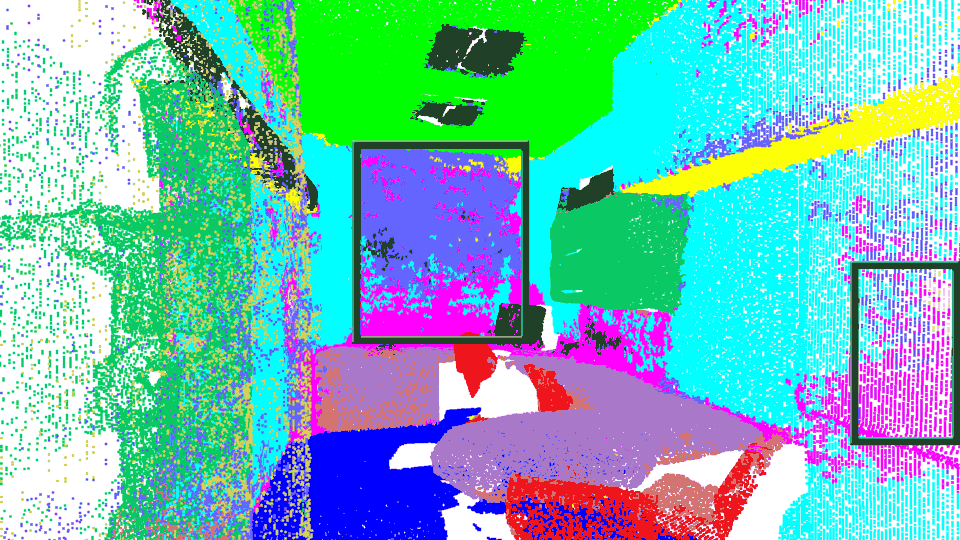}} & \raisebox{-.5\height}{\includegraphics[width=0.23\textwidth]{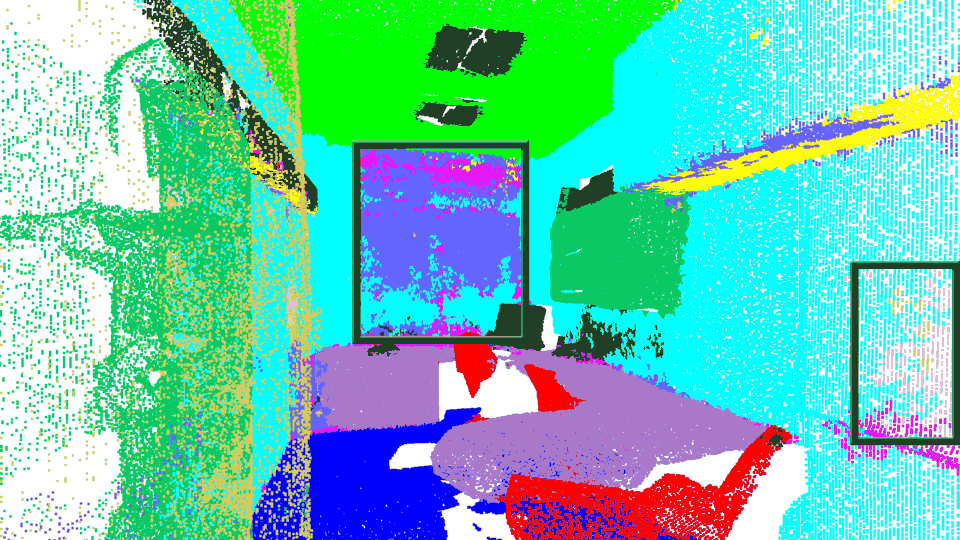}} \\
    \multicolumn{5}{c}{\includegraphics[width=0.7\textwidth]{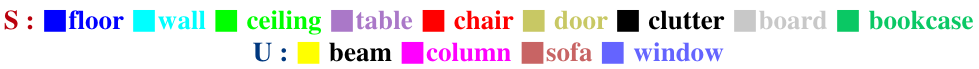}}\\ 
    \end{tabular}
    \caption{Qualitative comparison of the proposed model with other methods on S3DIS. The labels ``S'' in red denote seen classes, and ``U'' in blue denote unseen classes. } \label{fig:qualitative_comparison_S3DIS}
\end{figure*}

\clearpage

\noindent \textbf{Failure Cases.} \fref{fig:failure_case} presents representative failure cases observed in the ScanNet v2~\cite{c:304} and S3DIS~\cite{c:305} datasets. In the first columns of both datasets, we highlight examples where the model encounters difficulties in classifying unseen categories. While the proposed evidence-based dynamic calibration effectively detects unseen points, the final predictions still depend heavily on the classifier, revealing that calibration alone may be insufficient to resolve class ambiguity in certain scenarios.

The second column of ScanNet v2 and the second and third columns of S3DIS illustrate cases of semantic confusion caused by ambiguous object shapes. In ScanNet v2, which uses only geometric information (i.e., point cloud coordinates), the similar structures of the seen classes ``table'' and ``counter'' with the unseen class ``desk'' often lead to misclassification. Similarly, the ``door'' class is frequently confused with ``wall'' due to their comparable planar geometry. In S3DIS, despite the use of RGB information, visually similar categories such as ``column'' and ``beam'' remain difficult to distinguish from ``wall,'' indicating that geometric and color features alone may not fully resolve such ambiguities.

Finally, the third column of ScanNet v2 highlights a case of attribute-level confusion. The object ``shower curtain,'' which shares structural features with the ``curtain'' class, is inconsistently segmented—where the portion near the ``bathtub'' is correctly labeled as ``shower curtain,'' while the rest is misclassified as ``curtain.'' This reflects the challenge of fine-grained differentiation between semantically related classes, especially when only local context is considered.

\begin{figure*}[!h]
    \centering
    \begin{tabular}{@{~}c@{~}c@{~}c@{~}|c@{~}c@{~}c@{~}}
    \multicolumn{3}{c}{ScanNet v2} & \multicolumn{3}{c}{S3DIS} \\
     {\includegraphics[width=0.16\textwidth]{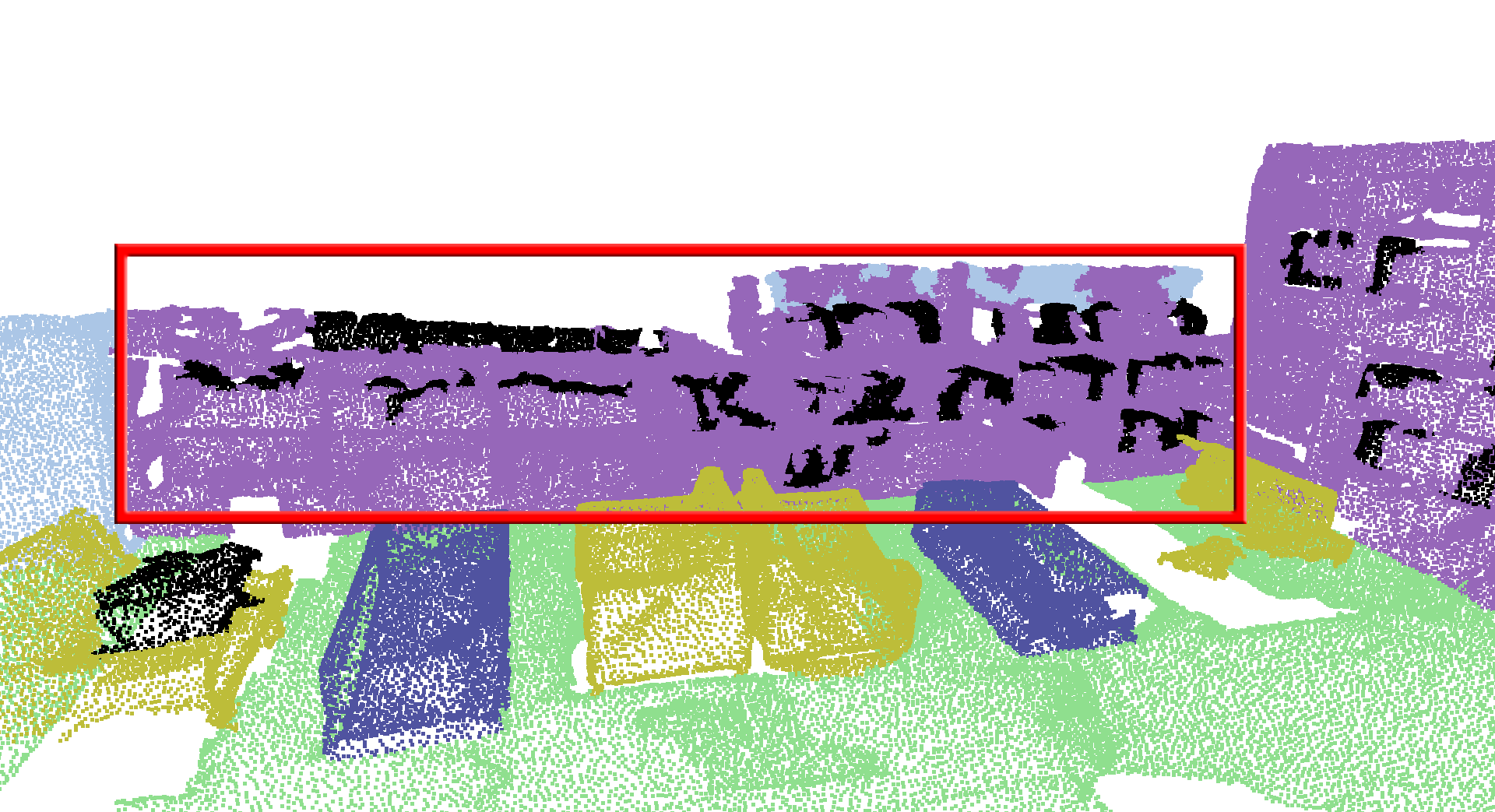}} & {\includegraphics[width=0.16\textwidth]{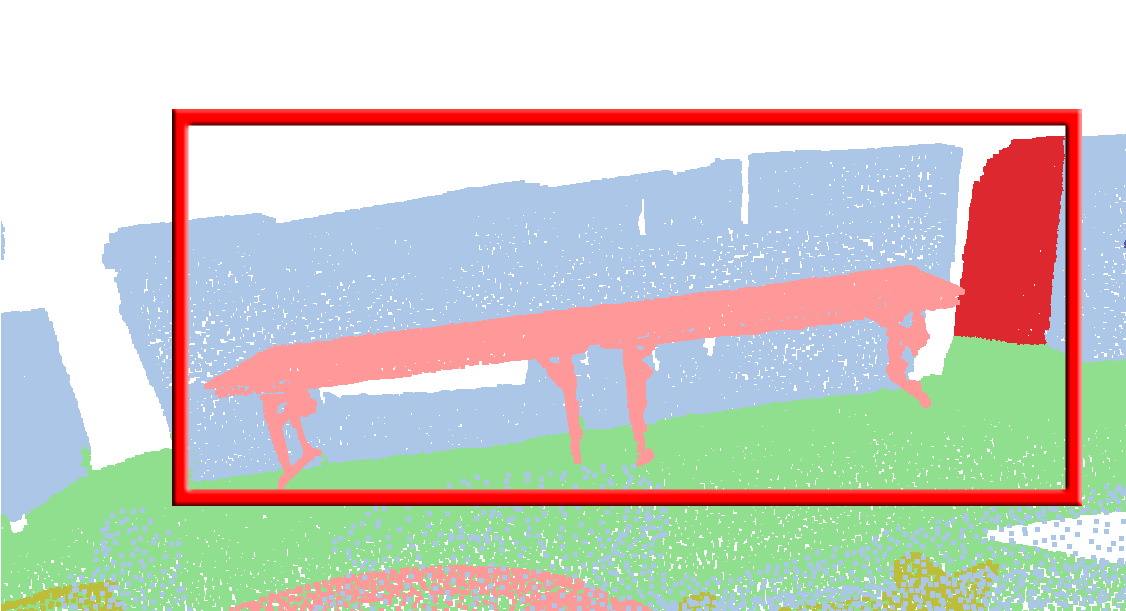}} & {\includegraphics[width=0.16\textwidth]{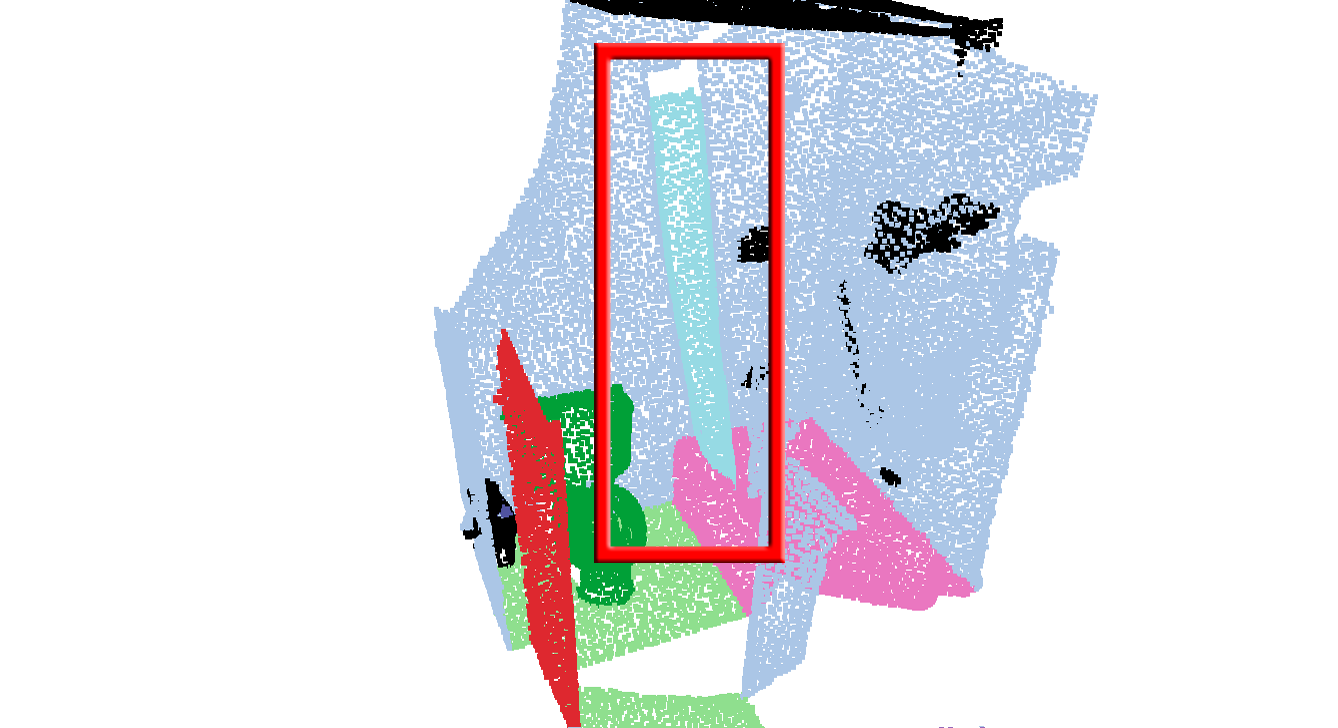}} &{\includegraphics[width=0.16\textwidth]{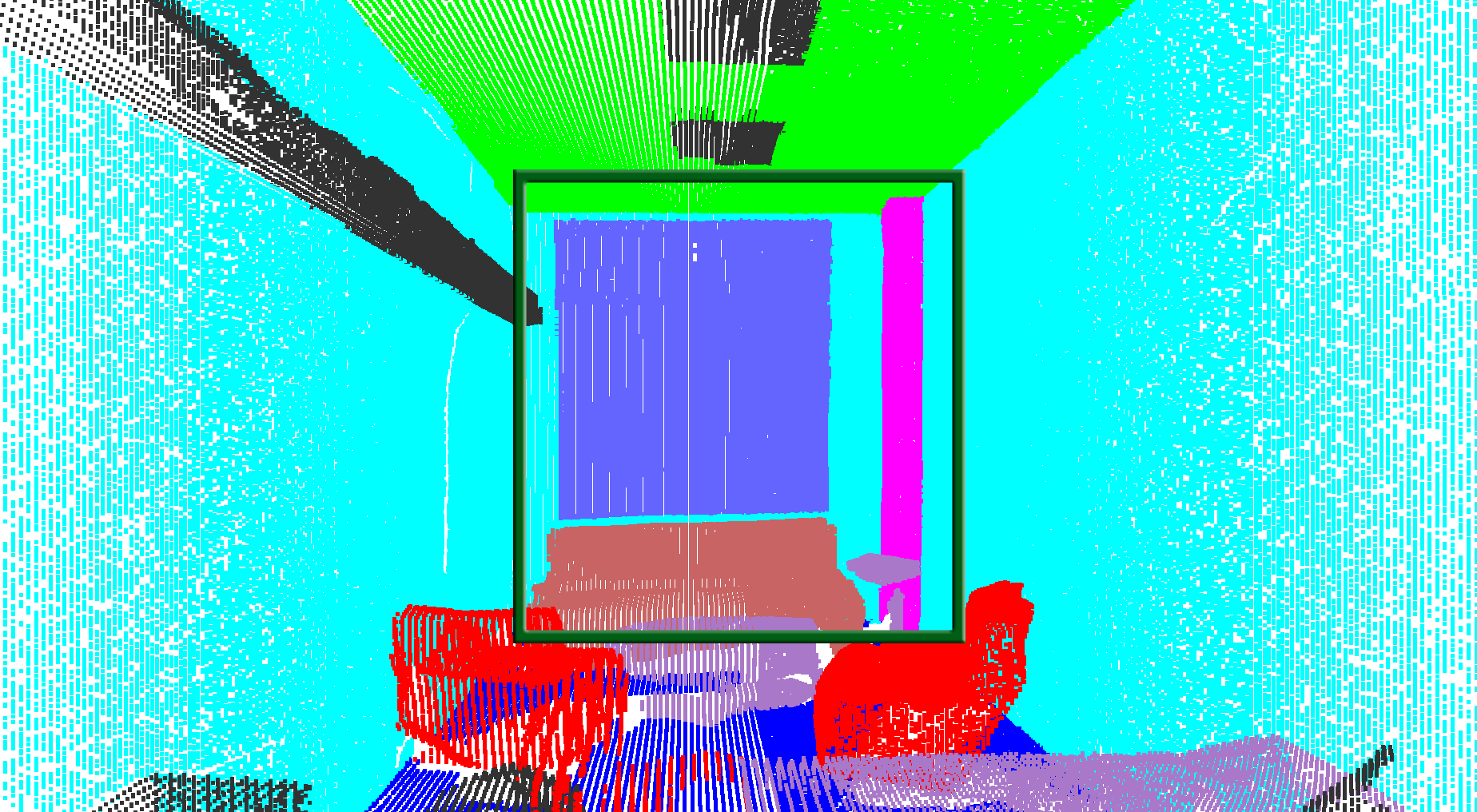}} & {\includegraphics[width=0.16\textwidth]{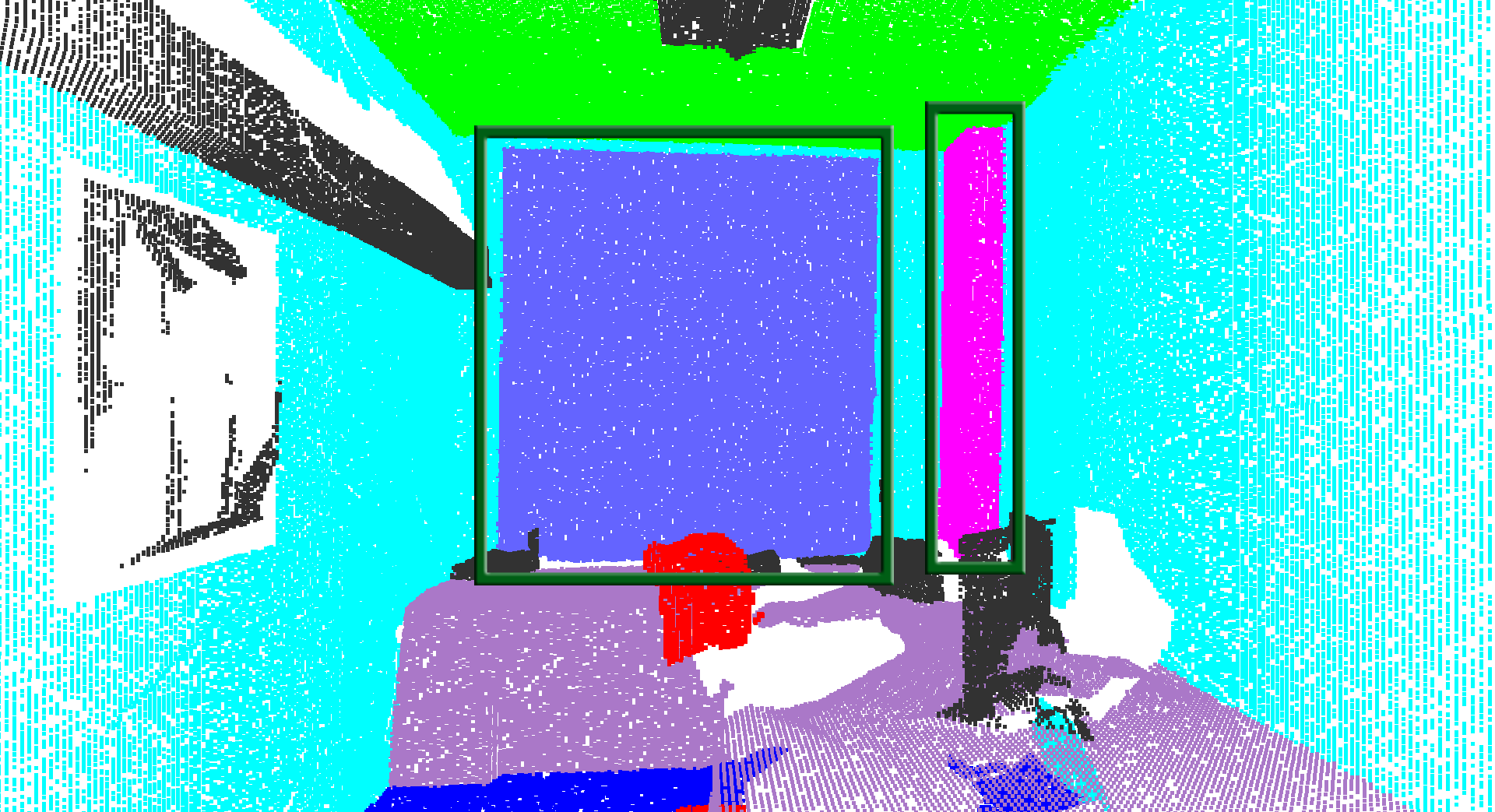}} & {\includegraphics[width=0.16\textwidth]{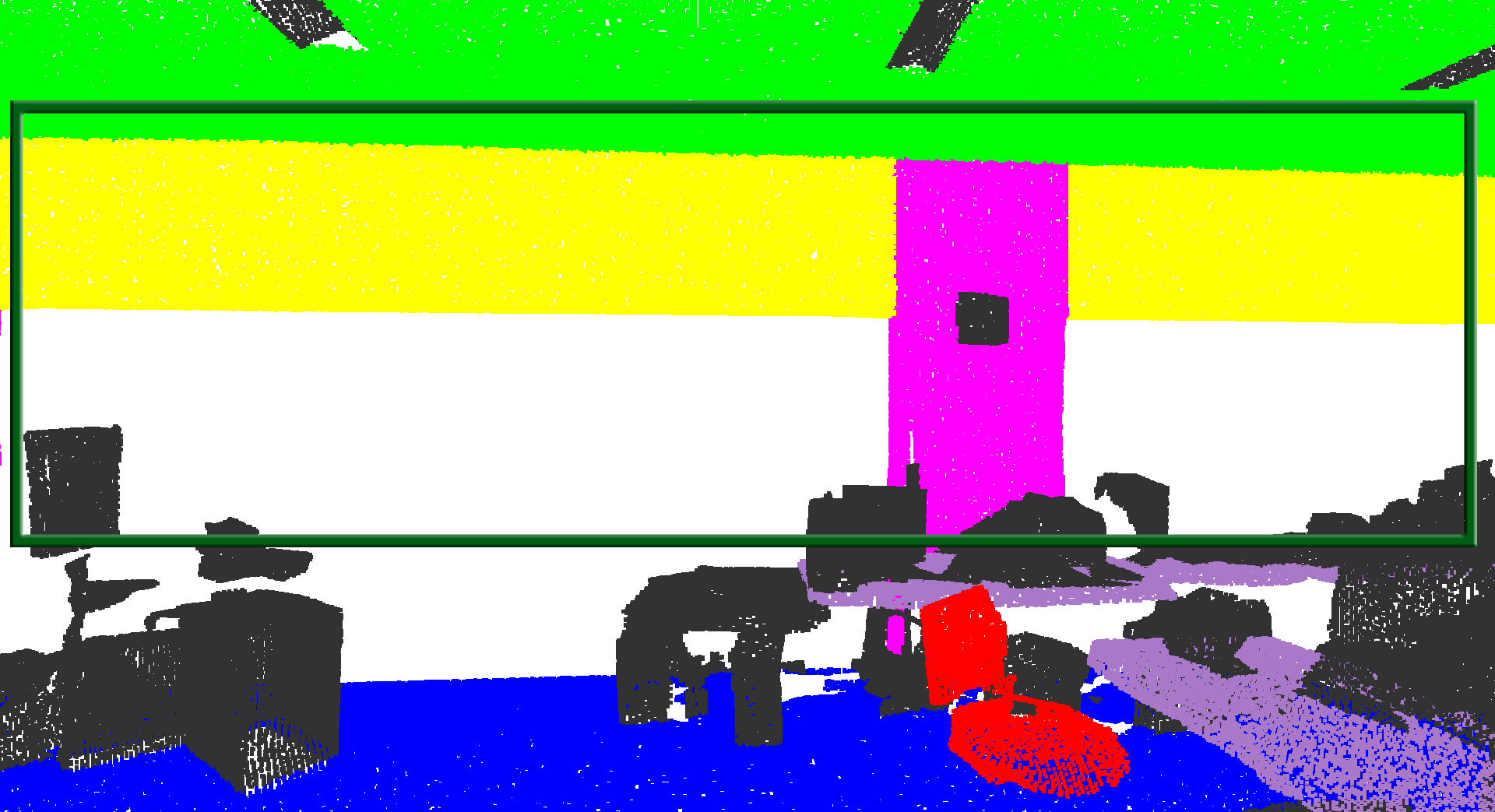}} \\ 
     {\includegraphics[width=0.16\textwidth]{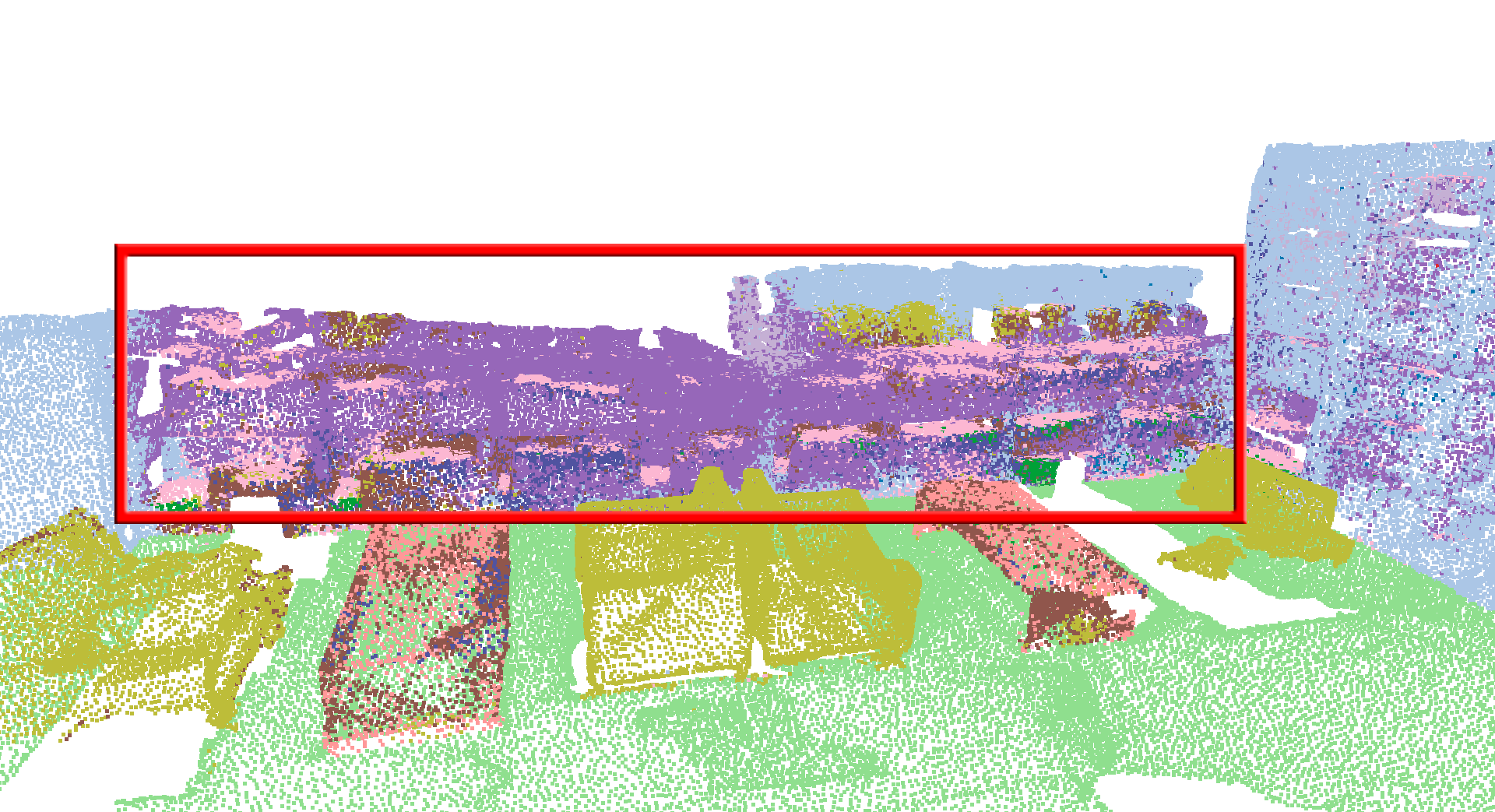}} & {\includegraphics[width=0.16\textwidth]{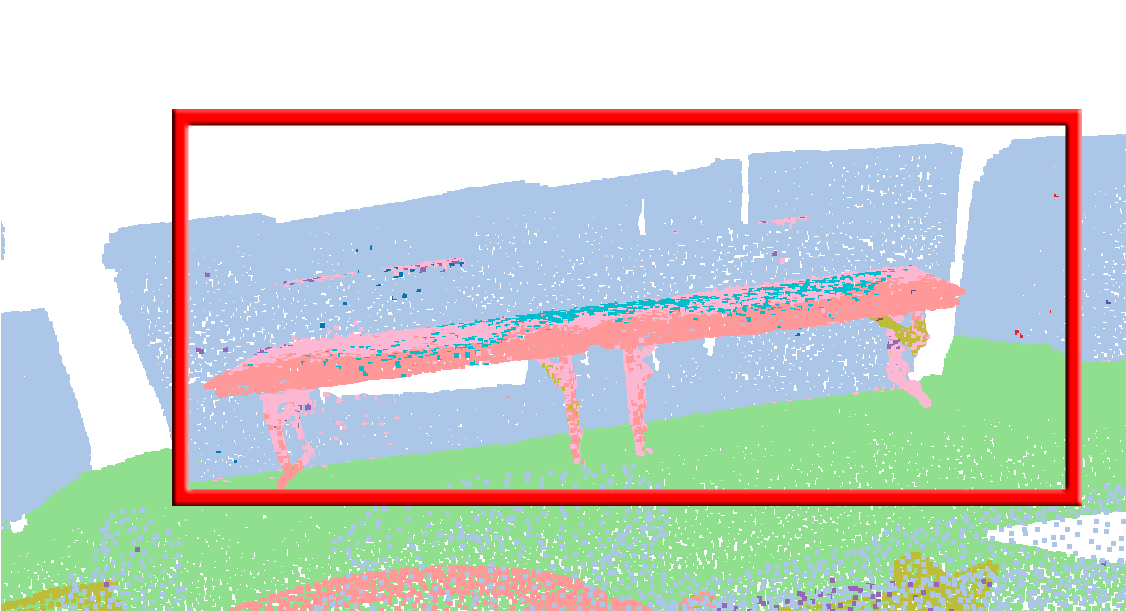}} & {\includegraphics[width=0.16\textwidth]{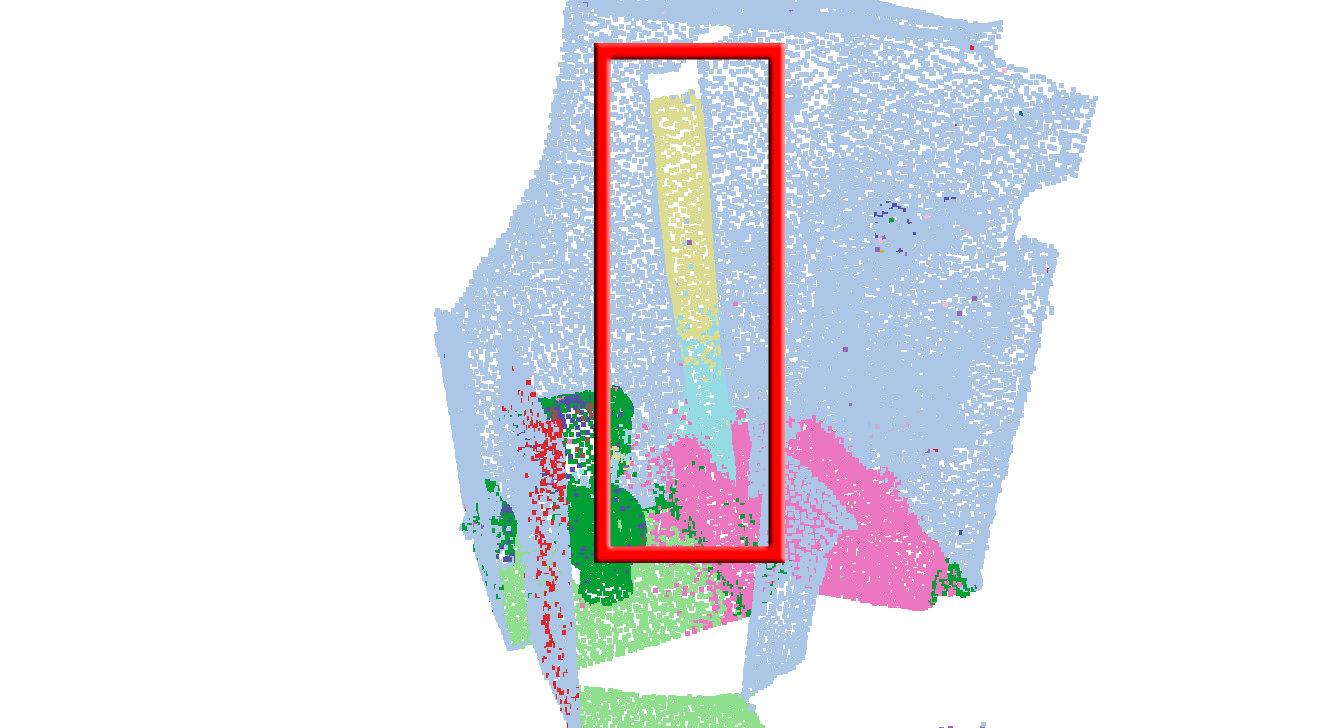}} & 
     {\includegraphics[width=0.16\textwidth]{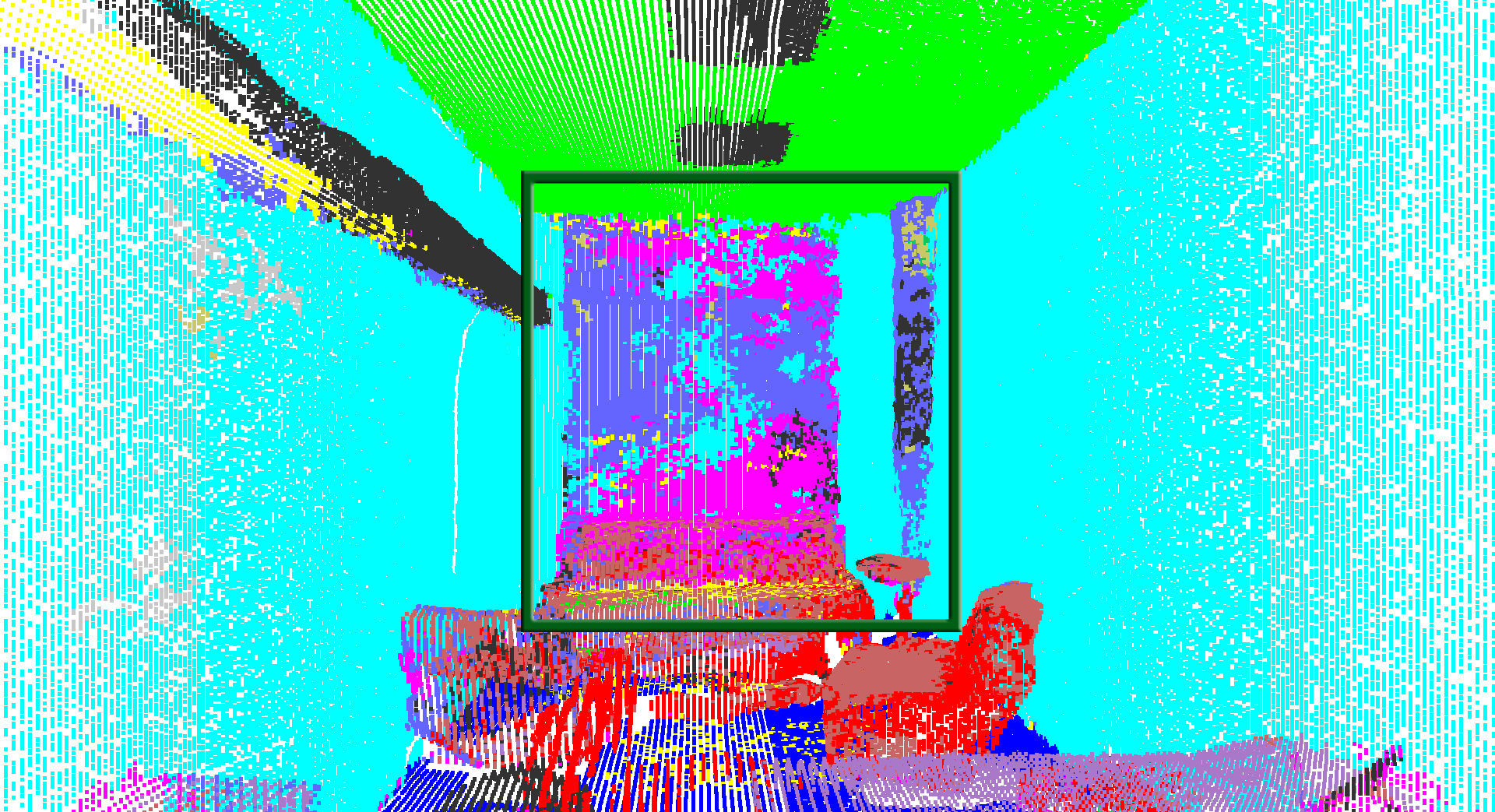}} & {\includegraphics[width=0.16\textwidth]{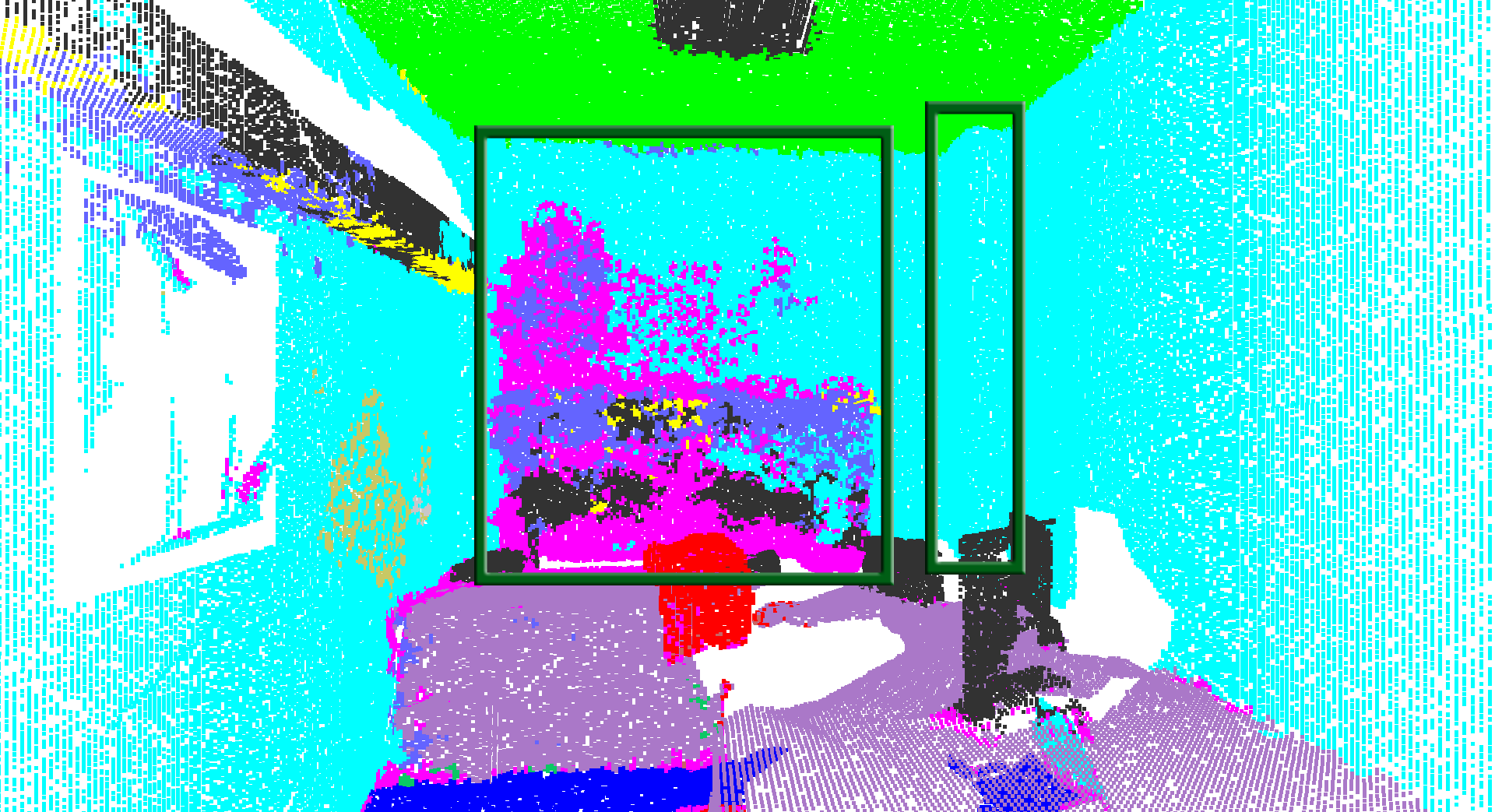}} & {\includegraphics[width=0.16\textwidth]{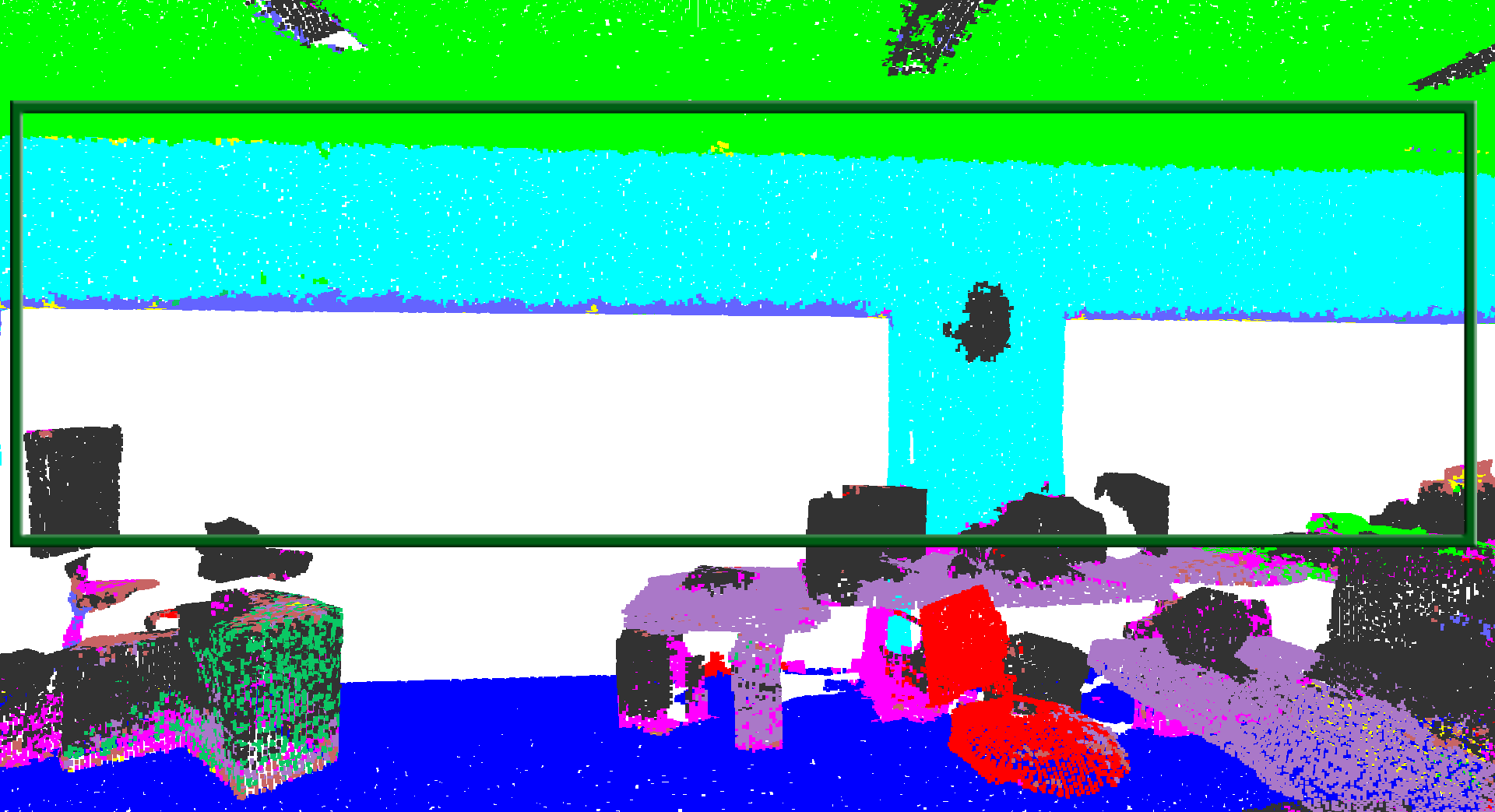}} \\
    \end{tabular} 
    \caption{Examples of failure cases on ScanNet v2 and S3DIS. The top row shows the ground truth, and the bottom row shows results with inaccurate segmentation. } \label{fig:failure_case}
\end{figure*}

\noindent \textbf{\Large C. Impact of Loss Weighting on Performance. }
\vspace{0.5cm}

We analyze the effect of the weighting coefficients $\lambda_{\text{DL}}$ and $\lambda_{\text{BL}}$ in Equation (\ref{eq:overall_loss}) of the manuscript on the ScanNet v2~\cite{c:304} and S3DIS~\cite{c:305} datasets. \fref{fig:parameters} illustrates the changes in seen and unseen mIoU with respect to different coefficient settings for each dataset.

For ScanNet v2, the model shows limited responsiveness to changes in the weighting coefficients, with seen mIoU exhibiting negligible variation. A slight improvement in unseen mIoU is observed when $\lambda_{\text{DL}} = 0.005$, and under this setting, the performance tends to improve as $\lambda_{\text{BL}}$ decreases.

In contrast, the S3DIS dataset exhibits a clearer sensitivity to the weighting coefficients. In general, increasing both $\lambda_{\text{DL}}$ and $\lambda_{\text{BL}}$ leads to improved performance. The highest accuracy is achieved when $\lambda_{\text{DL}} = 0.005$ and $\lambda_{\text{BL}} = 0.1$, indicating the importance of appropriately balancing these loss components for optimal generalization.

\begin{figure}[!h]
    \centering
    \begin{tabular}{cccc}
    \multicolumn{2}{c}{ScanNet v2} & \multicolumn{2}{c}{S3DIS} \\
    \includegraphics[width=0.21\textwidth]{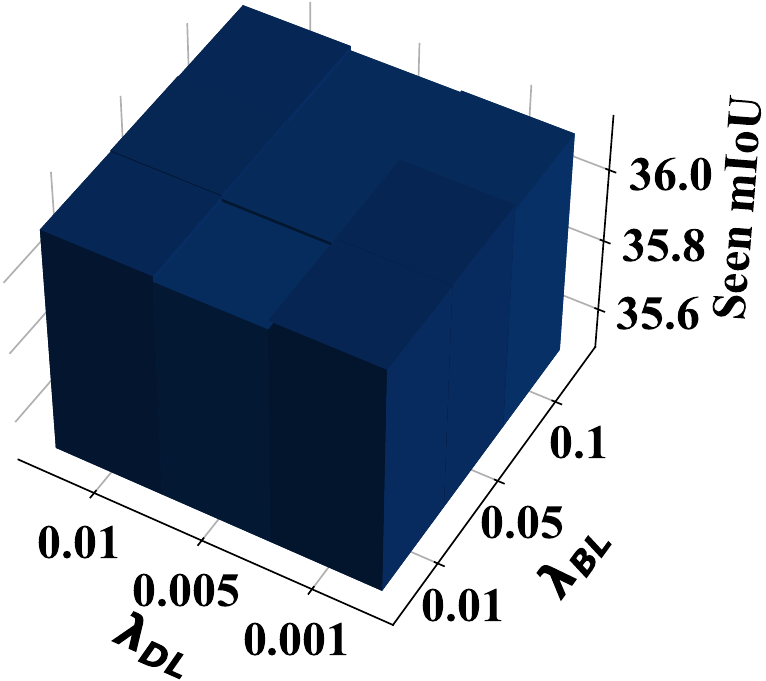} & \includegraphics[width=0.21\textwidth]{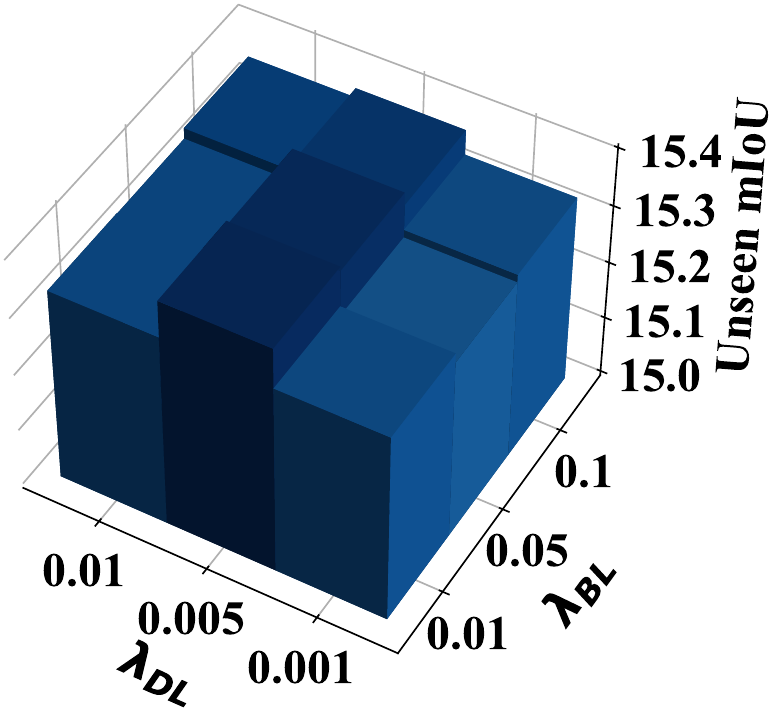} & \includegraphics[width=0.23\textwidth]{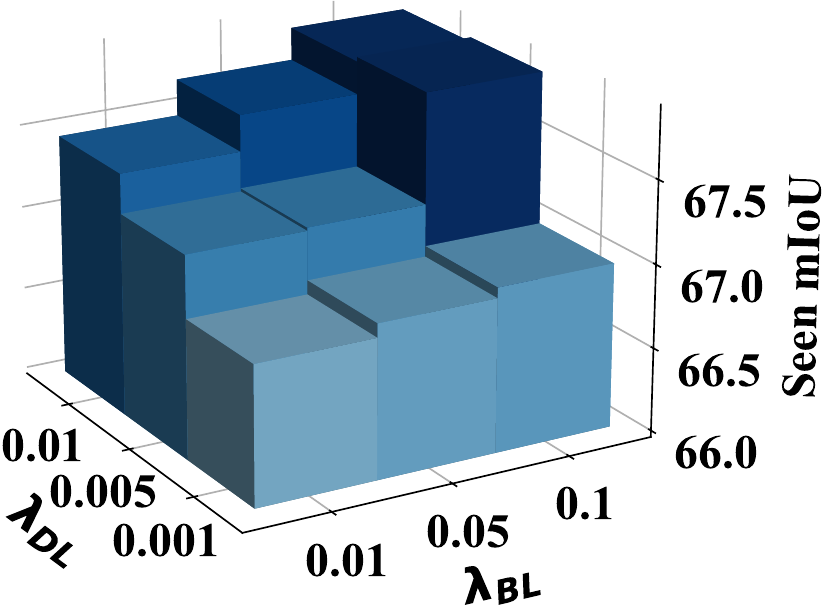} & \includegraphics[width=0.23\textwidth]{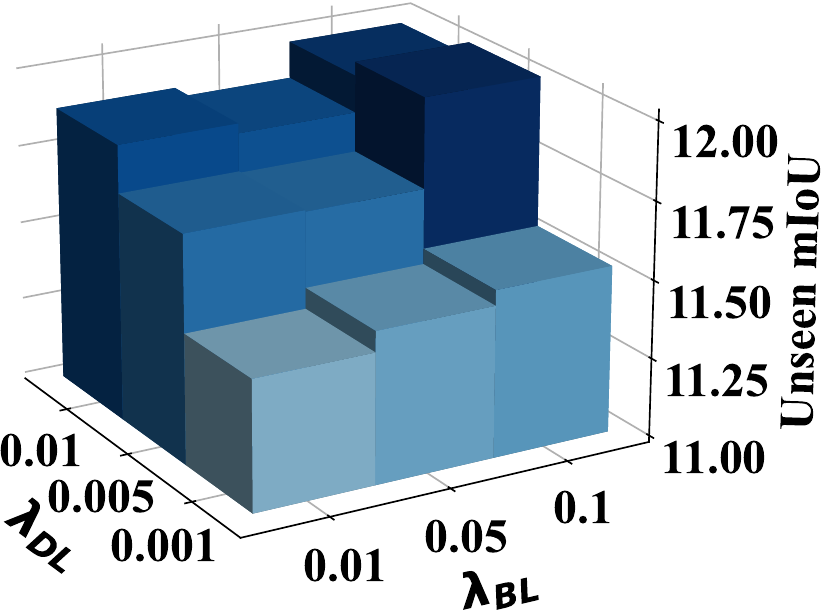} \\
    \end{tabular}
    \caption{Seen and unseen mIoU (\%) performance across different values of $\lambda_{DL}$ and $\lambda_{BL}$ } \label{fig:parameters}
\end{figure}
\clearpage

\noindent \textbf{\Large E. Comparison of Unseen Recognition}
\vspace{0.5cm}

To comprehensively assess the effectiveness of the proposed dynamic calibration, we compare the performance of the baseline model under varying $\eta$ values with that of our method, which incorporates dynamic calibration guided by uncertainty estimation. Specifically, we evaluate both F1 and mIoU scores across different $\eta$ settings for the baseline model, which does not include semantic tuning or uncertainty-aware calibration. The F1 score is computed as follows:
\begin{align}
    F1 = 2 \times \left(\frac{Precision \times Rrecall}{Precision + Recall}\right),
\end{align}
where precision and recall are computed as follows:
\begin{align}\begin{array}{ll}
    Precision  & = \displaystyle \frac{True\: Positive}{True\: Positive+False\: Positive}, \vspace{0.2cm}\\ 
    Recall  & = \displaystyle \frac{True\: Positive}{True\: Positive+False Negative}. \\
\end{array}\end{align}
\vspace{0.5cm}

\fref{fig:comparison_unseen_recognition} presents an analysis of segmentation performance on ScanNet v2~\cite{c:304} and S3DIS~\cite{c:305}. In each plot, the blue line indicates the performance trend across varying $\eta$ values, while the red dotted line marks the result of the proposed dynamic calibration as a reference. The results show that segmentation performance, measured by F1 and HmIoU, is sensitive to the choice of calibrated stacking factors. On S3DIS in particular, performance is notably influenced by the $\eta$ setting, highlighting the calibration’s impact. Similarly, on ScanNet v2, performance fluctuates across different $\eta$ values, underscoring the importance of selecting appropriate calibration parameters to avoid degradation in segmentation quality.

\begin{figure*}[!h]
    \centering
    \begin{tabular}{@{}cc@{}}
    ScanNet v2 \hspace{0.5cm} & S3DIS \\
    \raisebox{-.5\height}{\includegraphics[width=0.27\textwidth]{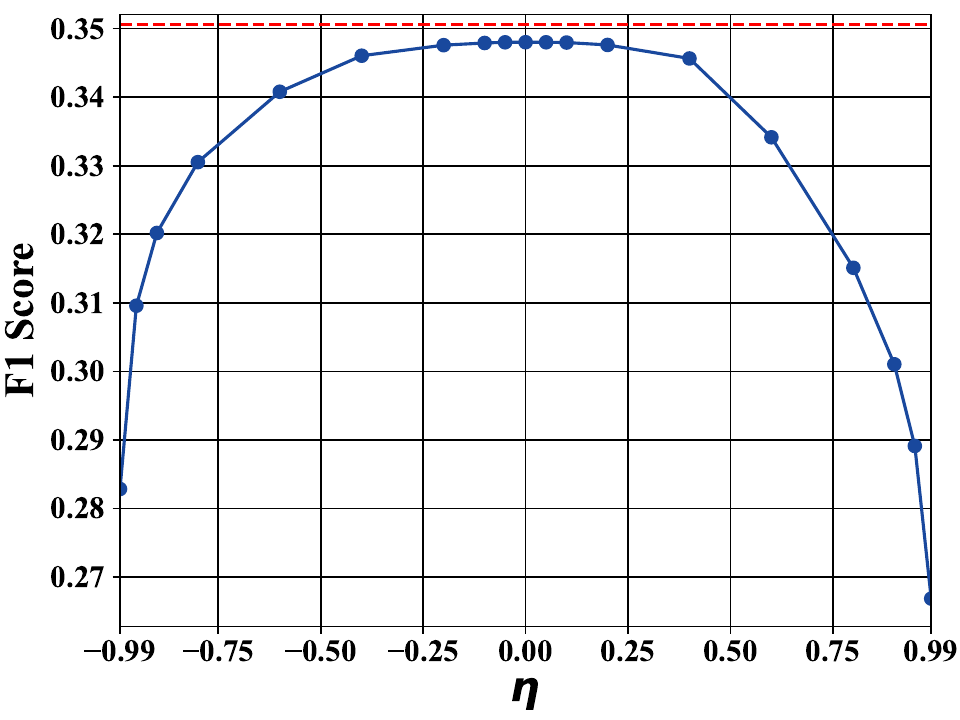}} \hspace{0.5cm} \vspace{0.1cm} & \raisebox{-.5\height}{\includegraphics[width=0.27\textwidth]{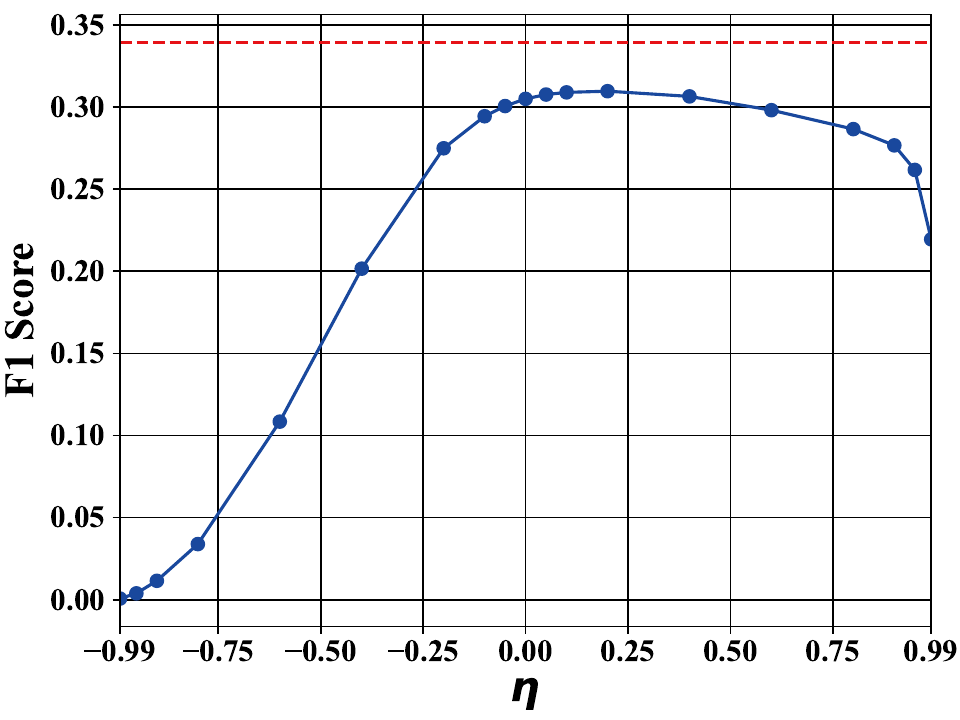}}  \vspace{0.1cm} \\ 
    \raisebox{-.5\height}{\includegraphics[width=0.27\textwidth]{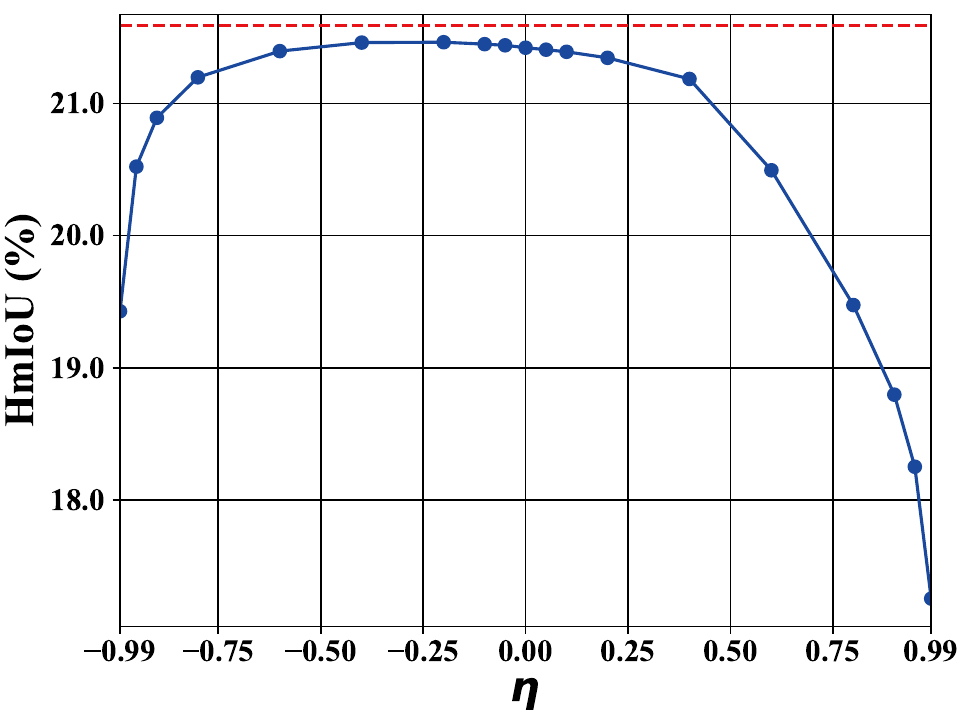}} \hspace{0.5cm} \vspace{0.1cm} & 
    \raisebox{-.5\height}{\includegraphics[width=0.27\textwidth]{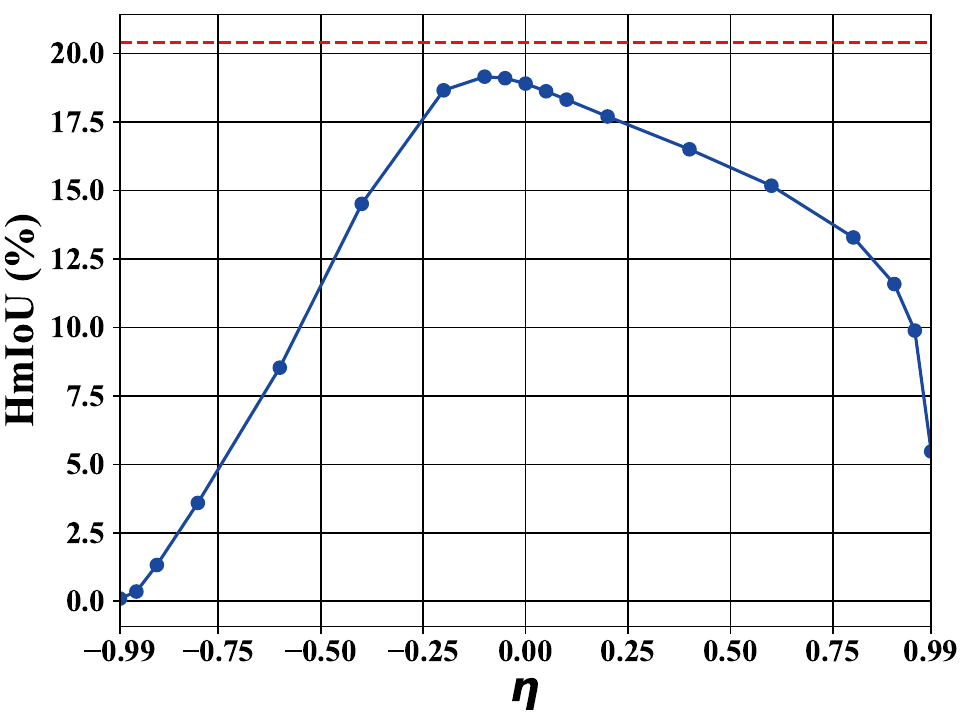}}  \vspace{0.1cm}\\ 
    \multicolumn{2}{c}{\includegraphics[width=0.35\textwidth]{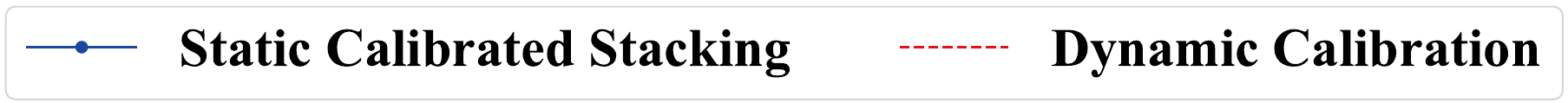}} \\ 
    \end{tabular}
    \caption{Variation in F1 score and HmIoU(\%) with different $\eta$ values.} \label{fig:comparison_unseen_recognition}
\end{figure*}

\clearpage

\noindent \textbf{\Large F. Outdoor Results}
\vspace{0.5cm}

In addition to indoor datasets such as ScanNet v2~\cite{c:304} and S3DIS~\cite{c:305}, we also evaluate our method on the outdoor LiDAR dataset SemanticKITTI~\cite{SemanticKITTI} to further validate its generalizability across diverse 3D environments.

\noindent \textbf{Datasets.} Consistent with other datasets~\cite{c:304, c:305}, the data splitting protocol for SemanticKITTI follows the settings established in prior works~\cite{c:225, c:226}. The SemanticKITTI~\cite{SemanticKITTI} dataset consists of $23,201$ point cloud scenes from $10$ outdoor sequences. Sequences 00–07, 09, and 10~($19,130$ scenes) are used for training with fifteen seen classes~($N_s = 15$), while Sequence 08~($4,071$ scenes) is designated as the evaluation dataset. This evaluation set includes four unseen classes~(bicyclist, motorcycle, traffic-sign, and truck, $N_u = 4$) and fifteen seen classes.

\noindent \textbf{Implementation Details.} Following the 3D GZSL semantic segmentation protocol~\cite{c:225, c:226}, we employ KPConv~\cite{c:214} as the encoder, resulting in a $128$-dimensional feature vector. The decoder, text embeddings, and uncertainty estimator components are consistent with those described in manuscript. For optimization, the model is trained using the SGD optimizer for $30$ epochs with a batch size of $8$. Point samples are generated using a subsampling cell size of $6 cm$, and the number of votes is set to $500$. The loss balancing coefficients are set to $\lambda_{DL}=0.5$ and $\lambda_{BL}=0.005$.

\noindent \textbf{Impact of Loss Weighting on Performance.} We further analyze the effect of the weighting coefficients $\lambda_{\text{DL}}$ and $\lambda_{\text{BL}}$ in Equation (\ref{eq:overall_loss}) on the SemanticKITTI~\cite{SemanticKITTI} dataset by presenting the changes in seen and unseen mIoU under different coefficient settings in \fref{fig:parameters_semanticKITTI}. Similar to the ScanNet v2~\cite{c:304} dataset, the seen mIoU on SemanticKITTI remains largely consistent across most configurations, except when $\lambda_{\text{DL}} = 0.5$ and $\lambda_{\text{BL}} = 0.01$. However, a notable difference from ScanNet v2 is that the seen mIoU approaches the performance obtained under a fully supervised setting, where the classifier is trained with both seen and unseen data, implying strong generalization under limited supervision. For unseen mIoU, we observe that when $\lambda_{\text{BL}} = 0.05$, performance tends to improve as $\lambda_{\text{DL}}$ decreases, indicating a negative correlation between $\lambda_{\text{DL}}$ and unseen-class segmentation performance under this setting.

\begin{figure}[!h]
    \centering
    \begin{tabular}{cc}
    \multicolumn{2}{c}{SemanticKITTI}\\
    \includegraphics[width=0.21\textwidth]{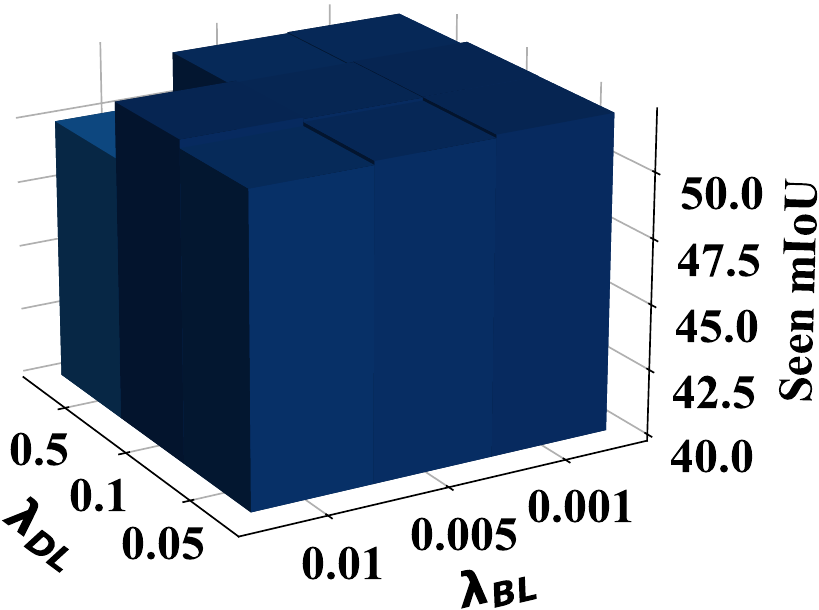} & \includegraphics[width=0.19\textwidth]{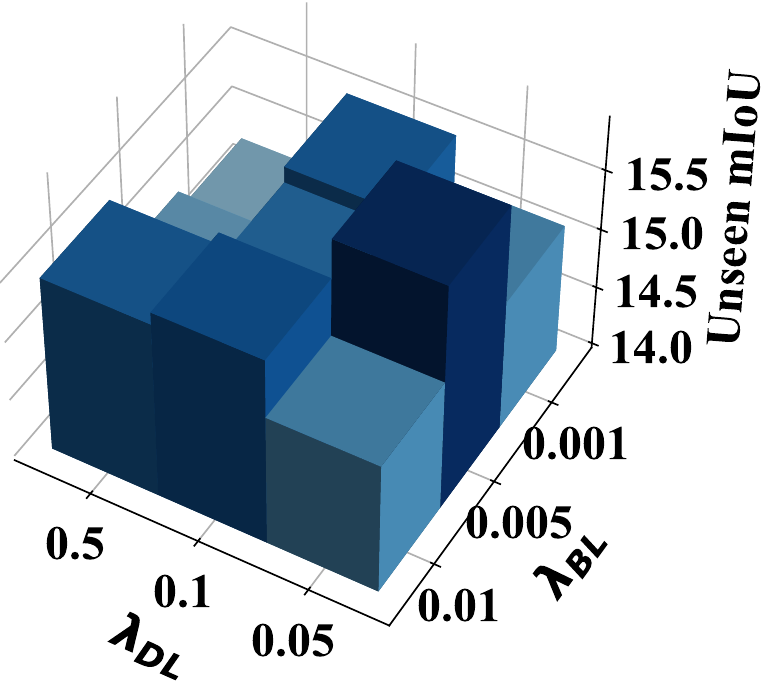} \\
    \end{tabular}
    \caption{Seen and unseen mIoU (\%) performance across different values of $\lambda_{DL}$ and $\lambda_{BL}$ on SemanticKITTI dataset.} \label{fig:parameters_semanticKITTI}
\end{figure}

\noindent \textbf{Experimental Results.} As in the main manuscript, we conduct a comprehensive performance comparison on the SemanticKITTI~\cite{SemanticKITTI} dataset against prior works~\cite{c:225, c:226}, as shown in Table~\ref{tab:SemanticKITTI_results}. The proposed E3DPC-GZSL method consistently outperforms existing approaches across all key metrics, achieving state-of-the-art(SOTA) results in seen, unseen mIoU, and harmonic mean IoU (HmIoU). Specifically, E3DPC-GZSL improves the seen mIoU by 9.4\% and 4.4\% over 3DGenZ~\cite{c:225} and 3DPC-GZSL~\cite{c:226}, respectively. For unseen mIoU, it achieves gains of 3.2\% and 1.2\% compared to 3DGenZ and 3DPC-GZSL. Furthermore, with respect to the critical generalized zero-shot learning metric, HmIoU, E3DPC-GZSL surpasses 3DGenZ by 4.8\% and 3DPC-GZSL by 1.8\%.

\begin{table}[!h]
\centering
  \setlength{\tabcolsep}{4.0mm}
  \begin{tabular}{c|c|c|ccc|c}
    \toprule
     & \multicolumn{2}{c|}{Training set} & \multicolumn{4}{c}{SemanticKITTI} \\ \cline{2-7}
     & \multirow{2}{*}{Encoder} & \multirow{2}{*}{Classifier} & \multicolumn{3}{c|}{mIoU} & \multirow{2}{*}{HmIoU} \\ \cline{4-6} 
     & & & Seen & Unseen & All &\\
     \midrule
     Full supervision & $\mathcal{Y}^{s} \cup \mathcal{Y}^{u}$ & $\mathcal{Y}^{s} \cup \mathcal{Y}^{u}$ & 59.4 & 50.3 & 57.5 & 54.5 \\
     Full supervision only for classifier & $\mathcal{Y}^{s}$ & $\mathcal{Y}^{s} \cup \mathcal{Y}^{u}$ & 52.9 & 13.2 & 42.3 & 21.2 \\
     Supervision with seen & $\mathcal{Y}^{s}$ & $\mathcal{Y}^{s}$ & 55.8 & 0.0 & 44.0 & 0.0 \\
     \midrule
     3DGenZ \cite{c:225} & $\mathcal{Y}^{s}$ & $\mathcal{Y}^{s} \cup \mathcal{Y}^{\tilde{u}}$ & 41.4 & 10.8 & 35.0 & 17.1 \\
     3DPC-GZSL \cite{c:226} & $\mathcal{Y}^{s}$ & $\mathcal{Y}^{s} \cup \mathcal{Y}^{\tilde{u}}$ & 46.4 & 12.8 & 39.4 & 20.1 \\
     \midrule
     \textbf{E3DPC-GZSL (ours)} & $\mathcal{Y}^{s}$ & $\mathcal{Y}^{s} \cup \mathcal{Y}^{\tilde{u}}$ & \textbf{50.8} & \textbf{14.0} & \textbf{43.0} & \textbf{21.9} \\
    \bottomrule
\end{tabular} 
\caption{Performance Comparisons of 3D GZSL semantic segmentation benchmarks in terms of mIoU(\%) and HmIoU(\%) on SemanticKITTI dataset.}
\label{tab:SemanticKITTI_results}
\end{table}

\clearpage

\noindent \textbf{Per-Class Semantic Segmentation Results.} Table \ref{tab:SemanticKITTI_classwise} presents the class-wise semantic segmentation results on the SemanticKITTI~\cite{SemanticKITTI} dataset. Our method successfully recognizes the ``other-ground'' and ``pole'' classes within the seen categories, which were previously undetected by prior approaches~\cite{c:225, c:226}. Additionally, for the ``other-vehicle'' class, our method achieves a substantial improvement of +18.1 IoU over 3DPC-GZSL~\cite{c:226}. Regarding the unseen classes, the ``traffic-sign'' class shows an increase of +3.3 IoU compared to 3DGenZ and +2.7 IoU compared to 3DPC-GZSL. Similarly, for the ``truck'' class, our method outperforms 3DGenZ~\cite{c:225} by +9.5 IoU and 3DPC-GZSL by +5.5 IoU.

\begin{table*}[h]
\centering
  \setlength{\tabcolsep}{0.7mm}
  \begin{tabular}{l|c|ccccccccccccccc|cccc}
    \toprule
    & & \multicolumn{15}{c|}{Seen classes} & \multicolumn{4}{c}{Unseen classes} \\
    \cline{3-21}
    SemanticKITTI& HmIoU& \rotatebox[origin=c]{90}{bicycle} & \rotatebox[origin=c]{90}{building} & \rotatebox[origin=c]{90}{car} & \rotatebox[origin=c]{90}{fence} & \rotatebox[origin=c]{90}{motorcyclist} & \rotatebox[origin=c]{90}{other ground} & \rotatebox[origin=c]{90}{other vehicle} & \rotatebox[origin=c]{90}{parking} & \rotatebox[origin=c]{90}{person} & \rotatebox[origin=c]{90}{pole} & \rotatebox[origin=c]{90}{road} & \rotatebox[origin=c]{90}{sidewalk} & \rotatebox[origin=c]{90}{terrain} & \rotatebox[origin=c]{90}{trunk} & \rotatebox[origin=c]{90}{vegetation} & \rotatebox[origin=c]{90}{bicyclist} & \rotatebox[origin=c]{90}{motorcycle} & \rotatebox[origin=c]{90}{traffic sign} & \rotatebox[origin=c]{90}{truck}\\
    \midrule
    Full-Sup & 54.5 & 42.0 & 88.6 & 93.6 & 65.8 & 0.0 & 2.7 & 41.1 & 28.9 & 69.7 & 63.7 & 89.4 & 77.1 & 70.5 & 70.7 & 87.5 & 74.4 & 58.6 & 26.7 & 41.6\\
    \midrule
    3DGenZ & 17.1 & 0.0 & 87.3 & 86.9 & \textbf{61.8} & 0.0 & 0.0 & 0.0 & 18.6 & 0.0 & 0.0 & 88.8 & \textbf{78.6} & 73.6 & 38.2 & 87.8 & \textbf{28.0} & 11.5 & 0.9 & 2.6\\
    3DPC-GZSL & 20.1 & 0.0 & \textbf{89.1} & 91.7 & 61.6 & 0.0 & 0.0 & 26.9 & 26.7 & 0.0 & 0.0 & \textbf{89.5} & 77.8 & 73.8 & \textbf{71.3} & \textbf{88.2} & 26.8 & \textbf{16.4} & 1.5 & 6.6 \\
    \midrule
    \textbf{Ours} &  \textbf{21.9} & 0.0 & 89.0 & \textbf{93.0} & 57.9 & 0.0 & \textbf{1.6} & \textbf{45.0} & \textbf{30.3} & 0.0 & \textbf{46.9} & 89.4 & 75.9 & \textbf{75.4} & 69.4 & 88.1 & 27.7 & 12.0 & \textbf{4.2} & \textbf{12.1} \\
  \bottomrule
\end{tabular} 
\caption{Class-wise Performance Comparisons of 3D GZSL Semantic Segmentation Benchmarks in terms of mIoU(\%) and HmIoU(\%) on SemanticKITTI dataset.}
\label{tab:SemanticKITTI_classwise} 
\end{table*}

\noindent \textbf{Per-Class Semantic Segmentation Results.} \fref{fig:qualitative_comparison_SemanticKITTI} illustrates qualitative comparisons with prior works~\cite{c:225, c:226} on the SemanticKITTI~\cite{SemanticKITTI} dataset. Existing methods tend to misclassify seen classes such as ``car'' or ``other-vehicle'' as the unseen class ``truck''. Similarly, ``pole'', a seen class, is often incorrectly predicted as the unseen class ``traffic-sign''. In contrast, the proposed E3DPC-GZSL effectively mitigates such confusion, resulting in overall improved segmentation for both seen and unseen classes. Notably, our method accurately segments the ``trunk'' class in scene 08-0000000 and successfully identifies the ``parking'' class in scene 08-0001040, which prior methods struggled to detect. Additionally, in scene 08-0003700, our approach demonstrates enhanced recognition of the seen ``fence'' class. For the unseen ``bicyclist'' class in scene 08-0001680, prior works produce mixed predictions with ``motorcycle'', whereas E3DPC-GZSL provides more accurate segmentation.

\begin{figure*}[!t]
    \centering
    \begin{tabular}{@{}c@{}c@{}c@{}c@{}c@{}}
    & Ground-truth & 3DGenZ & 3DPC-GZSL & E3DPC-GZSL~(Ours) \vspace{0.2cm}\\
    \rotatebox[origin=c]{90}{08-0000000} \vspace{0.5cm} & \raisebox{-.5\height}{\includegraphics[width=0.23\textwidth]{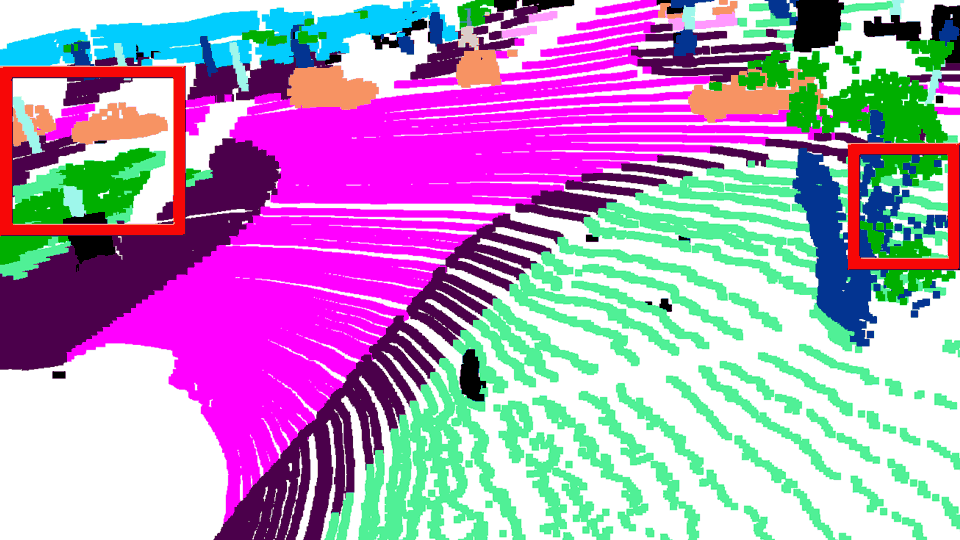}} & \raisebox{-.5\height}{\includegraphics[width=0.23\textwidth]{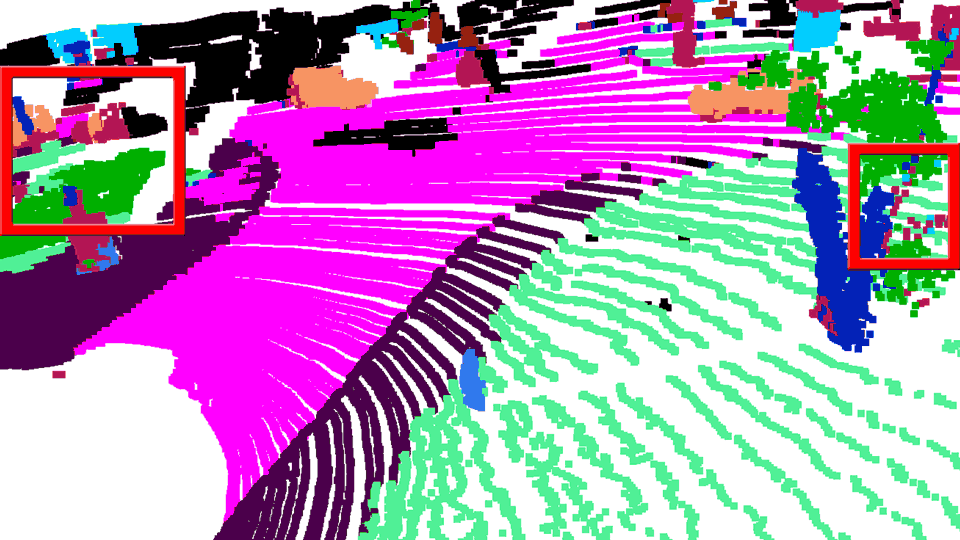}} & \raisebox{-.5\height}{\includegraphics[width=0.23\textwidth]{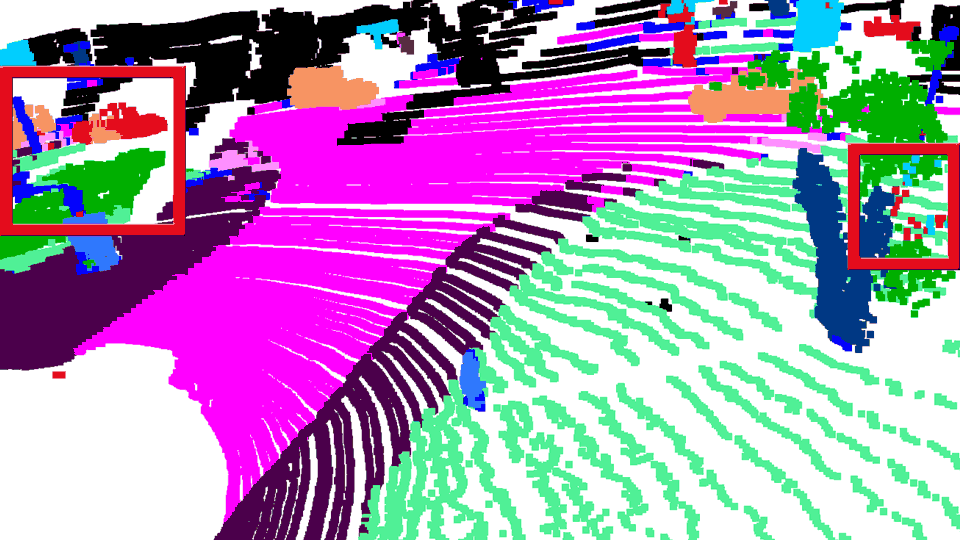}} & \raisebox{-.5\height}{\includegraphics[width=0.23\textwidth]{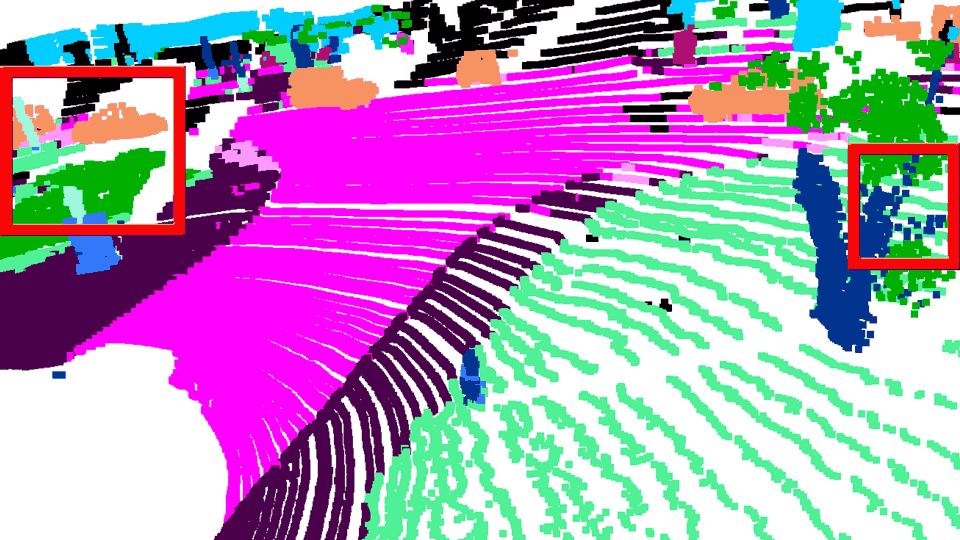}} \\ 
    \rotatebox[origin=c]{90}{08-0000470} \vspace{0.5cm} & \raisebox{-.5\height}{\includegraphics[width=0.23\textwidth]{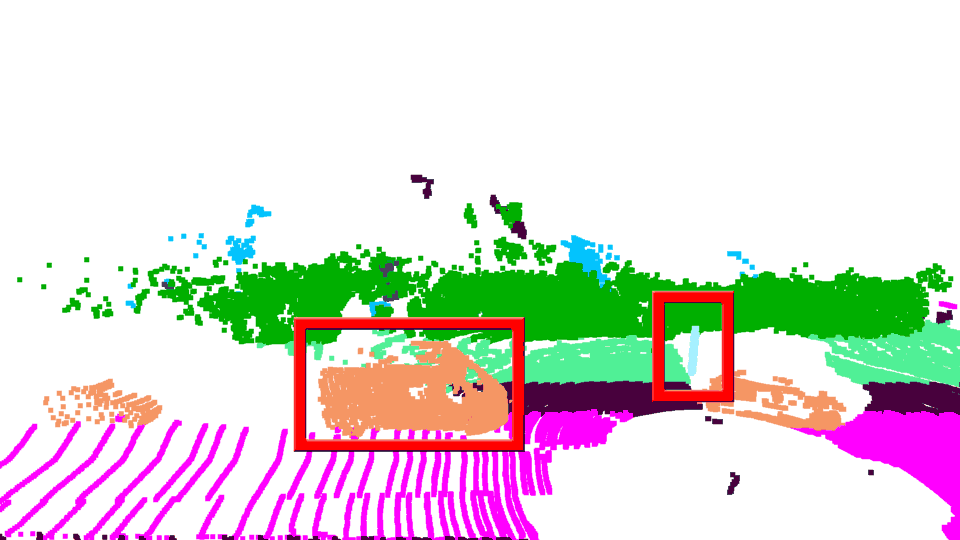}} & \raisebox{-.5\height}{\includegraphics[width=0.23\textwidth]{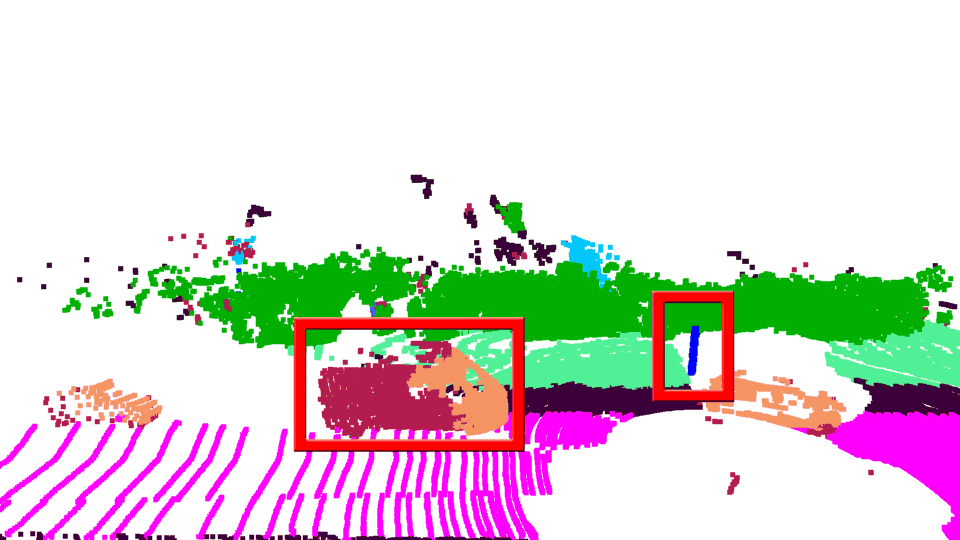}} & \raisebox{-.5\height}{\includegraphics[width=0.23\textwidth]{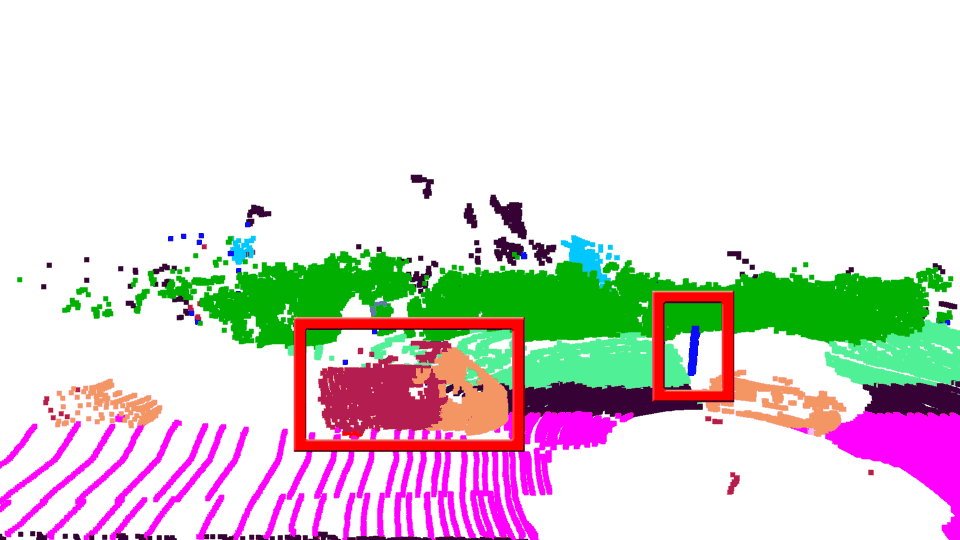}} & \raisebox{-.5\height}{\includegraphics[width=0.23\textwidth]{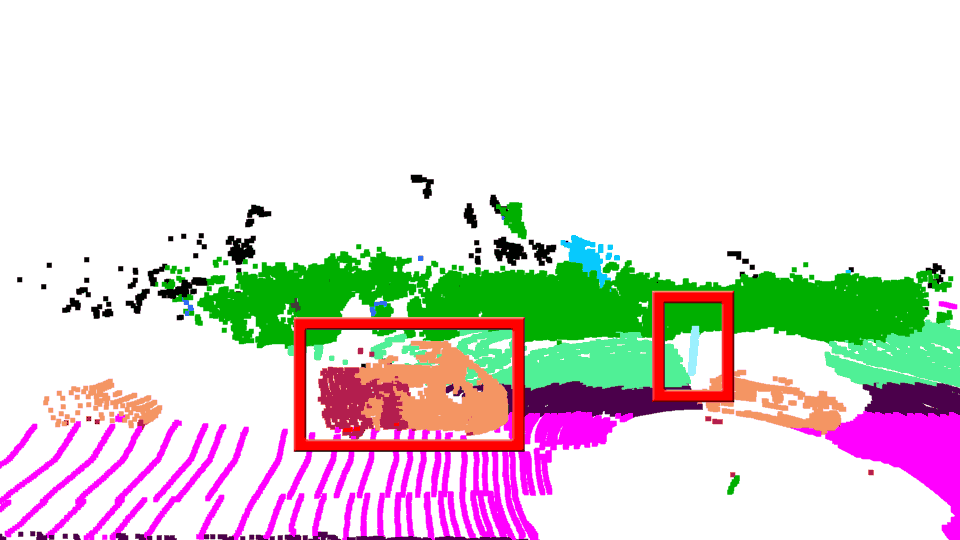}} \\ 
    \rotatebox[origin=c]{90}{08-0001040} \vspace{0.5cm} & \raisebox{-.5\height}{\includegraphics[width=0.23\textwidth]{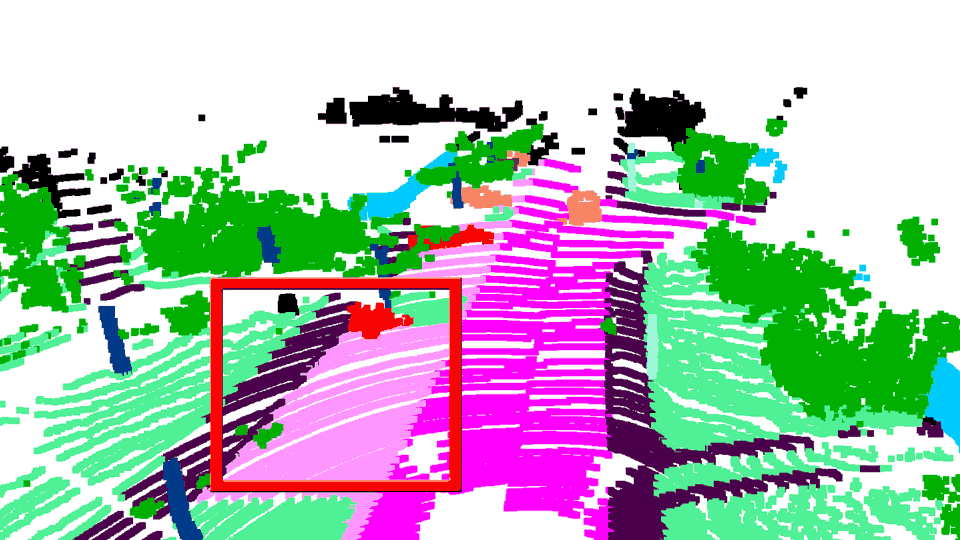}} & \raisebox{-.5\height}{\includegraphics[width=0.23\textwidth]{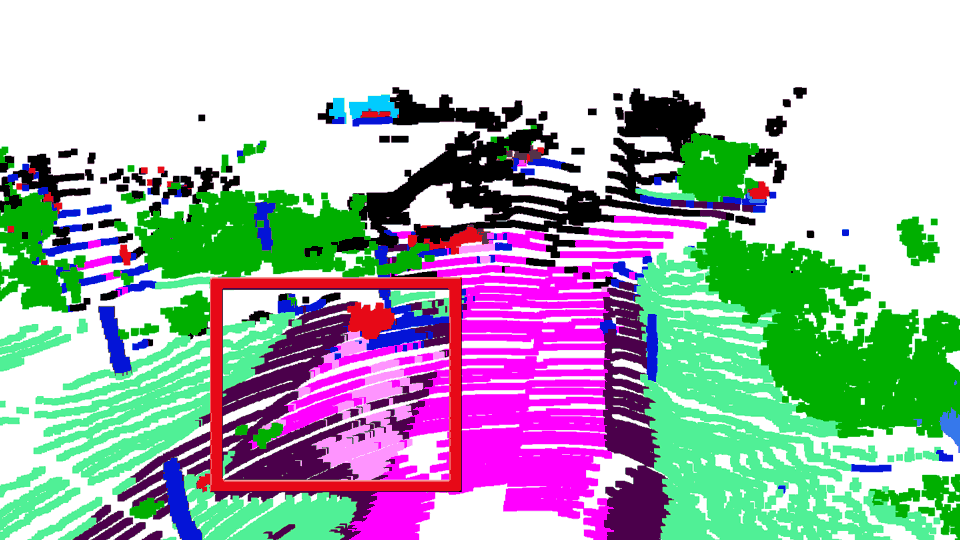}} & \raisebox{-.5\height}{\includegraphics[width=0.23\textwidth]{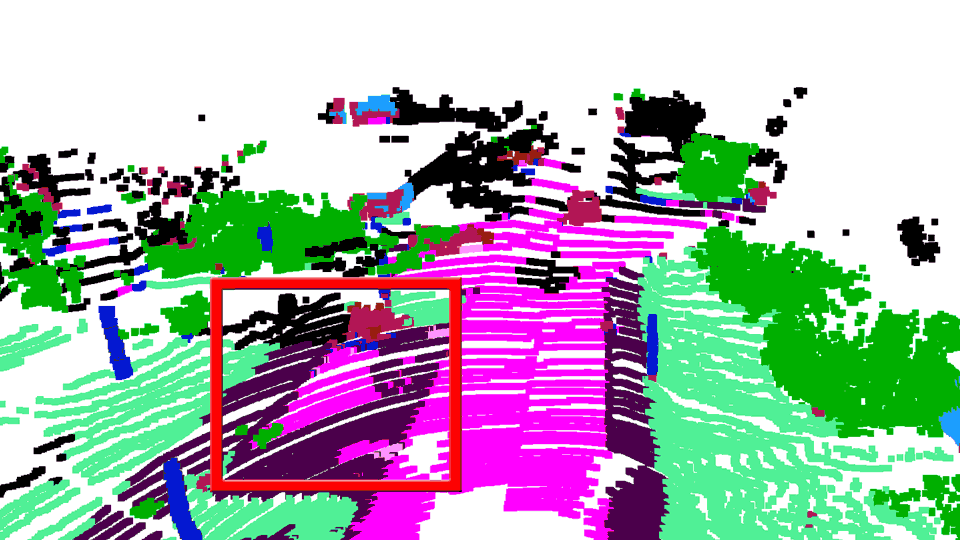}} & \raisebox{-.5\height}{\includegraphics[width=0.23\textwidth]{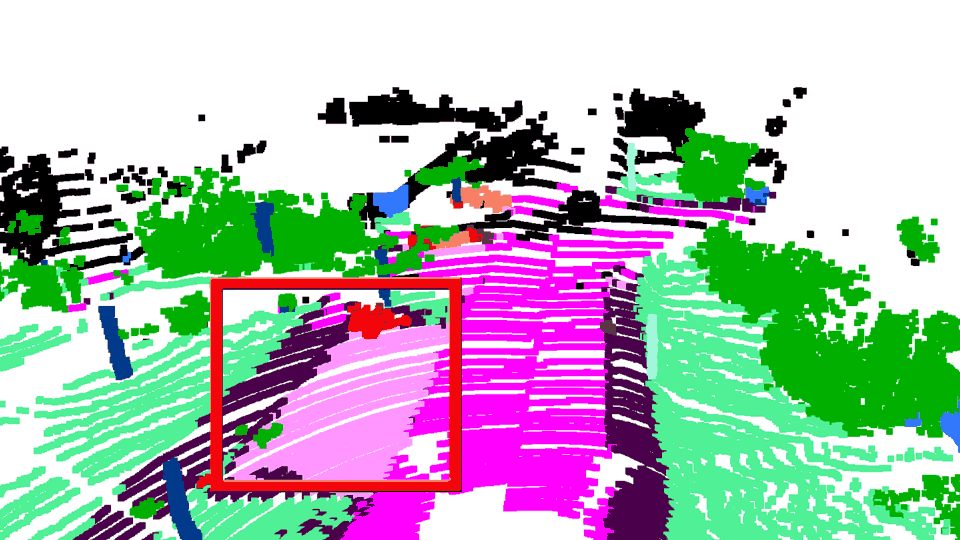}} \\
    \rotatebox[origin=c]{90}{08-0001680} \vspace{0.5cm} & \raisebox{-.5\height}{\includegraphics[width=0.23\textwidth]{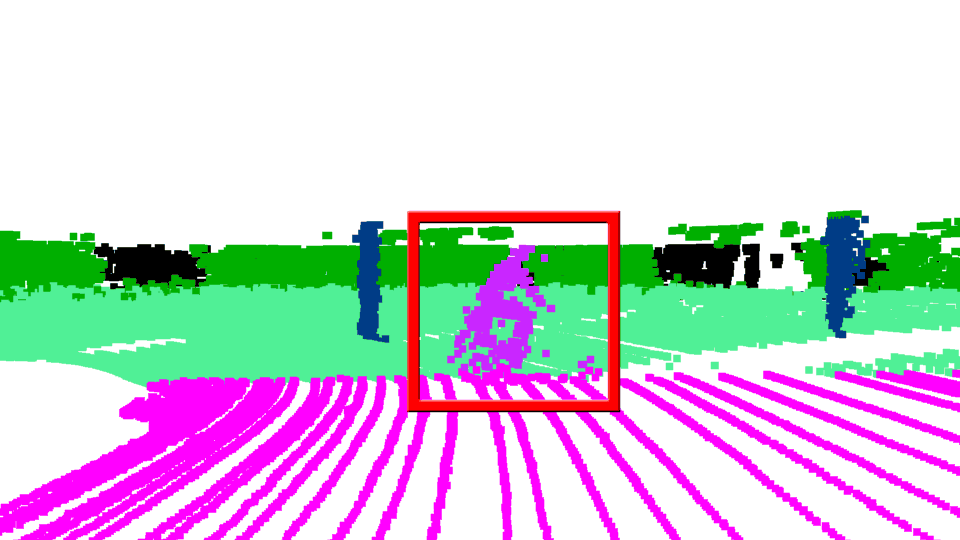}} & \raisebox{-.5\height}{\includegraphics[width=0.23\textwidth]{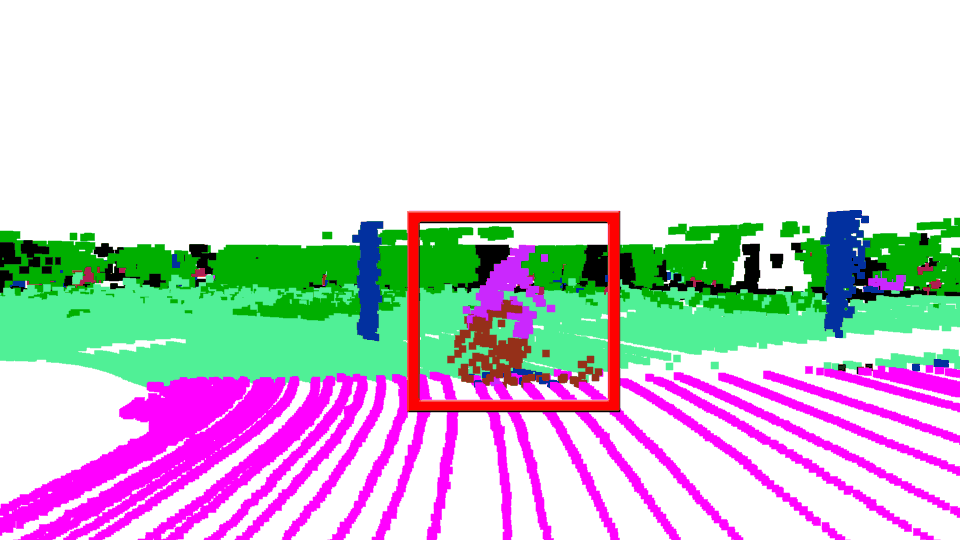}} & \raisebox{-.5\height}{\includegraphics[width=0.23\textwidth]{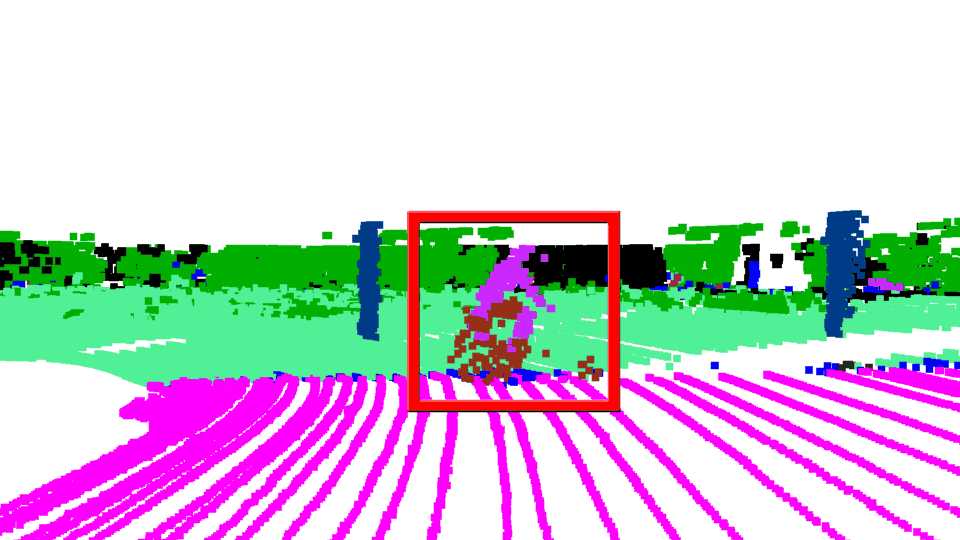}} & \raisebox{-.5\height}{\includegraphics[width=0.23\textwidth]{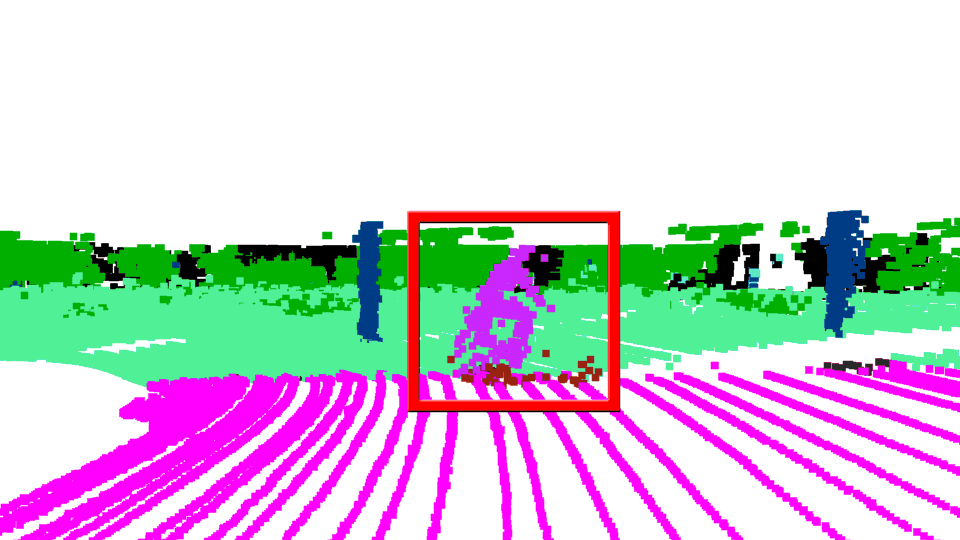}} \\
    \rotatebox[origin=c]{90}{08-0002440} \vspace{0.5cm} & \raisebox{-.5\height}{\includegraphics[width=0.23\textwidth]{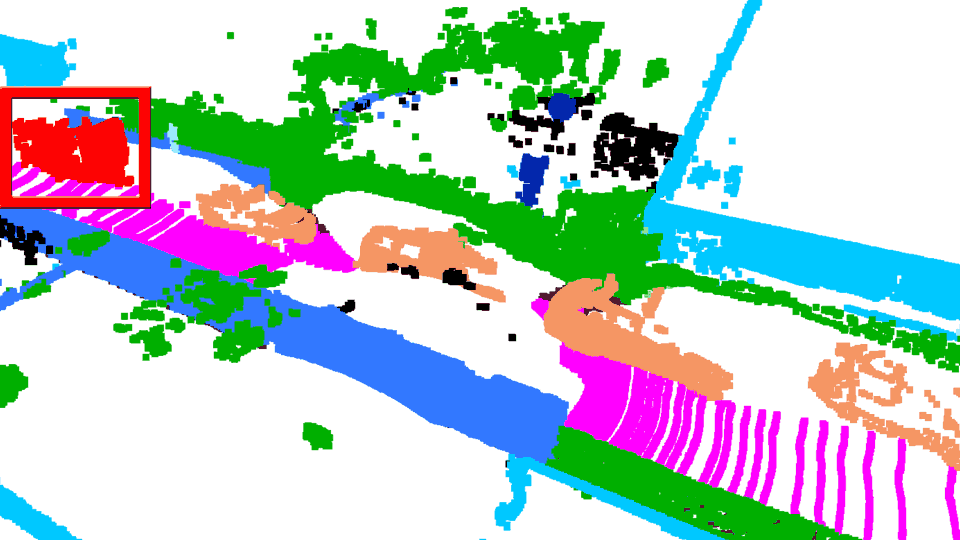}} & \raisebox{-.5\height}{\includegraphics[width=0.23\textwidth]{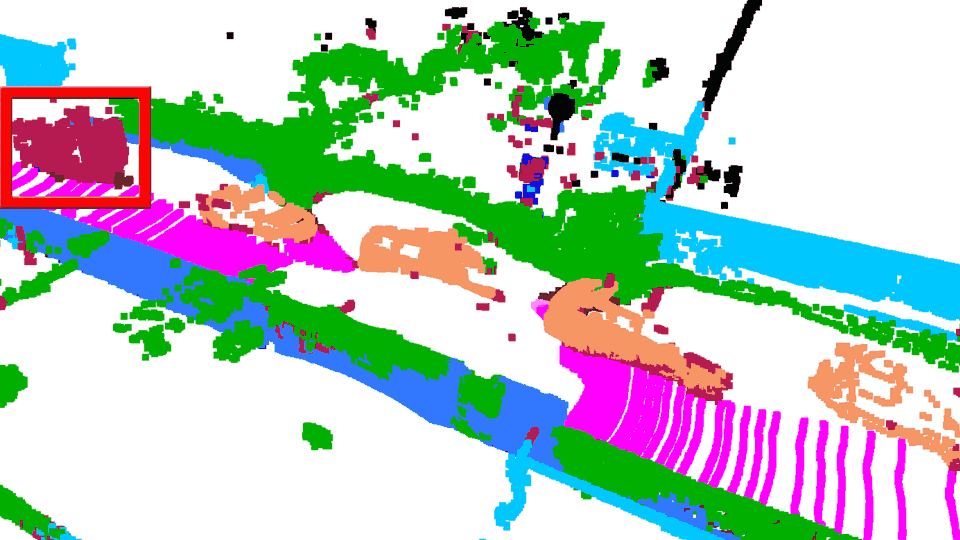}} & \raisebox{-.5\height}{\includegraphics[width=0.23\textwidth]{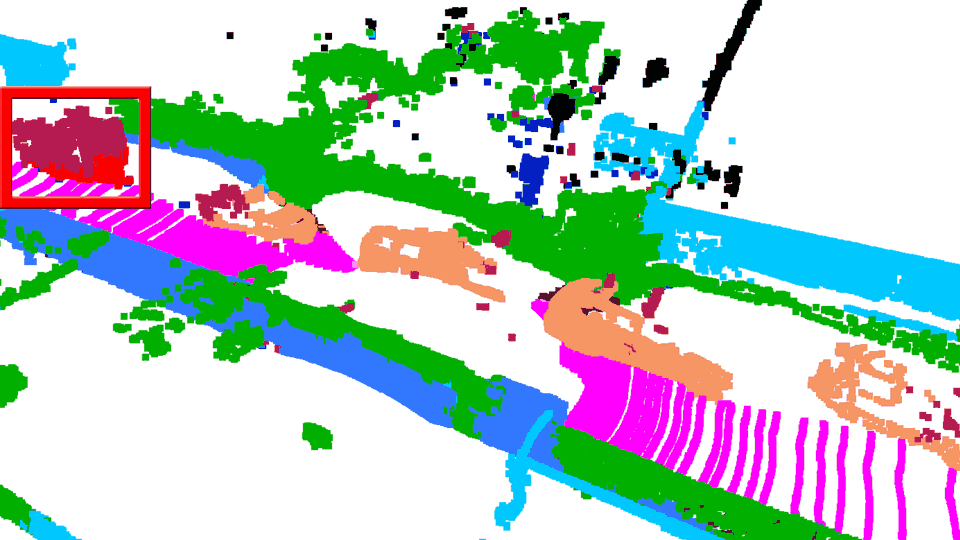}} & \raisebox{-.5\height}{\includegraphics[width=0.23\textwidth]{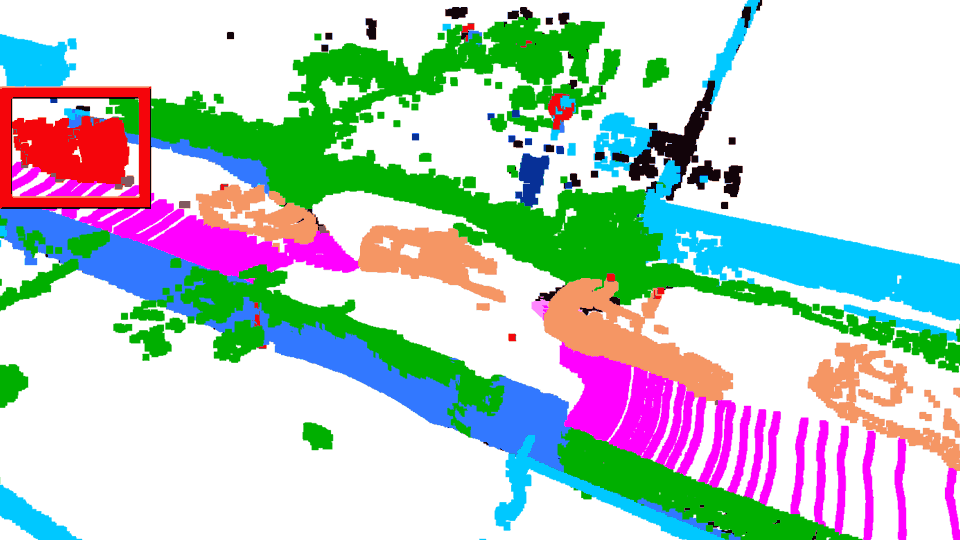}} \\
    \rotatebox[origin=c]{90}{08-0003000} \vspace{0.5cm} & \raisebox{-.5\height}{\includegraphics[width=0.23\textwidth]{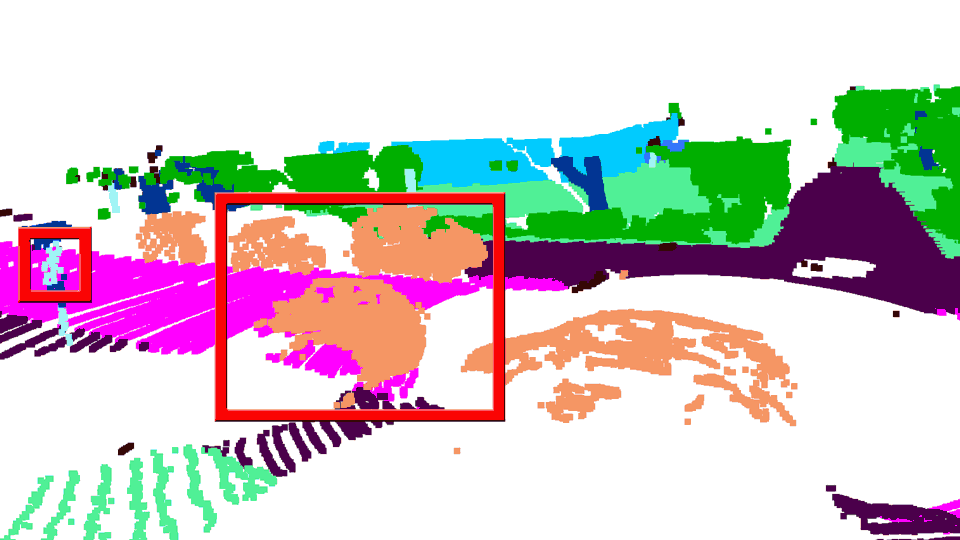}} & \raisebox{-.5\height}{\includegraphics[width=0.23\textwidth]{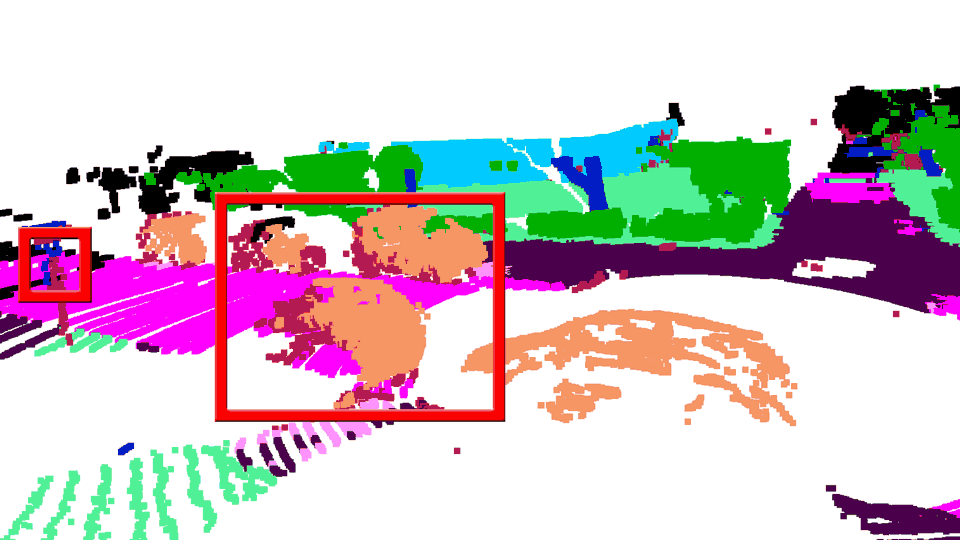}} & \raisebox{-.5\height}{\includegraphics[width=0.23\textwidth]{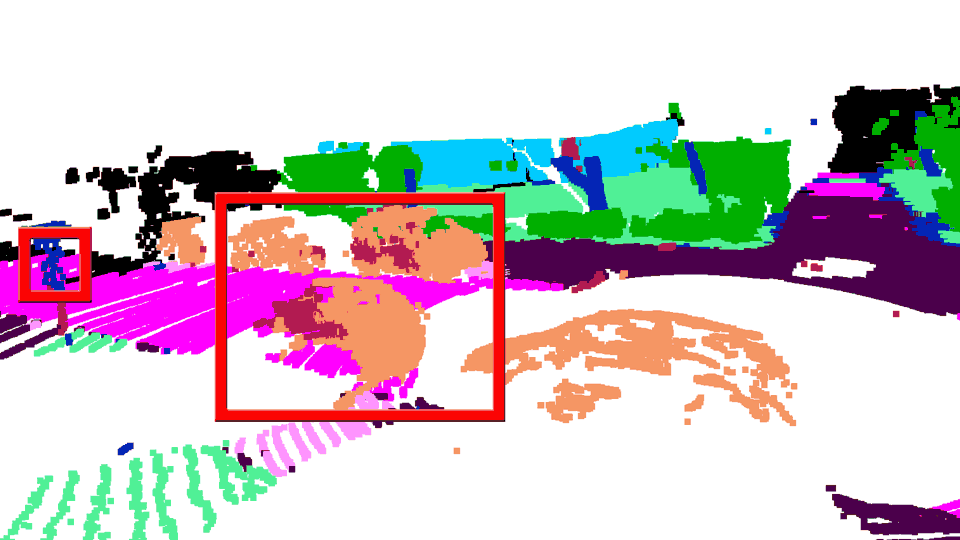}} & \raisebox{-.5\height}{\includegraphics[width=0.23\textwidth]{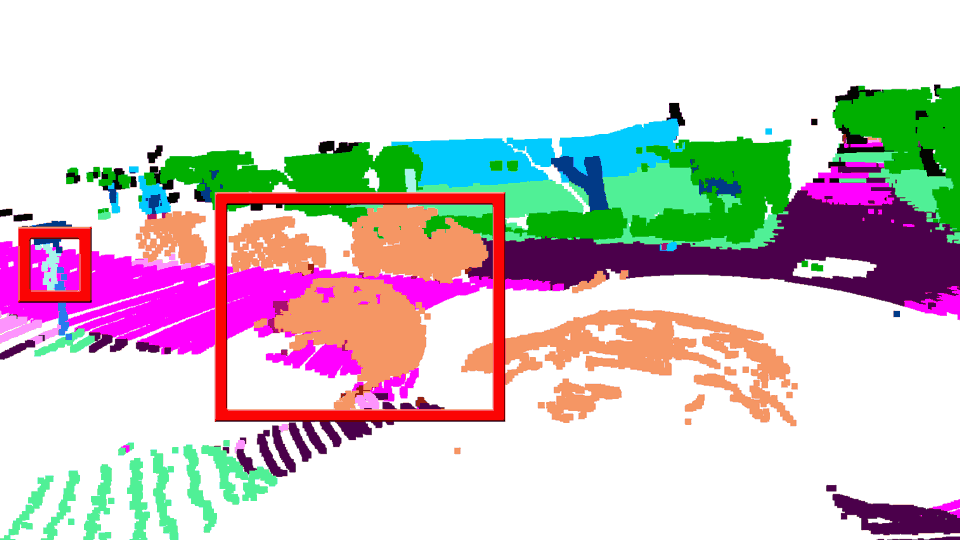}} \\
    \rotatebox[origin=c]{90}{08-0003700} \vspace{0.5cm} & \raisebox{-.5\height}{\includegraphics[width=0.23\textwidth]{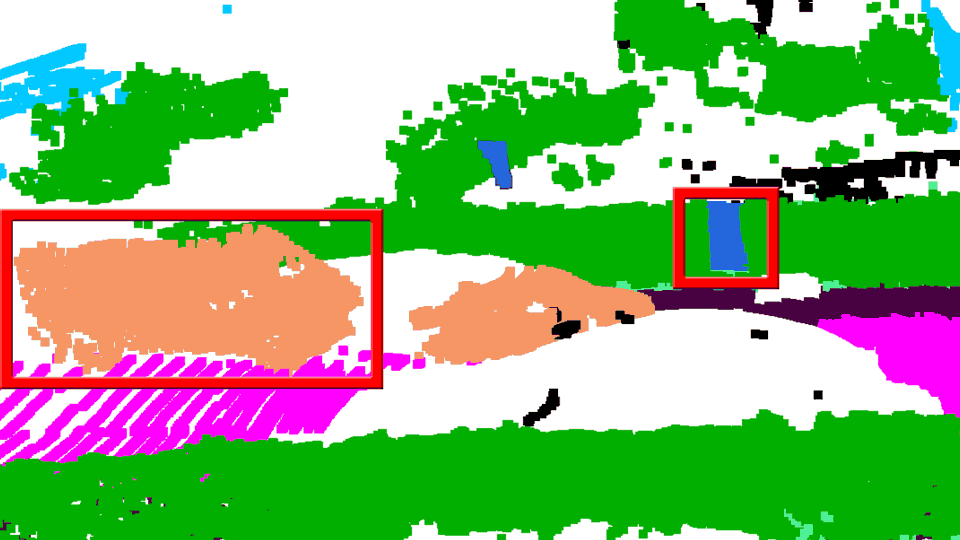}} & \raisebox{-.5\height}{\includegraphics[width=0.23\textwidth]{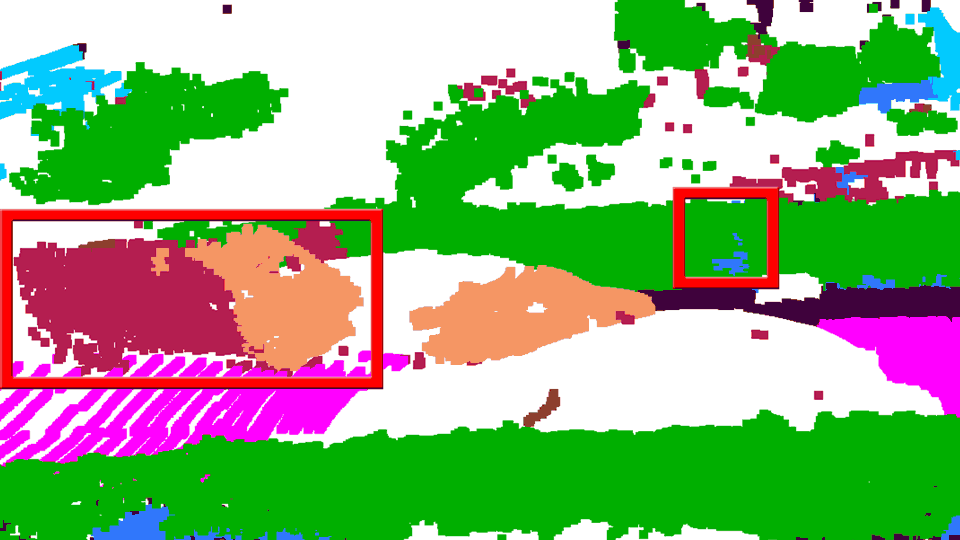}} & \raisebox{-.5\height}{\includegraphics[width=0.23\textwidth]{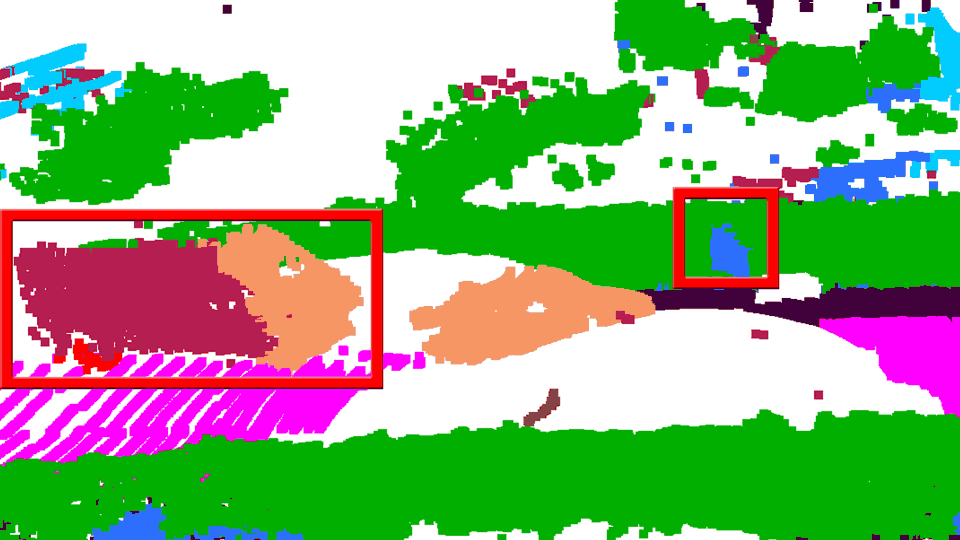}} & \raisebox{-.5\height}{\includegraphics[width=0.23\textwidth]{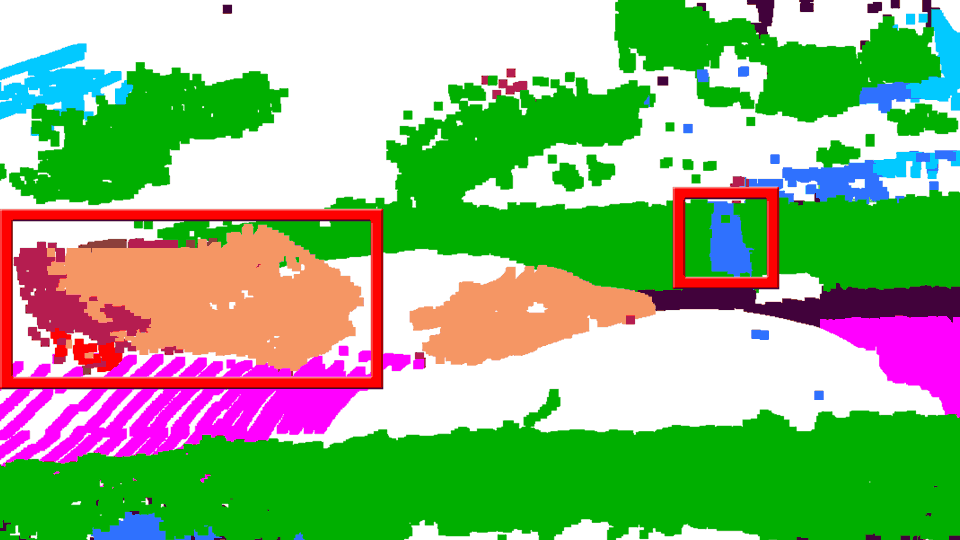}} \\
    \multicolumn{5}{c}{\includegraphics[width=0.9\textwidth]{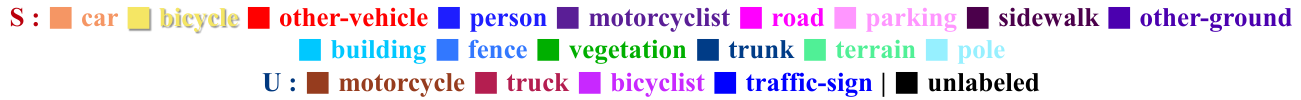}}\\
    \end{tabular}
    \caption{Qualitative comparison of the proposed model with other methods on SemanticKITTI. The labels ``S'' in red denote seen classes, and ``U'' in blue denote unseen classes. } \label{fig:qualitative_comparison_SemanticKITTI}
\end{figure*}

}

\end{document}